\documentclass[runningheads]{styles/llncs}
\usepackage[svgnames]{xcolor}
\usepackage{graphicx}
\usepackage{enumitem}
\usepackage{tikz}
\usepackage{comment}
\usepackage{amsmath,amssymb} 
\usepackage{colortbl}
\usepackage[accsupp]{axessibility} 
\usepackage{capt-of}

\usepackage[pagebackref=true,breaklinks=true,colorlinks,bookmarks=false]{hyperref}

\usepackage{subcaption}
\usepackage{float} 
\usepackage{tabularx} 
\usepackage{xcolor}
\usepackage{blindtext}
\usepackage{graphicx}
\usepackage[export]{adjustbox}
\usepackage{algorithm2e}
\RestyleAlgo{ruled}
\SetKwComment{Comment}{/* }{ */}
\usepackage{pifont}

\makeatletter
\newcommand\tabcaption{\def\@captype{table}\caption}
\newcommand\figcaption{\def\@captype{figure}\caption}
\makeatother
\newcommand{\orcid}[1]{\href{https://orcid.org/#1}{\includegraphics[width=10pt]{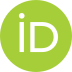}}}
\begin{document}
\pagestyle{headings}
\mainmatter

\title{Max Pooling with Vision Transformers reconciles class and shape in weakly supervised semantic segmentation } 

\titlerunning{ViT-PCM for WSSS by Image-Class Labels}

\author{Simone Rossetti\inst{1,2}\orcid{0000-0002-5344-7872},
Damiano Zappia\inst{1}\orcid{0000-0002-5726-2488},
Marta Sanzari\inst{2}\orcid{0000-0002-4640-9122},\\
Marco Schaerf\inst{1,2}\orcid{0000-0002-2016-1966},
Fiora Pirri\inst{1,2}\orcid{0000-0001-8665-9807}}
\authorrunning{Rossetti et al.}
%
\institute{DeepPlants,
\email{@deepplants.com}\\
\and
DIAG, Sapienza
\email{@diag.uniroma1.it}}

\maketitle
\typeout{--------------------Abstract -----------------------}
\begin{abstract}
Weakly Supervised Semantic Segmentation (WSSS) research
has explored many directions to improve the typical pipeline CNN plus
class activation maps (CAM) plus refinements, given the image-class label as the only supervision.
Though the gap with the fully supervised methods is reduced,
further abating the spread seems unlikely within this framework. 
On the other hand, WSSS methods based on Vision Transformers (ViT) have not yet explored valid alternatives to CAM. ViT features have been shown to retain a scene layout, and object boundaries in self-supervised learning. To confirm these findings, we prove that the advantages of transformers in self-supervised methods are further strengthened by Global Max Pooling (GMP), which can leverage patch features to negotiate pixel-label probability with class probability. This work proposes a new WSSS method dubbed ViT-PCM (ViT Patch-Class Mapping), not based on CAM.
The end-to-end presented network learns with a single optimization process, refined shape and proper localization for segmentation masks. Our model outperforms the state-of-the-art on baseline pseudo-masks (BPM), where we achieve 69.3\% mIoU on PascalVOC 2012 \textit{val} set. We show that our approach has the least set of parameters, though obtaining
higher accuracy than all other approaches. In a sentence, quantitative and qualitative results
of our method reveal that ViT-PCM is an excellent alternative to
CNN-CAM based architectures.

\end{abstract}

\typeout{---------------------Introduction -----------------------}
\section{Introduction}\label{sec:intro}
\begin{figure}[t]
\centering
\includegraphics[width=0.8\linewidth]{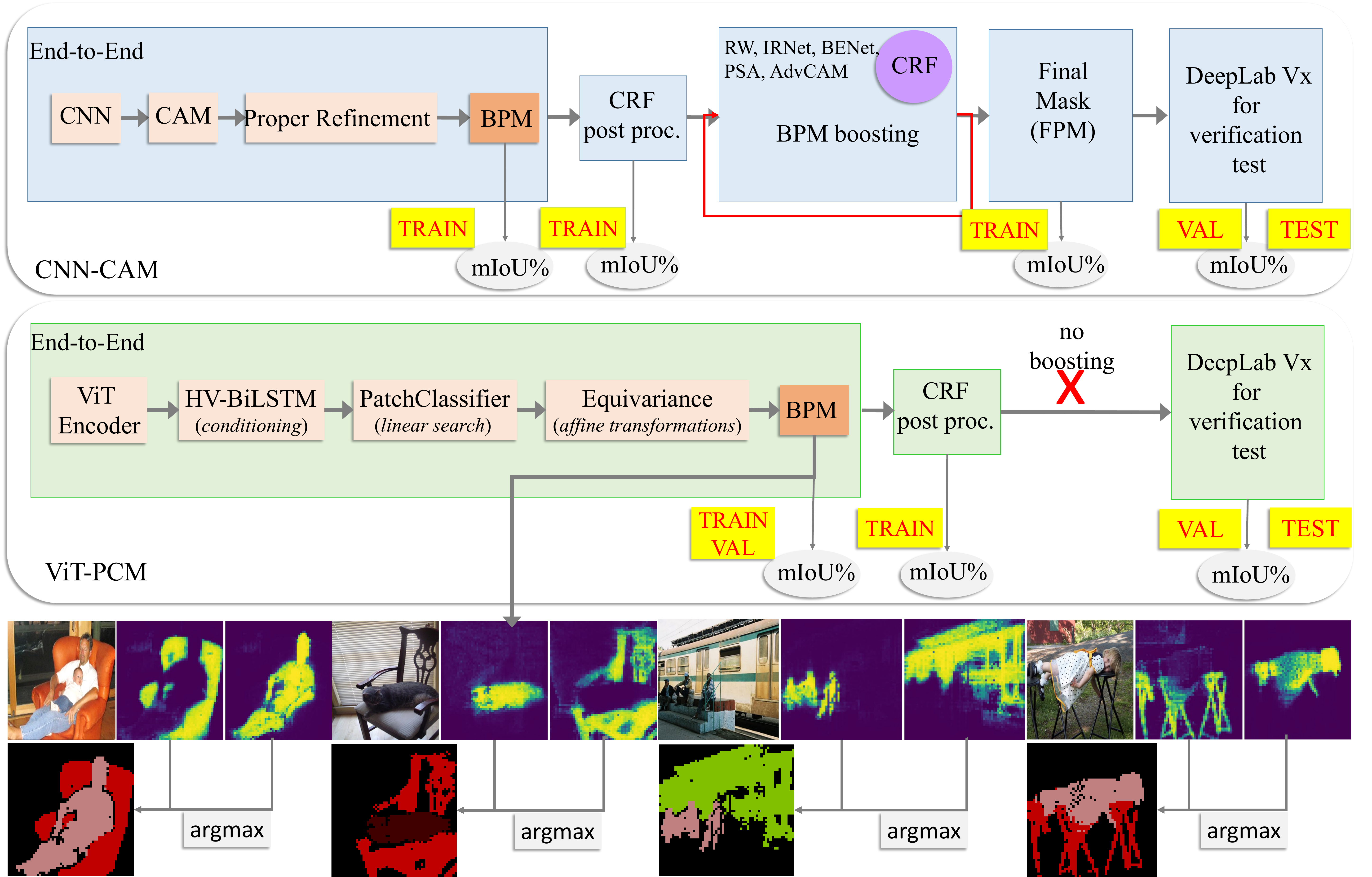}
\caption{The figure compares the basic structure of a CNN-CAM method, above in light blue, with our proposed ViT-PCM method, below in light green. ViT-PCM learns to estimate the BPM, shown in the last two strips, with a single optimization.  Our BPM  are then refined with a CRF (see Figure \ref{fig:ablatiob_argmax}) and, without  further processing, are passed to the verification task (DeepLab). Differently from ViT-PCM, a CAM-based method demands a multi-stage optimization. All recent approaches require boosting the BPM, improved by the CRF, before passing them to the verification task.} \label{fig:good-seeds}
\end{figure}
Weakly supervised semantic segmentation (WSSS) is about segmenting object classes with no pixel-label supervision and using the less demanding supervision possible. The most economic supervision is via image-level class labels, out of which a WSSS method computes pseudo-masks for each object class in an image. To test a WSSS method accuracy,  a supervised segmentation network, such as DeepLab \cite{Chen2018DeepLabSI}, is trained on the devised pseudo-masks, and the induced accuracy is compared with the fully supervised methods. The segmentation task, supervised by the pseudo-mask labels, is a {\em verification task} aiming at demonstrating the computed pseudo-mask quality. In principle, the verification task adds equal improvement to all methods.

So far, methods based on image-level class labels generate pseudo-mask using class activation maps (CAM) \cite{zhou2016learning}. CAM are obtained from a multi-label classification network, such as a CNN. 

CAM limitations in estimating both shape and localization of the classes of interest \cite{chen2022class,guo2019visual,Bae2020,wang2020self} induce many researchers to resort to extra refinements between the baseline pseudo-masks (BPM), often called  {\em seeds}, and the final pseudo-masks production for test verification. These refinements mostly often bring into play multi-stage architectures, as noted in PAMR \cite{araslanov2020single}. 
Several authors resort to saliency maps as subsidiary supervision for good localization \cite{wei2016stc,Lee_2021_CVPR,xu2021leveraging,sun2022inferring,zhou2022regional}. Other authors adopt image operations such as region erasing \cite{singh2017hide,wei2017object}, or region growing 
 to expand the
seed region during training \cite{kolesnikov2016seed,huang2018weakly}, and  multi-scale map fusion to improve background and foreground \cite{wei2018revisiting}. 
Jang {\em et Al.} \cite{jiang2019integral} reviewed the  feature layers selection for CAMs using attribute propagation methods \cite{montavon2017explaining}.  Sun  {\em et Al.} \cite{sun2020mining}  estimate the similarity of the foreground features of the same class with two co-attention networks to capture better the context and \cite{Fan2020CIANCA} look into relations across different images. 

Yet, the greatest success in refinement strategies has been earned  by IRNet \cite{ahn2019weakly},  PSA \cite{ahn2018learning},  and AdvCAM \cite{lee2021anti}. Also,  PAC \cite{su2019pixel} and BENet \cite{chen2020weakly}, have been recently used. For example, SEAM \cite{wang2020self}, Chang  
{\em et Al.} \cite{chang2020weakly} and \cite{shimoda2019self} use PSA;  CONTA \cite{dong_2020_conta} and ReCAM \cite{chen2022class} use IRNet while \cite{lee2021anti} using both.  AFA \cite{ru2022learning} use PAC \cite{su2019pixel}. 

CRF\cite{krahenbuhl2011efficient} are trained on PascalVOC, fully supervised, and introduced in WSSS by \cite{kolesnikov2016seed}. CRF used as post-processing out of a training loop, improve the BPM, on average, 3-4\% mIoU, on Pascal VOC 2012. On the other hand, multi-stage methods, refining BPM with IRNet \cite{ahn2019weakly},  PSA \cite{ahn2018learning},  and AdvCAM \cite{lee2021anti} use dense CRF in the training loop, which gives a substantial boost in accuracy.   Using dense CRF, optimized on PascalVOC, likewise using saliency (e.g. \cite{hou2017deeply}, which operates dense CRF too)   in the refinement loop to obtain the final pseudo-mask, beside being resource intensive,  fails to generalize a method beyond the PascalVOC dataset. This lack of generalization power is common to any  WSSS approach using biased methods in a refinement training loop. 

\begin{figure}[t!]
  \centering
  \includegraphics[width=1\linewidth]{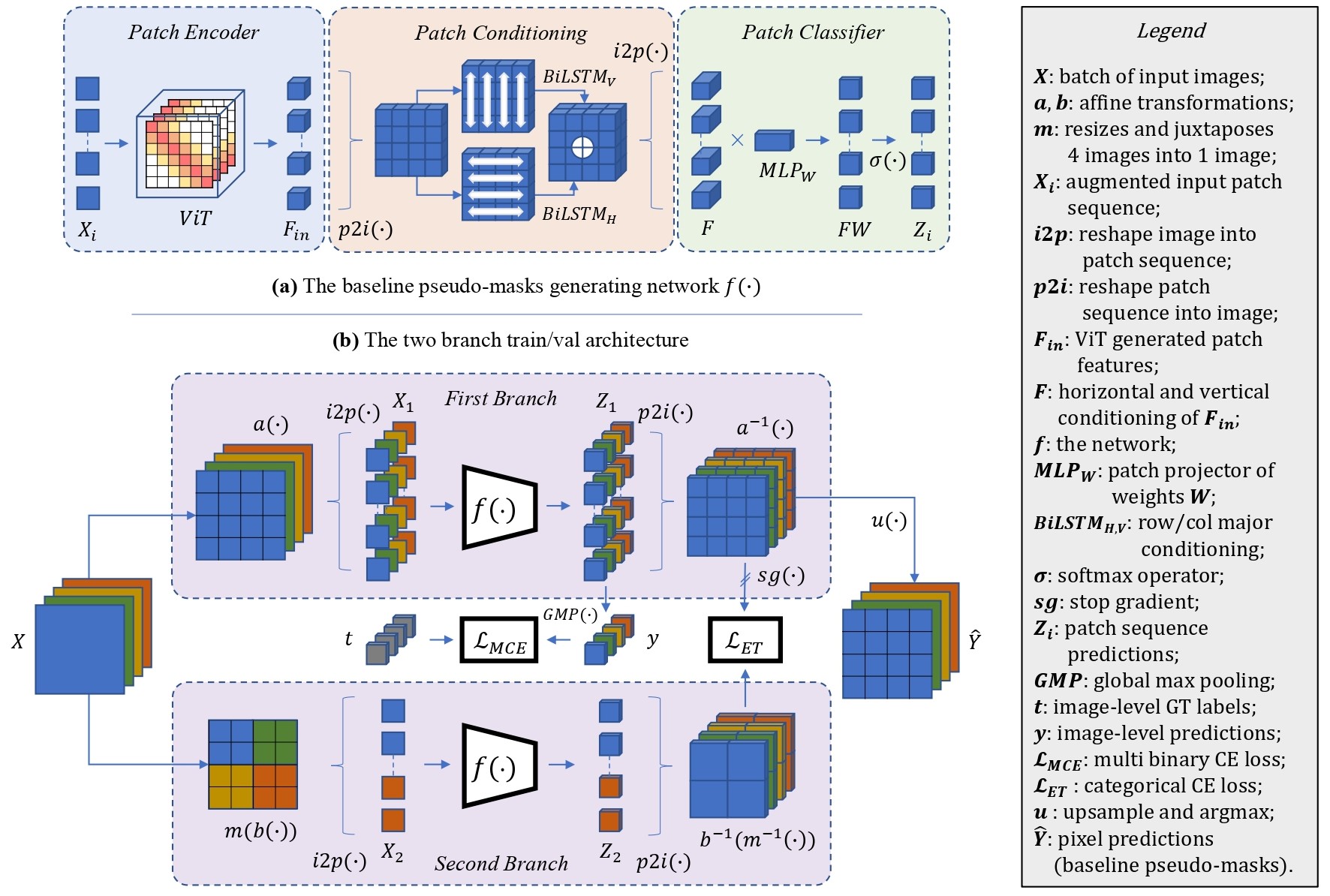}
   \caption{The above schema shows the end-to-end ViT-PCM, a semantic segmentation method supervised by image-level class labels $t$. The plate in (a) shows the core network $f(\cdot)$ implementing the {\em linear search method}, which maps the image-level class labels to patch-labels. The plate (b) shows the two-branches architecture, including $f(\cdot)$ in both branches.  }
   \label{fig:network}
\end{figure}

The challenge is to raise the bar of the baseline pseudo-mask accuracy so that the only supervision truly sticks to the image-level label. To this end, we introduce a new model for computing pseudo-masks, which bypasses the CAM bottleneck.
The main contributions of the paper are the followings:
\begin{itemize} \setlength{\parskip}{0pt}  \setlength{\itemsep}{0pt plus 1pt}
\item  We introduce a novel model for weakly supervised semantic segmentation (WSSS) based on ViT \cite{dosovitskiy2021image}. The model, dubbed ViT-PCM, is represented in Figure \ref{fig:network}. 
\item We propose a new pseudo-mask computation method {\em Explicit Search}  without resorting to CAM. The method leverages the locality properties of ViT to come close to an effective mapping between multi-label classification and semantic segmentation. We use the Global Max Pooling (GMP) to fetch the relevance of each patch, given the patches' categorical distribution over the classes of interest. This way, we project the patch features to class predictions (PCM) using a multi-label BCE loss (MCE).   We ensure equivariance to translation and scaling transformations defining two branches, see Figure \ref{fig:network}. 
\item The proposed pseudo-mask computation outperforms all state-of-the-art methods: we obtain BPM accuracy of 67.7 mIoU\% on Pascal VOC 2012 \textit{train} set which improves the current best BPM (\cite{ru2022learning}) of 3.91\% . On average, we improve more than 5\% mIoU than all the other competitors. On MS-COCO 2014 we obtain 45.03\% mIoU on \textit{val} set.
\item For the verification task, using  DeepLab as a segmentation method,  we do not need to boost our BPM to obtain masks more suitable for DeepLab, yet we obtain comparable validation and test scores.
\item We also prove the advantages of our method in terms of computational effort. In particular, we obtain the final segmentation with 89.4 M of parameter size, the minimal cost amid competitors. 
\end{itemize}
\vspace{-\topsep}
Beyond the novelty of our contribution, which is the first proposal to compute pseudo-mask baselines bypassing CAM, we show that both quantitative and qualitative results prove that exploring new methods for baseline pseudo-masks can be rewarding. We establish a new state of art on baseline pseudo-mask computation, using image-level class labels without refinement. 

\typeout{---------------------Related -----------------------}
\section{Related Works}\label{sec:related}

Current WSSS methods mostly operate with image-level class labels as the cheapest supervision. Approaches using image-level class labels are based so far on CAM \cite{zhou2016learning} methods using a plain multi-label classification network. The class activation maps are obtained via the global average pooling (GAP) averaging the feature maps of the last layer, further concatenated into a weights vector. This last is connected with the class prediction, using a BCE prediction loss. More recently,  Vision Transformers \cite{dosovitskiy2021image}  are emerging as an alternative to generate CAM \cite{xu2022multi,ru2022learning}. Our method is the first one using only  ViT without CAM to generate baseline pseudo-masks. 
\vskip 0.5\baselineskip
\noindent
{\bf CNN plus CAM.}\  These methods contribute to two complementary research directions:  {\em Baseline Pseudo-Mask generation}, to control and expand the activation of CAM regions, and  {\em Pseudo-Mask refinement} to obtain the full mask of objects.   

\noindent
{\em Baseline Pseudo-Mask generation} extends  CAM by revising the loss, or by augmenting the dataset, or by perturbing CAM devised regions, or using pretrained saliency maps. In ReCAM \cite{chen2022class} the authors propose softmax cross-entropy (SCE)  as a valid solution for CAM, since it bypasses the non-exclusive class problem of BCE. In OoD \cite{lee2022weakly} the authors propose an out of distribution dataset taken from OpenImages \cite{kuznetsova2020open}, to better capture background semantics. Other methods to expand CAM perturb the generated regions to capture new areas \cite{kweon2021unlocking,stammes2021find,lee2021anti}, by either erasing or masking. 
Since \cite{wei2016stc}, pretrained methods for saliency detection and saliency maps have been adopted in \cite{oh2017exploiting,zhang2020splitting,Lee_2021_CVPR,yao2021non,wu2021embedded,xu2021leveraging}, and in \cite{jiang2019integral,jiang2021online}. The latter propose an online attention accumulation (OAA) strategy based on attribute propagation methods.  Pseudo-mask generation is contaminated by self-supervised learning in \cite{wang2020self},  via downstream tasks and transformations ensuring CAM features equivariance, or via contrastive representation learning, as in RCA \cite{zhou2022regional}, C2AM \cite{xie2022c2am} and PPC \cite{du2022weakly}.

\noindent
{\em Pseudo-Mask refinement.} In recent works, all CAM-based approaches explore refinement strategies, ensuring some control on pixel-level labelling. The most common strategies are PSA \cite{ahn2018learning},  AdvCAM \cite{lee2021anti} and IRNet \cite{ahn2019weakly}. PSA  refines the baseline masks by propagating pixel semantic values to their neighbours, collecting confidence for the target classes. AdvCAM \cite{lee2021anti} uses iterative adversarial climbing performed on an image to iteratively involve its features in the classification to increase CAM confidence in activated regions. IRNet \cite{ahn2019weakly} explores class equivalence relations of pixels and refines pixel-labels by evaluating the displacement w.r.t. computed centroids. 
Recently BENet \cite{chen2020weakly} has been used for pseudo-mask baseline refinement, too; it refines object boundaries, together with foreground and background. We observed in the introduction that all these strategies use in the training loop dense CRF of \cite{krahenbuhl2011efficient}, which is trained on PascalVOC2012.

\vskip 0.5\baselineskip
\noindent
{\bf Transformers.} ViT have so far gathered a significant success with self-supervised learning \cite{dosovitskiy2021image}, as witnesses Dino \cite{caron2021emerging}, \cite{chen2021empirical,li2021efficient}, and recently SDMP \cite{ren2022simple}.  Dino \cite{caron2021emerging} downstream task segments foreground from background for single class images, differing from WSSS. Only recently ViT contributed to WSSS with MCTformer \cite{xu2022multi} and AFA \cite{ru2022learning}, though both resort to CAM. MCTformer exploits ViT attention mechanism to obtain localization maps. To generate pseudo-masks, they resort to PSA \cite{ahn2018learning}. AFA uses ViT multi-head-self-attention (MSA) to capture global dependencies and develop an affinity-based attention module to propagate the initial pseudo-masks, namely the obtained CAM. Refinement of the initial pseudo-mask is attained by affinity propagation with RAWK \cite{vernaza2017learning}, in turn, pretrained on a scribble dataset. 

Differently from the above approaches, we use ViT as a backbone for building our explicit search method. Indeed, we devise an end-to-end internal refinement to obtain a baseline pseudo-mask (BMP) without resorting to external strategies.

\typeout{---------------------Motivations -----------------------}
\section{Motivations of using ViT and bypass CAM}\label{sec:motivations}
\begin{figure}[t]
\centering
\includegraphics[width=0.7\linewidth]{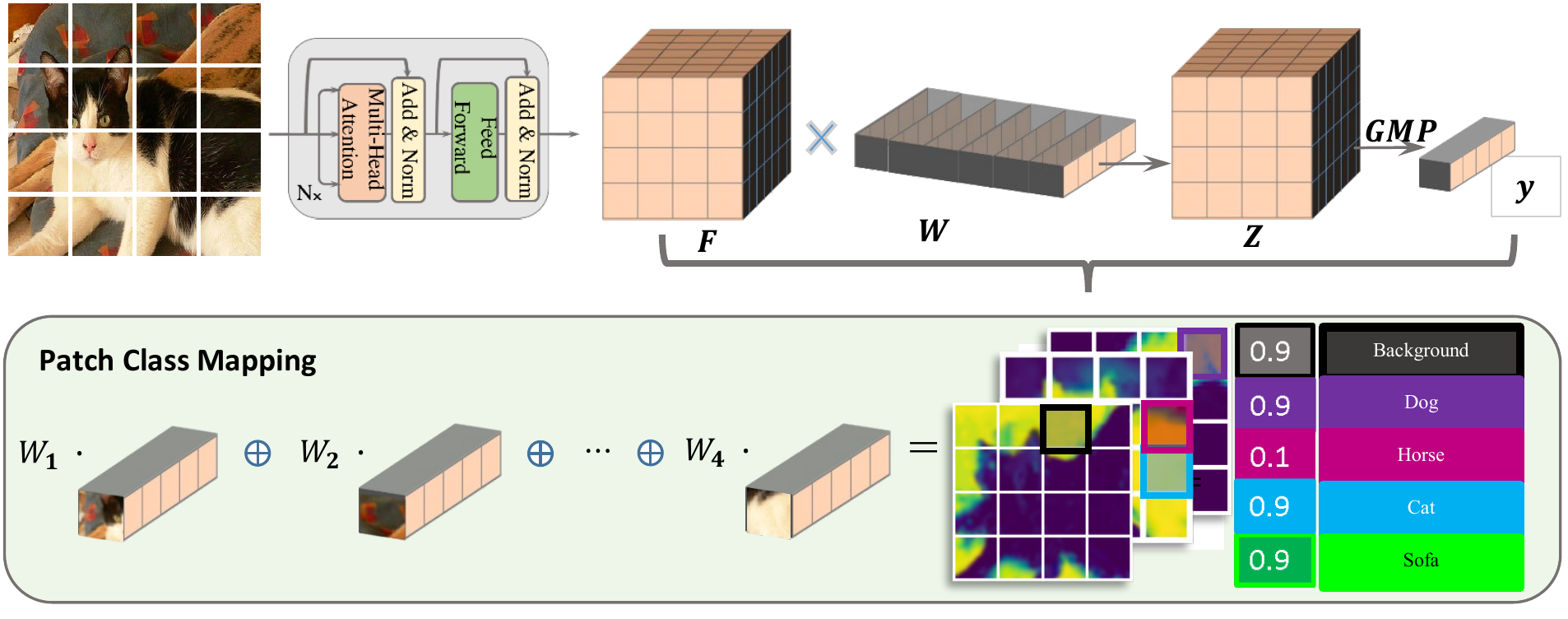}
\caption{PCM: Patch Class Mapping.} \label{fig:lls}
\end{figure}
At the core of semantic segmentation, supervised by image-level labels, is the mapping between multilabel image classification and pixel-level classification. 
This mapping requires linking the abstract image feature space, encoding classes into an index vector, to a completely different space in which features encode classes into a fine grid structure.  
How could this be possible? CNN have an inductive bias on the image features local structure because of convolution kernels,  which  CAM leverages. The inductive bias of CNN entitles CAM to indicate the pixels which mainly contribute to the specific class prediction. The produced map is appealing though misleading: it does not induce a mapping between image features and pixels. 

On the other hand, ViT \cite{dosovitskiy2021image} have much less bias because images are split into flattened patches and encoded. Thus, the spatial relations are learned from scratch using attention and position embedding. This learning from a {\em tabula rasa} generates a number of basis functions for each patch, specifying their internal structure. These basis functions account implicitly for the class a patch belongs to. On these grounds, the mapping problem amounts to unravelling the implicit class representation brought on by the patch principal components. Our proposed {\em explicit search method} models this mapping.

We describe here the intuition. Let us assume that patches are pixels, the classes (categories) are denoted by ${\mathcal C}$, having cardinality $K$ and $X{\in} \mathbb{R}^{h {\times} w {\times} 3}$ is an image. Let also assume that the ViT inferring the image multiclass labels is the function $f(\cdot|\varphi)$ with parameters $\varphi$, mapping an image $X$ to a vector of values in $(0,1)$ for each category in ${\mathcal C}$. On the other hand, let us represent the basis functions specifying the patches' internal structure, implicitly accounting for the patch classes, by a tensor $Z$. We shall see below how Z is computed. $Z$ has height and width as the image $X$, and it also has a third axis for the categories ${\mathcal C}$. We make $Z$ a stochastic tensor along the categories axis: summing up along that axis, we obtain a matrix of ones. Let $f(\cdot | \theta)$, with parameter $\theta$, play the role of the segmentation model; namely, it evaluates the likelihood that a patch of the original image belongs to some precise class in ${\mathcal C}$. 

We argue that  Global Max Pooling (GMP) relates the two models $f(\cdot | \theta)$ and $f(\cdot|\varphi)$ as follows. 
Let $Z^k$ be the slice of $Z$, along the categories axis, which should specify the patches internal structure for the category $k\in {\mathcal C}$. $GMP$  selects the most relevant element of $Z^k$, namely the element with the highest confidence to belong to the category $k$, and returns a probability value $y^k$ that it  belongs to class $k\in {\mathcal C}$.
The selected element $Z_{ij}^k$,  at the same time, is the one in highest consideration to tell whether or not the category $k$ appears in the image. In this way, GMP links image class prediction and patch class prediction. 
\typeout{--------------------Method -----------------------}

\section{The explicit search method}\label{sec:explicit_search}

This section considers the optimization method leading to estimating the map between image classes and patch classes. The  end-to-end architecture enclosing the method is described in Figure \ref{fig:network}, and in Section \ref{sec:architecture}. 

Let us indicate by $f$ the network taking inputs from a dataset $\mathcal{D}{=}\{\langle X_{in},t\rangle\}$.  Here $X_{in} {\in} {\mathbb R}^{h{\times} w{\times}3}$ indicates an input images, possibly obtained from an augmented and transformed set,  $t{\in} \{0,1\}^{K}$ are the ground truth binary labels, and $K$ is the number of classes defined by the category set ${\mathcal C}{=}\{0,1,{\ldots}, K\}$. The output of $f$ is a tensor $\hat{Y}\in {\mathcal C}^{h{\times} w}$ which is a {\em baseline pseudo-mask}.

ViT is part of $f$. We recall that ViT partitions the image $X$, resized image of the original  $X_{in}$,  into $s$ patches of size $(d{\times} d {\times} 3)$. In particular, we are interested in the feature maps $F{\in} {\mathbb R}^{s\times e}$, with $s{=}(n/d)^2$, with $n{=}w{=}h$.  The feature maps $F$  are the encoded representations of the patches, obtained by ViT.  $F$ represent the basis functions specifying the patches internal structure.  

\subsubsection{Explicit search by Global Max-Pooling}
Given $F{\in} {\mathbb R}^{s\times e}$, we consider also a weight matrix $W{\in}{\mathbb R}^{e{\times}K}$  whose weights are taken into account in the  optimization method described below. More precisely, we estimate the baseline pseudo-mask $\hat{Y}$, training the weights $W$  with only image-level class labels as supervision,  minimizing the multilabel classification error. 

The first objective  is to minimize the multilabel classification prediction error (MCE). Thus, given the ground truth binary labels $t$ defined above, and recalling that $K$ are the number of classes,  we model the multi-label classification using $K$ independent Bernoulli distributions and $K$ binary cross-entropy losses (BCE):
\begin{equation}\label{eq:bce}
    \mathcal{L}_{MCE}=\frac{1}{K}\sum_{k\in{\mathcal C}} BCE(t_k,y_k)=-\frac{1}{K}\sum_{k\in {\mathcal C}} t_k\log(y_k)+(1-t_k)\log(1-y_k).
\end{equation}
Let us consider first how $y{\in} {\mathbb R}^K$  is obtained.  Let:
\begin{equation}
A = FW\ \mbox{ and } Z = softmax(A) \mbox{, with } F{\in} {\mathbb R}^{s\times e}, W{\in}{\mathbb R}^{e{\times}K} \mbox{ hence } Z{\in}{\mathbb R}^{s{\times}K}.
\end{equation}
$Z$ represents the semantic segmentation predictions, needing to be projected into class predictions\footnote{Note that we are representing here $Z$ as a matrix, which is simply a reshaping of the tensor $Z$ discussed in Section \ref{sec:motivations}.}. We do so using Global Max Pooling (GMP):
\begin{equation}\label{eq:gmp}
y_k = GMP(Z^k) = \max (Z^k) = Z^k_i, \mbox{ for some  } i{\in}\{1,{\ldots},s\}.
\end{equation}
Here:
\begin{equation}\label{eq:output}
Z^k = softmax(A^k) \mbox{ and }  A_{j}^k= F_jW^k
\end{equation} 
The feature maps $F$  are the encoded representation of patches $U$, and $F_j$ is the feature map of patch $U_j$, while  $A_j^k$ is the logit of patch $U_j$, $j=0,{\ldots},s$ with respect to class $k\in\{0,1,{\ldots},K\}$. 

Given the vector $y_k$, we show how the optimization obtains the terms separating the feature space by the relative error backpropagation of ${\mathcal L}_{MCE}$, with respect to weights $W$. Computing the gradient of Eq. (\ref{eq:bce}) w.r.t. the weight $W$, we obtain:
\begin{equation}
    \frac{\partial \mathcal{L}_{MCE}}{\partial W}=\sum_{k\in{\mathcal C}}\frac{\partial BCE(t_k,y_k)}{\partial W}
\end{equation}
Let us analyze the gradient of the weights $W$, with respect to each column $h$, of size $e$, with $h{\in}\{0,1,{\ldots}K\}$.  Applying the chain rule, w.r.t. the generic class $k$:

\begin{equation}\label{eq:chain}
    \frac{\partial BCE(t_k,y_k)}{\partial W^h}=\frac{\partial BCE(t_k,y_k)}{\partial y_k}\frac{\partial Z_i^k}{\partial A^h}\frac{\partial A^h}{\partial W^h}
\end{equation}
Here we used the fact that $y_k{=}max(Z^k)$, and  $max(Z^k){=}Z^k_i$ from eq. (\ref{eq:gmp}). Therefore,  the gradient dimension is $\frac{\partial BCE(t_k,y_k)}{\partial W^h}{\in} \mathbb{R}^{e}$. 
The derivation of each term is provided in the supplementary.


Let   us select, now,  the column $h$ of the weights $W$, this column will be updated by the quantity:
\begin{equation}\label{eq:backproploss}
\begin{split}
   \frac{\partial \mathcal{L}_{MCE}}{\partial W^h}&=
   \frac{\partial BCE(t_h,y_h)}{\partial W^h}
   +
   \sum_{k{\in} {\mathcal C}, k{\neq} h}\frac{\partial BCE(t_k,y_k)}{\partial W^h}\\
   &= -F_{i_h}(t_h - y_h) + \sum_{k{\in} {\mathcal C}, k{\neq} h} F_{i_k}Z_{i_k}^h \frac{t_k -y_k}{1-y_k}   
\end{split}
\end{equation}
Note that here the subscripts $i_h, i_k$ in $F$ and $Z^h$ indicate, respectively, the indexes at which  $Z_i$ have maximum value, w.r.t classes $h$ and $k$, where $F_i$ is obtained by the last two terms of equation \ref{eq:chain}, r.h.s.  We are using these indexes only in the updating rule for the weights; we are not using them in the derivation.

Eq. \ref{eq:backproploss} specifies the linear-search mechanism of the proposed optimization, iteratively selecting the most representative features $F_{i_h}$ of each category $h$. At each step, the optimization  updates the full column rank matrix $W{\in} {\mathbb R}^{s{\times}e}$ and returns the minimum error norm solution, which  separates the feature vector space $\mathbb{R}^e$ into $K$ linear sub-spaces. 
Considering the optimization manifold,   the vector  $W^h$ moves  in the direction of the best representative feature vector $F_{i_u}$, with either $u$ being of the same category of the chosen column $h$, or not. More precisely, at each iteration, $W^h$ moves in the direction of $F_{i_h}$  according to the error value $(t_h - y_h)$, and in the direction $F_{i_k}$ according to the term $Z_{i_k}^h \frac{t_k -y_k}{1-y_k}$, for any category $k$, with $k\neq h$.

More specifically, when the term $\frac{(t_k -y_k)}{1-y_k}{=}1$,  and the category $k{\neq} h$ is considered,   $W^h$ moves in the direction opposite to the best representative feature vector $F_{i_k}$. On the other hand, when $t_k=0$  the term considered is $ -(Z_{i_k}^k \frac{y_k}{1-y_k})$ which is added to $W^h$, for its  updating. Note that, in this case,  the update term is increasingly small, since $y_k{\ll}1- y_k$ as $y_k{\to}0$.  
This optimization method, based on iterative learning and stochastic gradient descent, induces a separation in the space of patch features, according to the multilabel classification.

\section{ViT-PCM model  structure}\label{sec:architecture}

The model architecture has two branches, as shown in Figure \ref{fig:network}. We describe its components in the following.

\noindent
{\bf Augmentation.} The batch of input images is augmented as usual in the first branch. In the second branch, images are translated, rotated and scaled. Furthermore, we merge four images from the batch into a single image after scaling them to have a different tiling of the images into patches.

 \noindent
{\bf ViT patch encoder.} The Vision Transformer encoder takes as input the augmented batch of images and returns the features $F_{in}$ and the $n$ patches described in the {\em explicit search} method, Section \ref{sec:explicit_search}.

\noindent
{\bf HV-BiLSTM patch conditioning.} Two bidirectional LSTM (BiLSTM) process row-wise and   column-wise  the features $F_{in}$ transformed to a tensor grid.  The two BiLSTM outputs are concatenated into a HV-BiLSTM (for Horizontal and Vertical), and their feature maps $F$ are fed to the Patch Classifier. The HV-BiLSTM improves information amid neighbour patches by conditioning each patch on all other ones in horizontal (H)  and vertical directions (V) \cite{van2016pixel}.

\noindent
{\bf Patch Classifier (PC).} While ViT and the two BiLSTM encode class information into the patch features, the Patch Classifier  implements the BPM generation, as described in the explicit search method, Section \ref{sec:explicit_search}. 

\noindent
{\bf Two branches for Equivariant regularization.} \label{par:equiv}
 ViT are not equivariant to translations because of the absolute positional encoding used for self-attention. Romero {\em et Al.} \cite{romero2020group} show that for self-attention to be equivariant to group transformations, they must act directly on positional encoding. 
 In our ViT-based method, though GMP is independent of the positional encoding and is invariant to transformations, the BPM generation is not. 
To remedy we resort to typical self-supervised learning tasks, using two branches enabling the network to learn equivariance properties. Equivariance encourages the feature representation to change coherently to the transformation applied to the input \cite{dangovski2021equivariant}. As discussed above, we apply affine transformations to both the network branches in the preprocessing step.  
After the same processing steps of the main branch, the sibling one applies an inverse merging of the features and upscales them to obtain the  $n$ patches feature maps as in the main branch. Finally,  inverse affine transformations are applied to both branches. 

The outcome is that these transformations cope both with positional encoding and spatial transformations.
The loss to be minimized is the cross entropy  loss ${\mathcal L}_{ET}$, taking into account the transformations in the two branches:
 
\begin{equation}
\begin{array}{lll}
    \mathcal{L}_{ET} & {=} & \displaystyle{-\frac{1}{s}\sum_{i=0}^s \sum_{X\in {\mathcal X}} \nu_i(X) \log  \mu_i(X) }\\ 
    & &\text{with }\mu_i(X)= a^{-1} f(a(X)) \mbox{ and } \nu_i(X) = c^{-1}f(c(X))
    \end{array}    
\end{equation}
Here, ${\mathcal X} $ is the images domain,  $a(\cdot), b(\cdot)$ are affine transformations in the first and second branch, $m(\cdot)$ is the above defined merging operation, and $c = m(b(\cdot))$. 

\noindent
{\bf Final loss}
We have the ${\mathcal L}_{MCE}$ loss, conveying the mapping between image classification and patch classification, and  ${\mathcal L}_{ET}$, which ensures equivariance and scales the images so that patches get pixel dimension. The  final loss is then:
\begin{equation}
{\mathcal L} = {\mathcal L}_{MCE} +{\mathcal L}_{ET}
\end{equation}
Training the end-to-end network by minimizing this final loss obtains the baseline pseudo-mask.
\typeout{---------------------Experiments and results -----------------------}
\section{Experiments and results}\label{sec:exp}
\subsection{Set-Up}\label{subsec:impdet}

\begin{figure}[t]
\centering
\begin{subfigure}[t]{.15\linewidth}
\includegraphics[width=\linewidth]{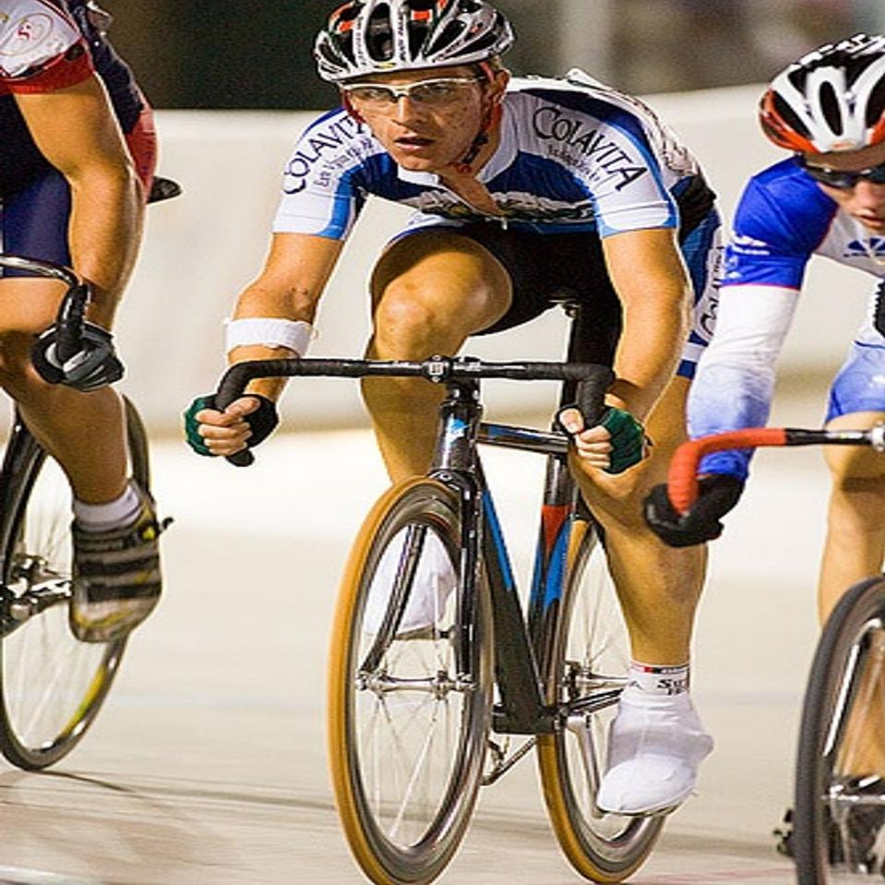}
\end{subfigure}
\begin{subfigure}[t]{.15\linewidth}
\includegraphics[width=\linewidth]{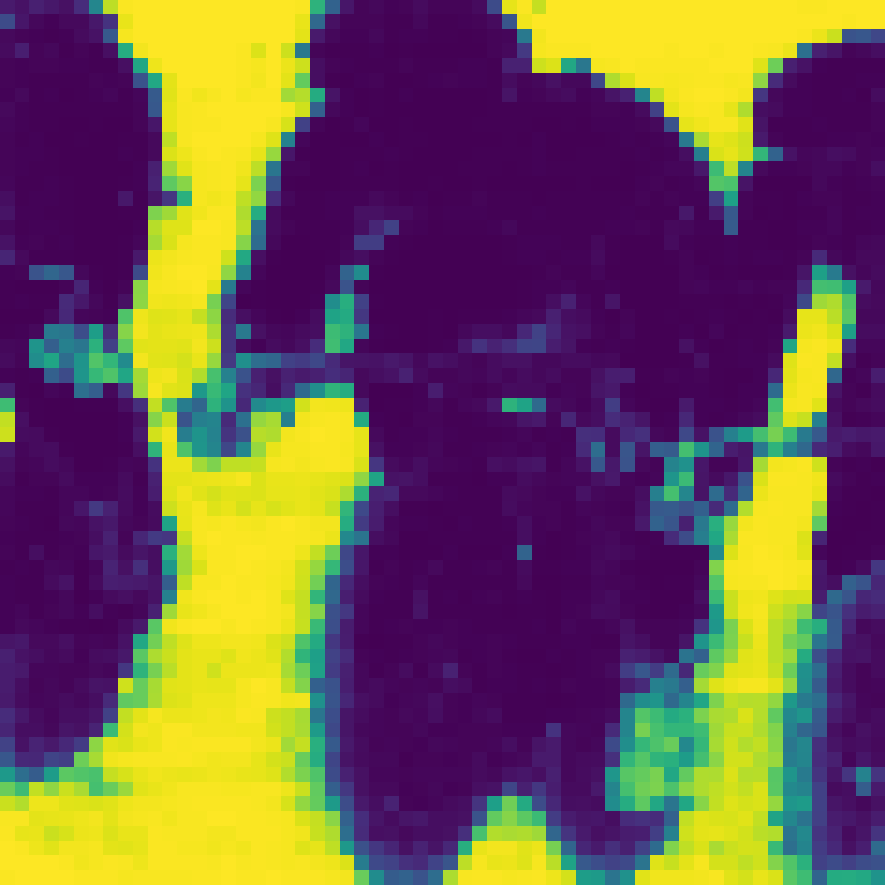}
\end{subfigure}
\begin{subfigure}[t]{.15\linewidth}
\includegraphics[width=\linewidth]{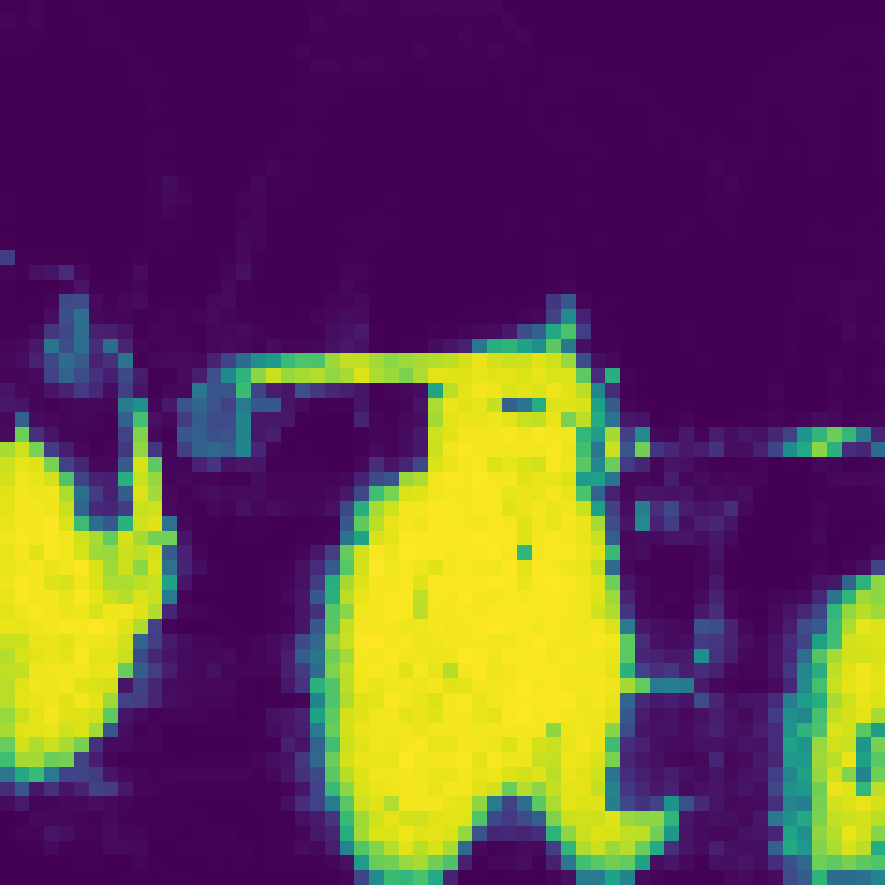}
\end{subfigure}
\begin{subfigure}[t]{.15\linewidth}
\includegraphics[width=\linewidth]{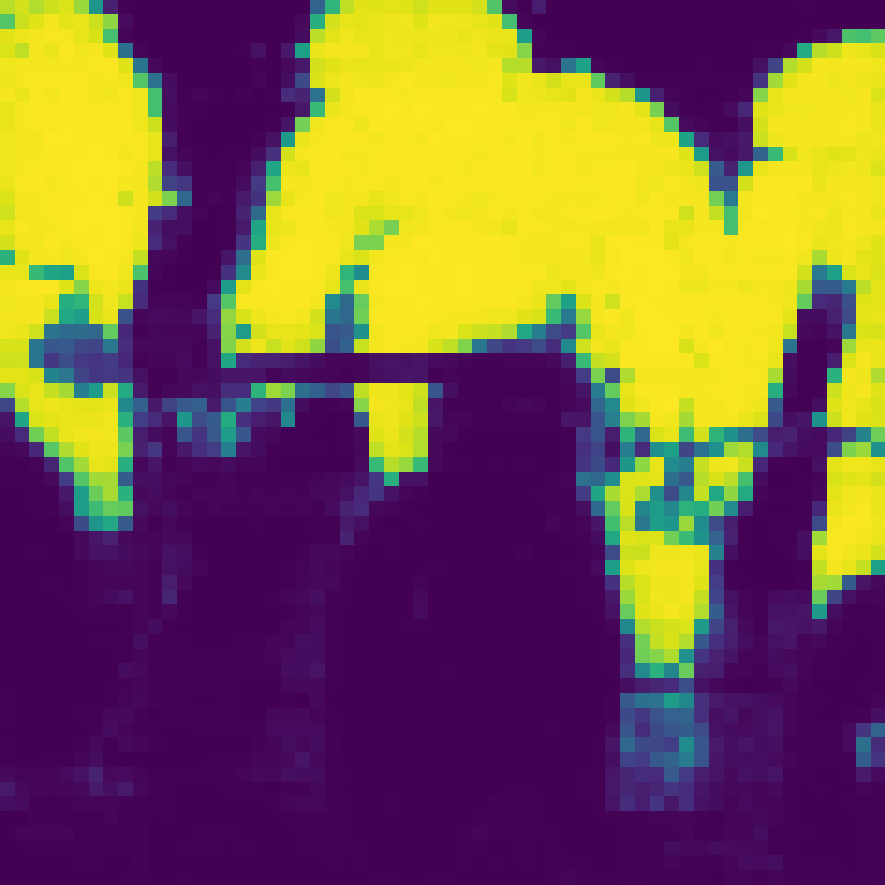}
\end{subfigure}
\begin{subfigure}[t]{.15\linewidth}
\includegraphics[width=\linewidth]{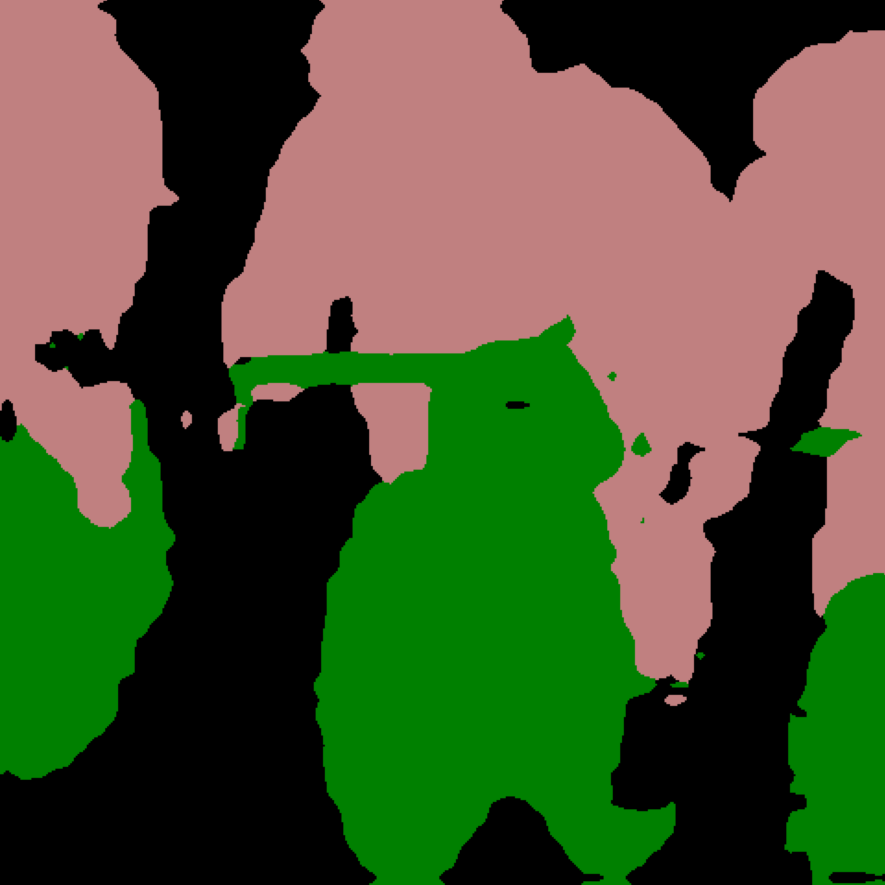}
\end{subfigure}
\begin{subfigure}[t]{.15\linewidth}
\includegraphics[width=\linewidth]{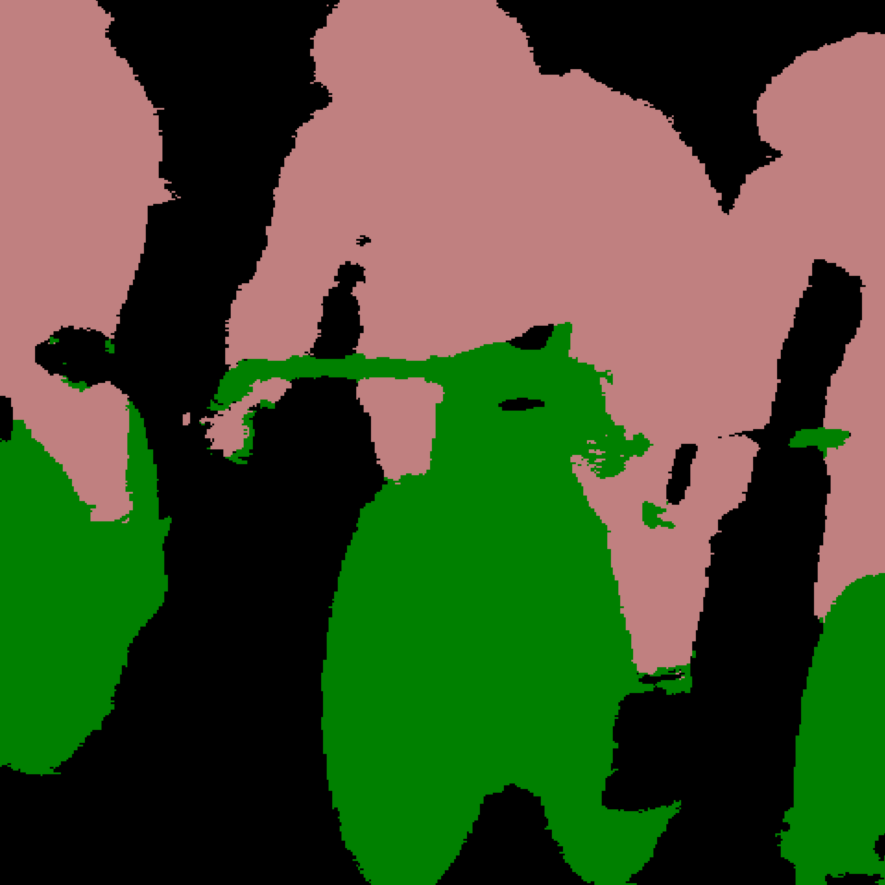}
\end{subfigure}\\
\begin{subfigure}[t]{.15\linewidth}
\includegraphics[width=\linewidth]{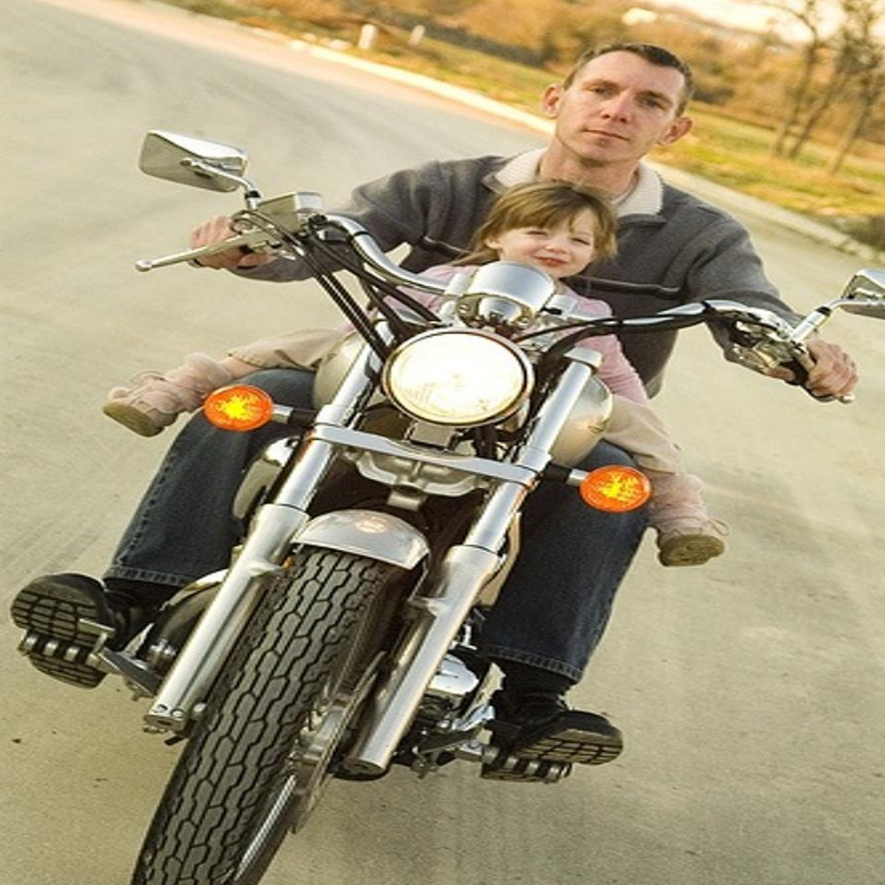}
\end{subfigure}
\begin{subfigure}[t]{.15\linewidth}
\includegraphics[width=\linewidth]{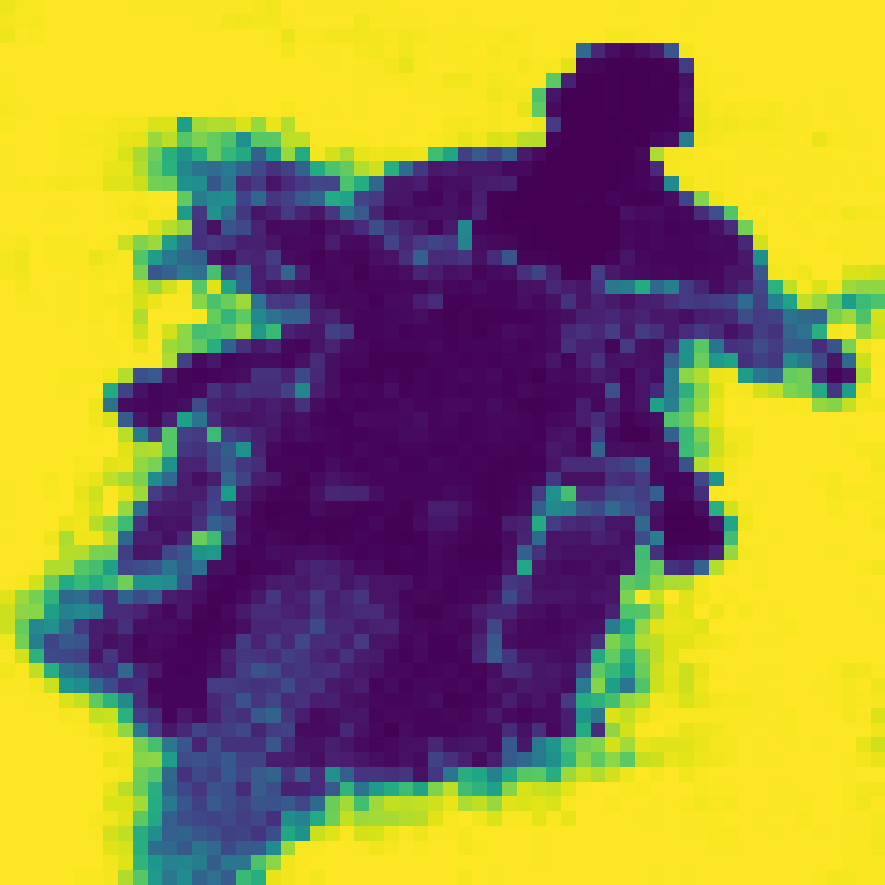}
\end{subfigure}
\begin{subfigure}[t]{.15\linewidth}
\includegraphics[width=\linewidth]{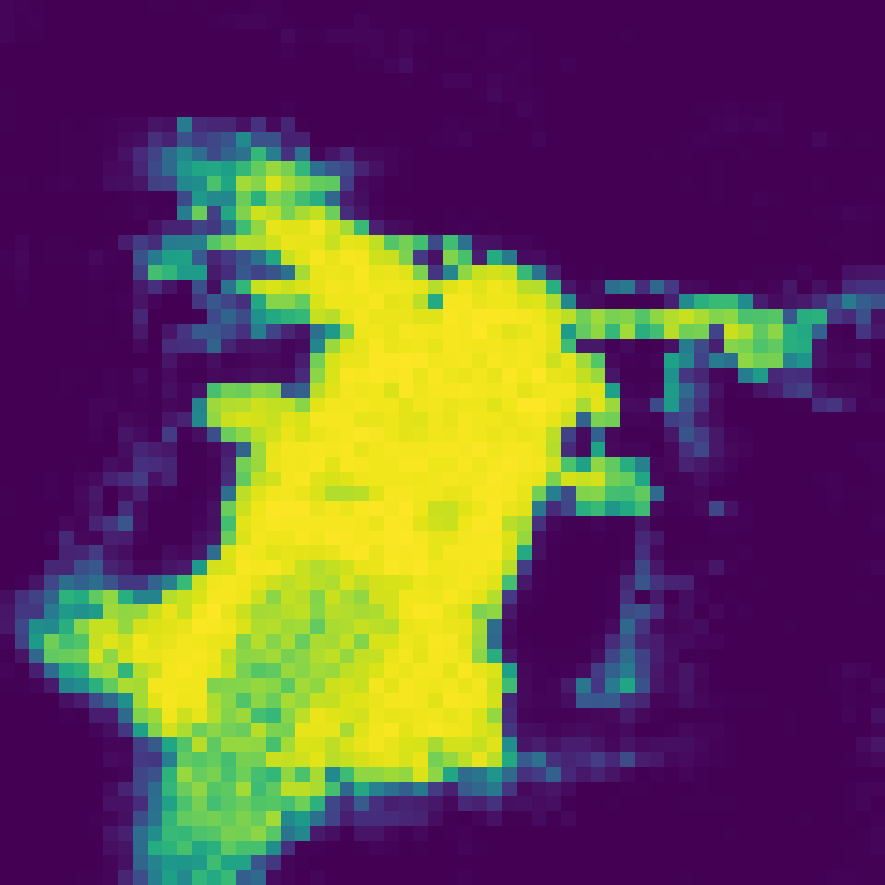}
\end{subfigure}
\begin{subfigure}[t]{.15\linewidth}
\includegraphics[width=\linewidth]{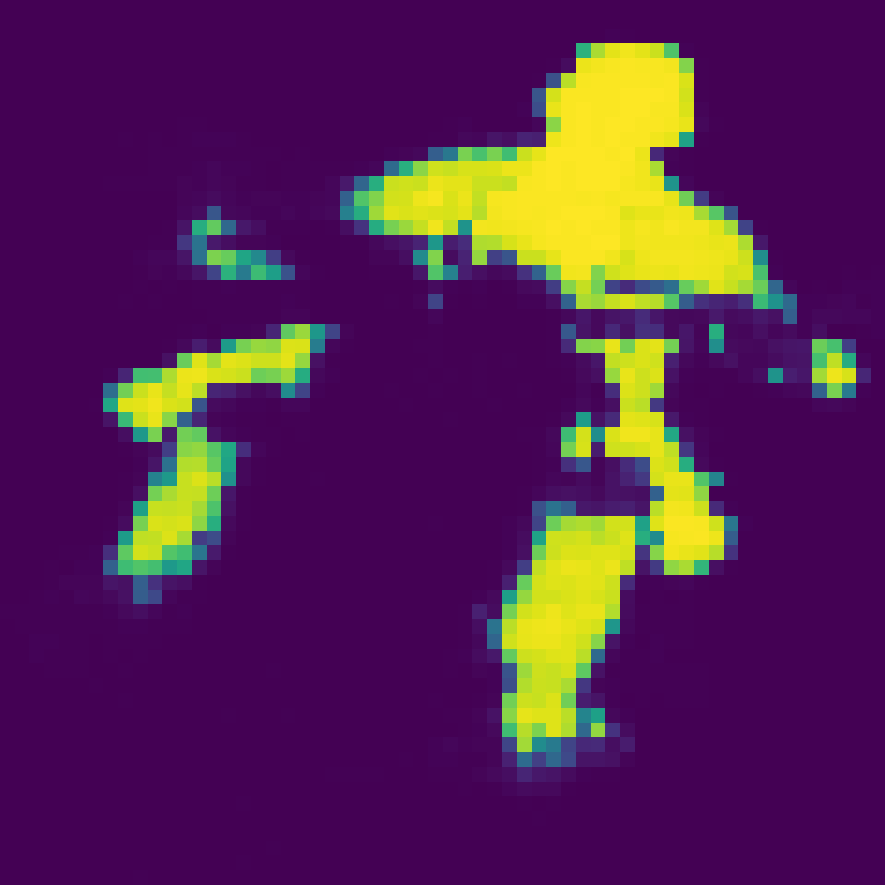}
\end{subfigure}
\begin{subfigure}[t]{.15\linewidth}
\includegraphics[width=\linewidth]{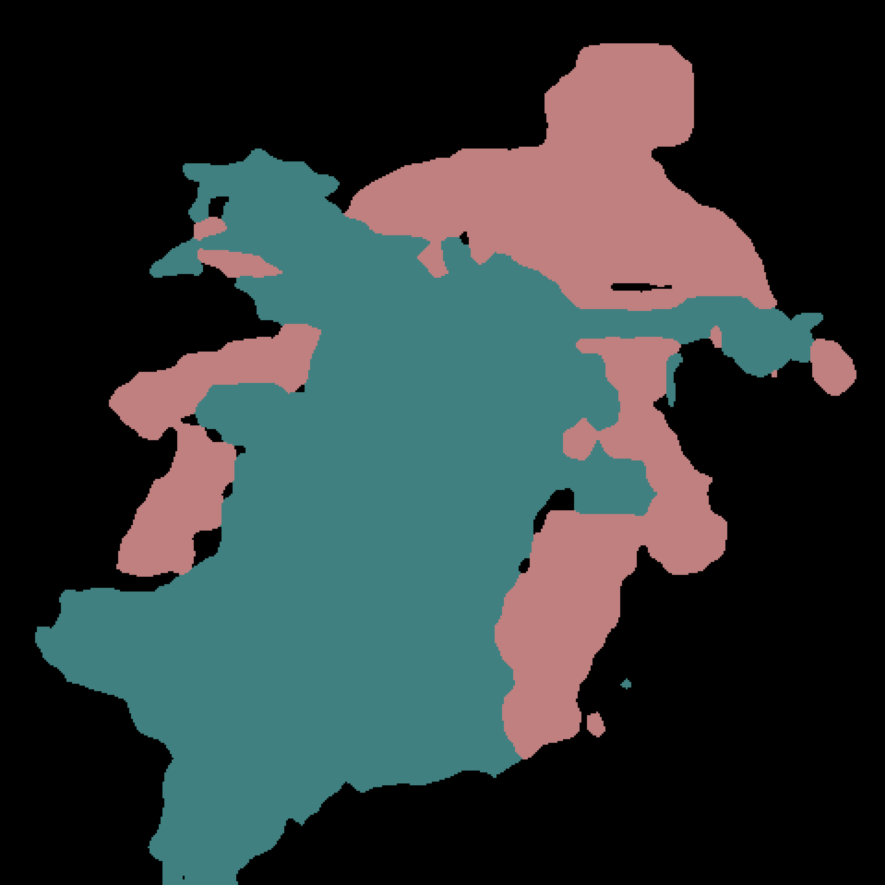}
\end{subfigure}
\begin{subfigure}[t]{.15\linewidth}
\includegraphics[width=\linewidth]{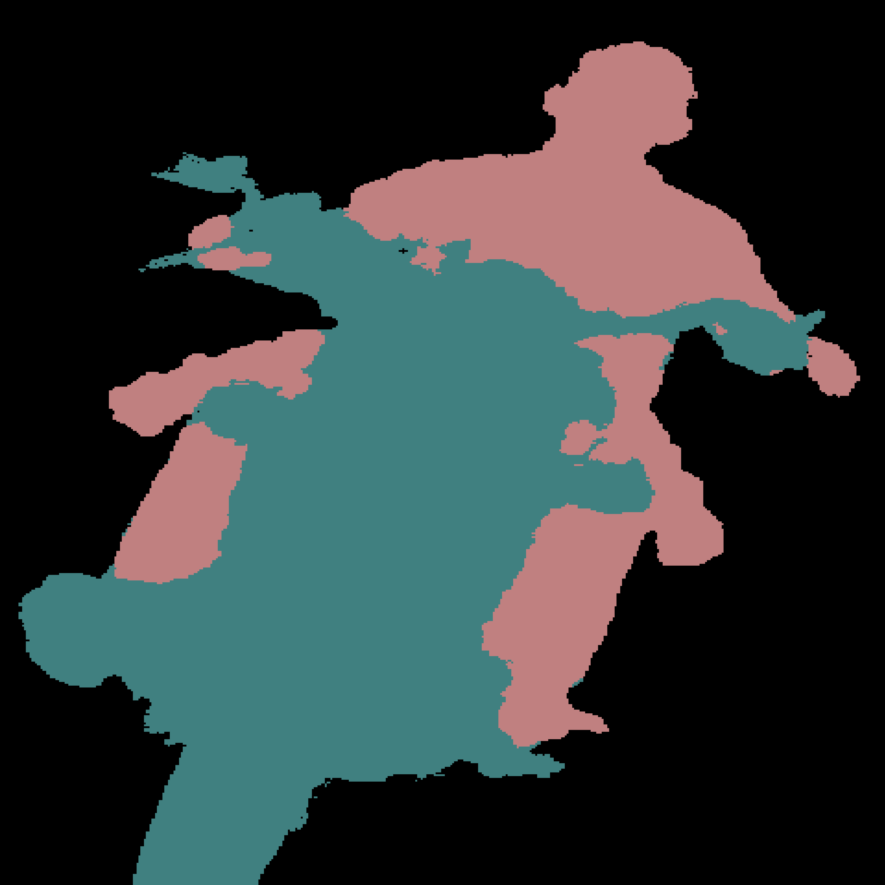}
\end{subfigure}\\
\begin{subfigure}[t]{.15\linewidth}
\includegraphics[width=\linewidth]{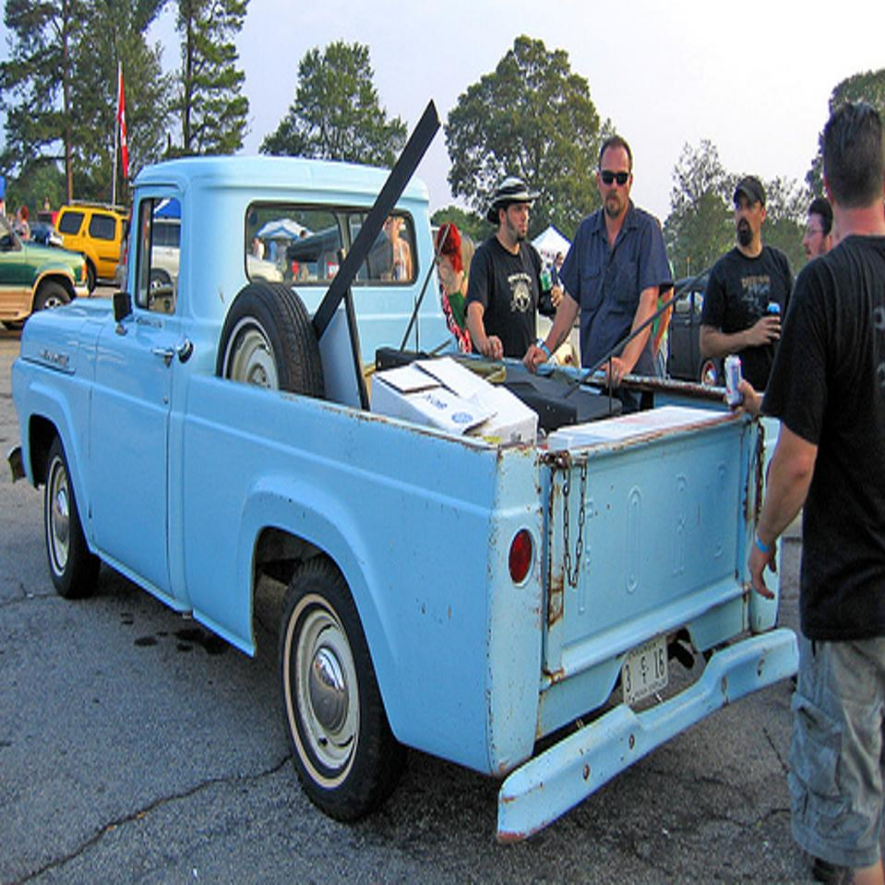}
\end{subfigure}
\begin{subfigure}[t]{.15\linewidth}
\includegraphics[width=\linewidth]{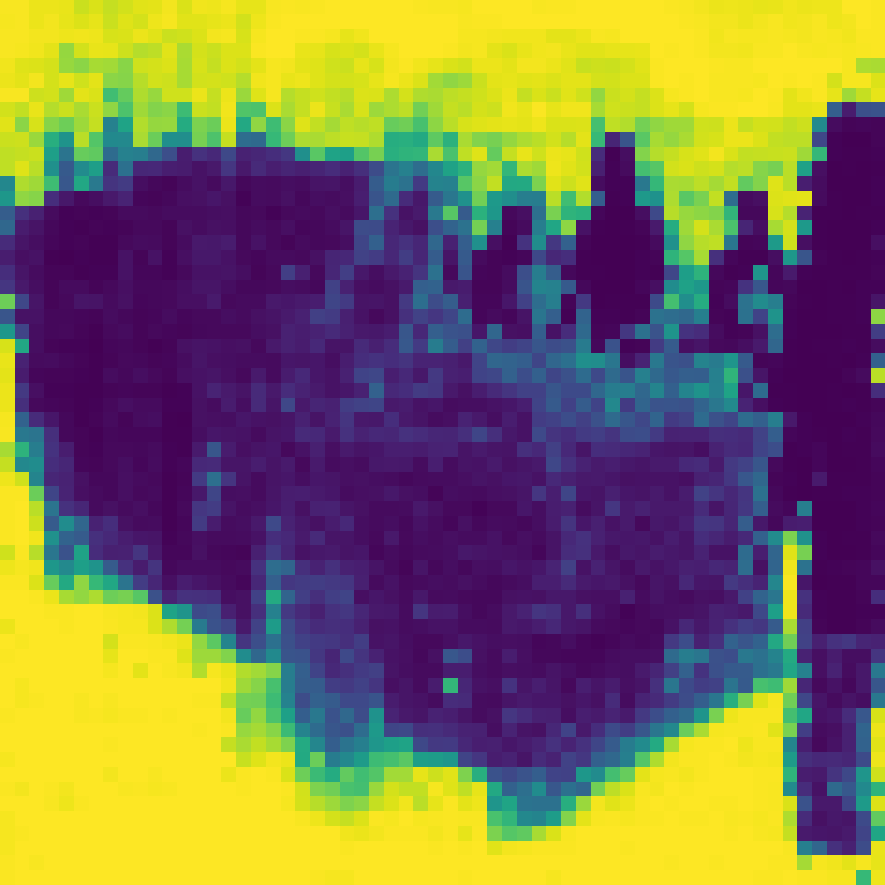}
\end{subfigure}
\begin{subfigure}[t]{.15\linewidth}
\includegraphics[width=\linewidth]{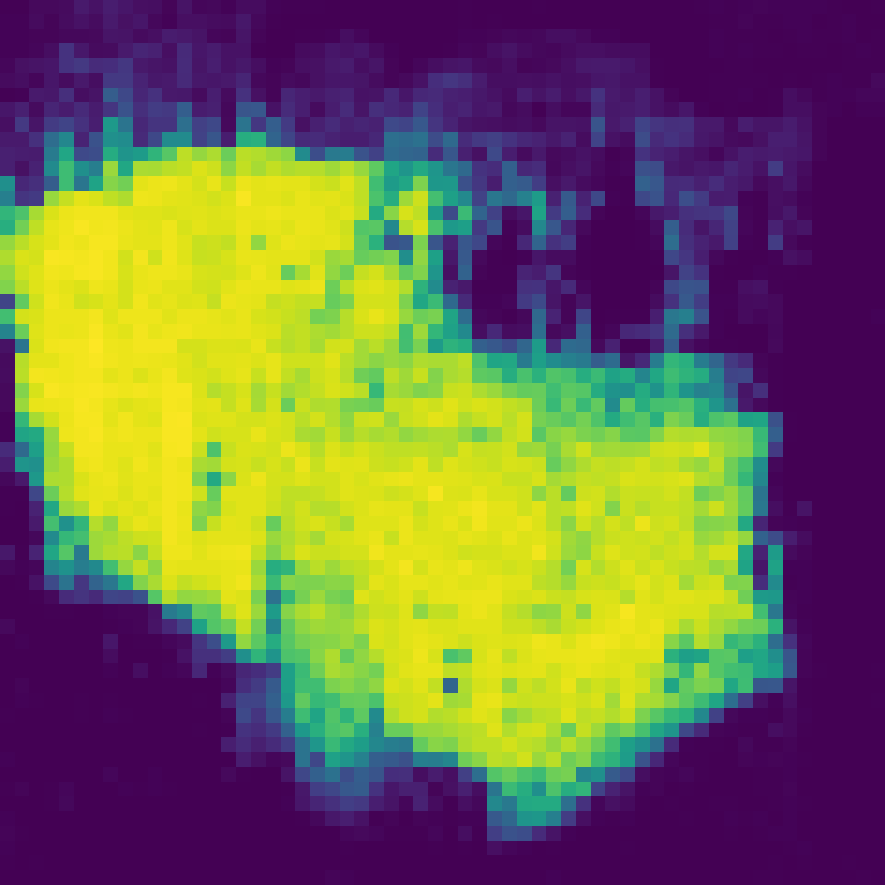}
\end{subfigure}
\begin{subfigure}[t]{.15\linewidth}
\includegraphics[width=\linewidth]{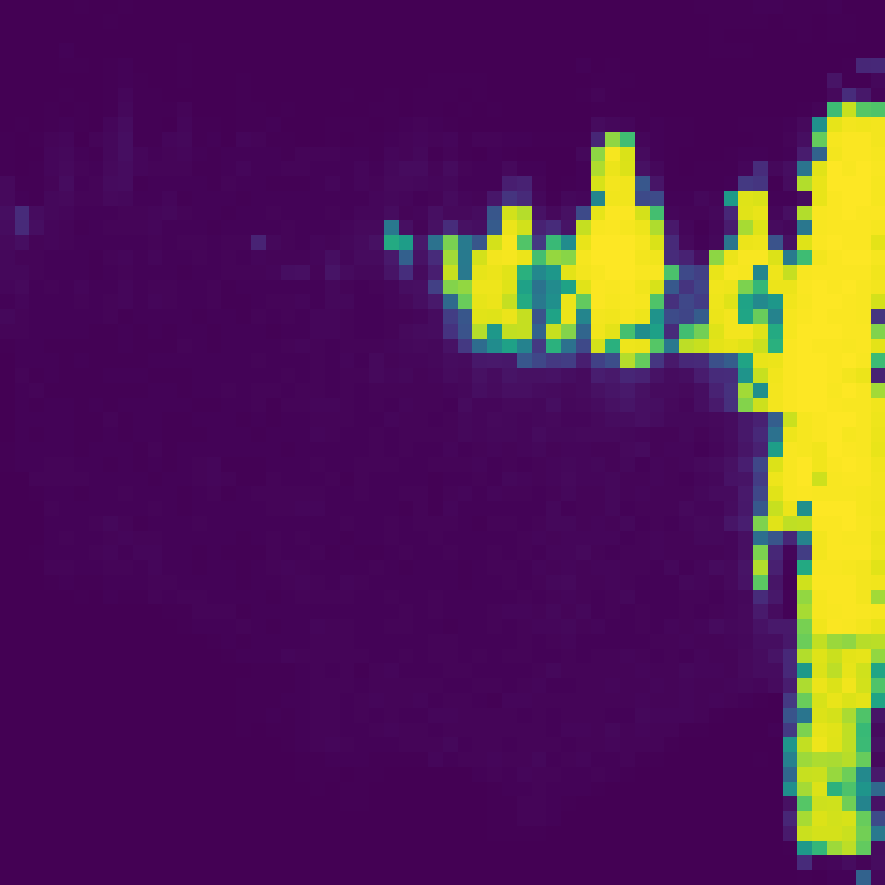}
\end{subfigure}
\begin{subfigure}[t]{.15\linewidth}
\includegraphics[width=\linewidth]{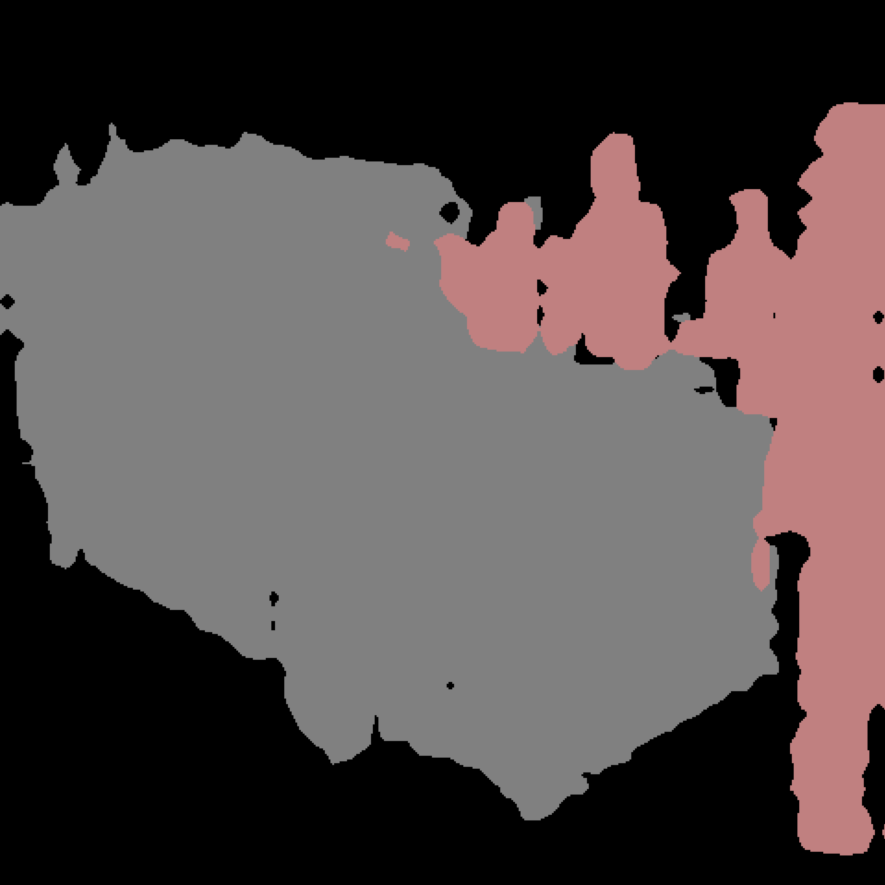}
\end{subfigure}
\begin{subfigure}[t]{.15\linewidth}
\includegraphics[width=\linewidth]{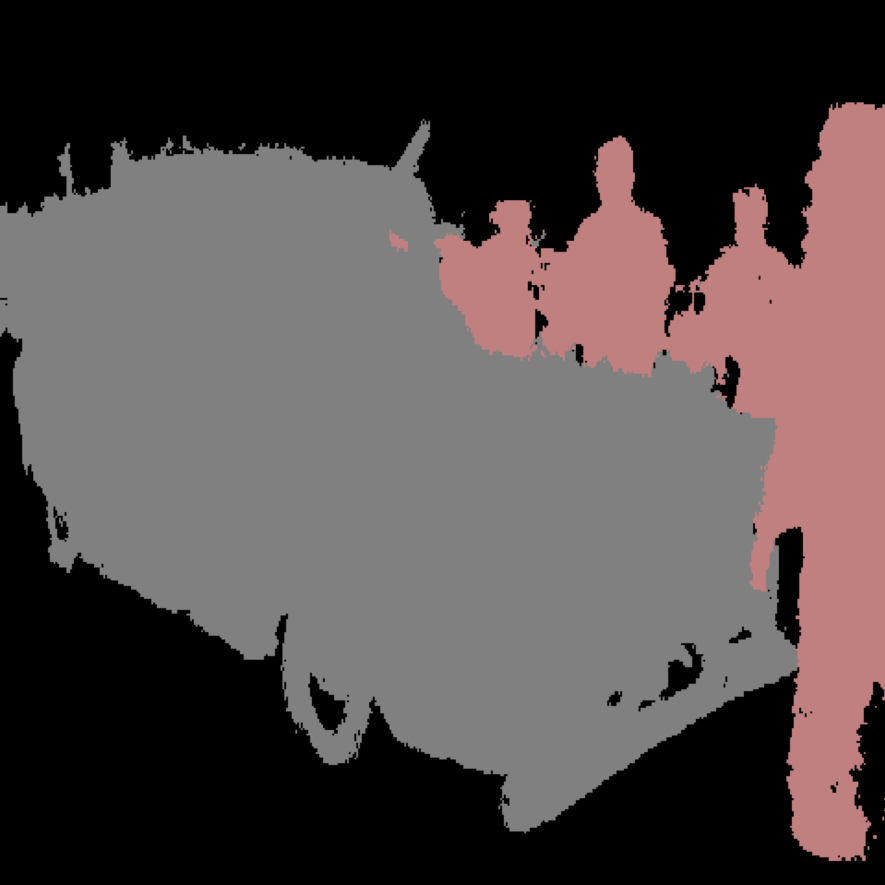}
\end{subfigure}

\includegraphics[width=0.97\linewidth,left]{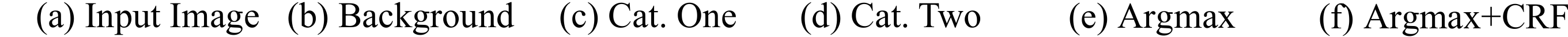}
\caption{Columns (b)-(c)-(d) show the BPM inferred by our ViT-PCM, with probabilities highlighted by $60{\times}60$ heatmaps: values in yellow indicate the pixels' probability of belonging to the predicted class. Column (e)  is the scaled BPM,  obtained by selecting from the distribution of each patch the category indices with maximum probability (argmax). Column  (f) displays the BPM argmax refined by CRF.}
\label{fig:ablatiob_argmax}
\end{figure}

\noindent
{\bf Datasets.}
We conducted our experiments on Pascal VOC 2012 \cite{everingham2010pascal} (20 categories) and on MS COCO 2014\cite{lin2014microsoft} (80 categories), the additional background class is inferred. The Pascal VOC 2012 Dataset \cite{everingham2010pascal} is usually augmented with the SBD dataset \cite{hariharan2011semantic}. The images in train sets of PASCAL VOC and MS COCO are annotated with image-level labels
only. We report mean Intersection-Over-Union
(mIoU) as the evaluation criteria.

\vskip0.3\baselineskip
\noindent
{\bf Networks Configuration.}
For the ViT transformer  backbones \cite{dosovitskiy2021image} we used ViT-S/16 and ViT-B/16 architectures, pre-trained on ImageNet22K  and fine-tuned on ImageNet2012 \cite{russakovsky2015imagenet}. We designed an MLP layer projecting the patch features into a categorical distribution on the $K$ classes as a baseline model for ablation purposes. For the verification task, we used DeepLab V2\cite{chen2018encoder}.\

\noindent
{\bf Reproducibility.}
Images are resized to $384{\times}384$ for training and augmented by random colour jitter, random grayscale,  $90^{\circ}$ rotation, and vertical and horizontal flip. 
Initially, we freeze the backbone and ignore the output feature for the $[cls]$ token. At the same time, we preserve the $24{\cdot} 24$ encoded patch features as input to the BiLSTM conditioning, whose outcome features are passed to the Patch Classifier. 
We initialize the MLP layer with standard Gaussian distribution and use L2 regularization with coefficient $l_2{=}10^{-1}$. 
We ran our training sessions iterating over the entire dataset, each epoch measuring the mIoU(\%) progresses on the PascalVOC 2012 and MS COCO2014 validation sets. We keep the input resolution to $384{\times}384$ to hasten the evaluation on a 4 NVIDIA
Titan V GPUs with 12GB RAM each, a deliberately limited resources setup. We use Adam optimization and schedule the learning rate as follows:  $10^{-3}$ learning rate for the first two epochs with a frozen backbone; then, we unfreeze the last four backbone layers and keep training until convergence with $10^{-4}$ learning rate. At inference time, we scale the input image to $960{\times}960$ to get pseudo-label segmentation maps of shape $60{\times}60$. As expected, we noticed an increase in performance of about $2{-}3\%$ mIoU scores for validation in the training session, confirming that ViTs scales very well on larger input size.

\subsection{Ablation studies}\label{subsec:ablations}
In Table \ref{tab:pseg} we evaluate ViT-PCM computation both with backbone ViT-S/16 and ViT-B/16, considering each component of the end-to-end network. We adopted a patch size of $16$ since the memory requirements grow quadratically with the number of patches. The low scores of the ($\mathcal{L}_{MCE}$) in Table \ref{tab:pseg} are due to the difficulty in encoding the background without equivariance. We observe that with the equivariance, ${\mathcal L}_{ET}$ there is an improvement of 15.2 mIoU\% on the \textit{train} set and 13.1 mIoU\% on the \textit{val} set for PascalVOC 2012. A further improvement of 4.4 on the \textit{train} set and 5.4 mIoU\% on the \textit{val} set is obtained by conditioning the patches with HV-BiLSTM. Finally, we add the dCRF\cite{krahenbuhl2011efficient}  as post-processing obtaining an improvement of 3.5 mIoU\% on \textit{train} set. 
 \noindent
\begin{table}[tbp]
  \centering
  \caption{Ablation on our ViT-PCM model for baseline pseudo-mask production, on PASCAL-VOC 2012 values in mIoU\%.}
   \resizebox{0.85\textwidth}{!}{
    \begin{tabular}{c|c|c|c|c|c|c}
    \hline
{\bf Backbone} & ${\mathcal L}_{MCE}$ &  ${\mathcal L}_{ET}$ & HV-BiLSTM & CRF   & {\bf train} & {\bf val}\\
   \hline
   \hline
     ViT-S/16 & $\checkmark$ &       &       &       &   44.0    & 43.3\\
     \cline{2-7}
  & $\checkmark$ & $\checkmark$ &       &       &   59.2 {\color{ForestGreen} {+}15.2}   & 56.4 {\color{ForestGreen} {+}13.1} \\
\cline{2-7}          
   & $\checkmark$ & $\checkmark$ & $\checkmark$&       & 63.6 {\color{ForestGreen} {+}4.4} & 61.8{\color{ForestGreen} {+}5.4} \\
\cline{2-7}          
 & $\checkmark$ & $\checkmark$ & $\checkmark$ & $\checkmark$ & 67.1 {\color{ForestGreen} {+}3.5} & 64.9{\color{ForestGreen} {+}3.1} \\
\hline
   ViT-B/16 & $\checkmark$ &       &       &       &   45.6    & 44.1\\
     \cline{2-7}
  & $\checkmark$ & $\checkmark$ &       &       &   65.1 {\color{ForestGreen} {+}19.5}   & 62.4 {\color{ForestGreen} {+}18.3} \\
\cline{2-7}          
   & $\checkmark$ & $\checkmark$ & $\checkmark$&       & 67.7 {\color{ForestGreen} {+}2.6} & 66.0{\color{ForestGreen} {+}3.6} \\
\cline{2-7}          
  & $\checkmark$ & $\checkmark$ & $\checkmark$ & $\checkmark$ & 71.4 {\color{ForestGreen} {+}3.7} & 69.3{\color{ForestGreen} {+}3.3} \\
    \hline
    \end{tabular}%
    }
  \label{tab:pseg}%
\end{table}%
\begin{table}[htbp]
\caption{ mIoU(\%) of BPM on  PascalVOC 2012 \textit{val}  set. w/wo CRF}
\label{tab:ablation_categories}
\resizebox{\textwidth}{!}{%
\begin{tabular}{c | c  c  c  c  c  c  c  c  c  c  c  c  c  c  c  c  c  c  c  c  c|  c} 
 \hline
 Method & bkg & plane & bike & bird & boat & btl & bus & car & cat & chair & cow & table & dog & horse & mbk & person & plant & sheep & sofa & train & tv &  mIoU(\%)\\ 
 \hline
 pseudo-masks w/o CRF & 87.2& 66.4 & 36.9 & 61.0 & 61.1 & 63.0 & 86.8 & 76.0 & 76.9 & 41.1 & 80.7 & 39.0 & 82.3 & 77.4 & 75.7 & 55.9 & 50.6 & 85.0 & 50.9.6 & 78.9 & 54.7 & 66.0 \\
 pseudo-masks w/ \ CRF & 88.8 &    78.2 &    39.1 &    69.2 &    67.2 &    67.2 &    88.0 &    77.7 &    78.5 &    42.5 &    83.9 &   39.2 &    85.2 &    82.8 &    79.8 &    56.2 &    51.0 &   91.3 &    51.0 &    81.9 &    57.0 & 69.3 \\
 \hline
\end{tabular}
}
\end{table}

\noindent
Figure \ref{fig:ablatiob_argmax} shows the BPM heat-maps for each class in the second, third and fourth columns, inferred by our end-to-end network, including the background. The BPM heat map highlights each pixel's likelihood of belonging to a specific category. Column (e) shows the pseudo-masks obtained by selecting the indices of the classes with maximum probability. Column (f) shows the pseudo-masks improved by CRF. We use these last masks for the verification task as input to DeepLab \cite{Chen2018DeepLabSI}.
 
\noindent
In Table \ref{tab:ablation_categories} we report the BPM mIoU\%  on Pascal VOC val set for each category,  w and w/o CRF. 
\subsection{Comparisons with state-of-the-art}
{\bf Comparison on baseline pseudo-masks}. We compare the mIoU(\%) accuracy of our ViT-PCM method with other methods, which compute BPM and post-process them with CRF \cite{krahenbuhl2011efficient}  similarly. Some methods such as CIAN \cite{Fan2020CIANCA} and EDAM \cite{wu2021embedded}  also incorporate saliency.

Results are reported in Table {\color{red} A}. Here we can observe that CRF, used as BPM post-processing,  improves the BPM, on average, by 3.97\%, with a standard deviation of 1.87. The statistics show that CRF out of a training loop behaves similarly on all methods. Observe that we improved BPM  state-of-the-art by 3.91 mIou\% points and BPM+CRF by 5.4 mIoU\%, both w.r.t. AFA\cite{ru2022learning}, owning so far the best accuracy on both. 

\noindent
\begin{minipage}[t]{\textwidth}
  \begin{minipage}[b]{0.499\textwidth}\label{tab:BPM}
     \centering
     \resizebox{0.98\textwidth}{!}
     {%
       \begin{tabular}{|c | c | c | c | } 
        \multicolumn{4}{c}{{\bf Table A}: mIoU(\%) on PascalVOC2012 train set.}\\
        \hline
        \rowcolor[gray]{.85}
         Method &  Backbone & BPM             & BPM{+}CRF \\ 
         \hline\hline
             ICD \cite{Fan_2020_CVPR}\tiny{\textsc{CVPR'20} } &  VGG16 &  57.00 &   62.20 \\
             SCE\cite{chang2020weakly}\tiny{\textsc{CVPR'20}} &  ResNet38 &  50.90 & -  \\
             SEAM \cite{wang2020self}\tiny{\textsc{CVPR'20} }&  ResNet38 & 55.41 & 56.83  \\
             CIAN\cite{Fan2020CIANCA}\tiny{\textsc{AAAI'20}} & ResNet101 &  58.10 & 62.50 \\ 
             ECSNet\cite{sun2021ecs}\tiny{\textsc{ICCV'20}} & ResNet38 &  56.60 & 58.60 \\
             PAMR\cite{araslanov2020single}\tiny{\textsc{CVPR'20} } &ResNet38 & 59.7  & 62.7\\
             AdvCAM\cite{lee2021anti}\tiny{\textsc{CVPR'21} } & ResNet50 & 55.60 &  62.10 \\ 
             CPN\cite{zhang2021complementary}\tiny{\textsc{ICCV'21} } & ResNet38 &  57.43 & - \\
             CSE\cite{kweon2021unlocking}\tiny{\textsc{ICCV'21} } & ResNet38 &56.0 & 62.8\\
             EDAM\cite{wu2021embedded}\tiny{\textsc{CVPR'21} } &  ResNet101 &  52.83 &  58.18 \\
             MCTformer\cite{xu2022multi}\tiny{\textsc{CVPR'22} } &  DeiT-S &  61.70 &  - \\
             PPC\cite{du2022weakly}\tiny{\textsc{CVPR'22} } & Resnet38 &  61.50 &  64.00\\
             CLIMS\cite{xie2022clims}\tiny{\textsc{CVPR'22} } & Resnet50 &  56.60 &  -\\
             SIPE\cite{chen2022self}\tiny{\textsc{CVPR'22} } & Resnet50 &  58.60 & 64.70\\
             AFA\cite{ru2022learning}\tiny{\textsc{CVPR'22} } & MiT-B1 &  63.80 & 66.00\\
             IRN+W-OoD\cite{lee2022weakly}\tiny{\textsc{CVPR'22} } & Resnet50&  53.30 & 58.40\\
       \bf{ViT-PCM Ours} &ViT-B/16 &  \textbf{67.71} &  {\bf 71.4} \\
     \hline
         \multicolumn{4}{c}{{\bf Table C}: mIoU(\%) on MS-COCO 2014 val  set.}\\
         \hline 
         \rowcolor[gray]{.85}
         Method &  Backbone &  \multicolumn{2}{c|}{Val}    \\ 
         \hline
         \hline
         MCTformer\cite{xu2022multi}\tiny{\textsc{CVPR'22} } &  Resnet38 &  \multicolumn{2}{c|}{42.0} \\
          SIPE\cite{chen2022self}\tiny{\textsc{CVPR'22} } & Resnet38 &  \multicolumn{2}{c|}{43.6} \\
          {\bf ViT-PCM Ours}  & ViT-B/16 & \multicolumn{2}{c|}{\bf 45.0}\\
          \hline
     \end{tabular}
    
     }%
\end{minipage}
\hfill
 \begin{minipage}[b]{0.495\textwidth}\label{tab:Deeplab}
   \centering
         \resizebox{0.99\textwidth}{!}
         {%
         \begin{tabular}{|c | c | c | c | }
          \multicolumn{4}{c}{{\bf Table B}: mIoU(\%) on PascalVOC2012 val and test set.}\\
              \hline
              \rowcolor[gray]{.85}
              Method &  Backbone &      Val & Test \\ 
             \hline\hline
             IRNet\cite{ahn2019weakly}\tiny{\textsc{CVPR'19} } & ResNet50 & 63.5 & 64.8\\
             SCE\cite{chang2020weakly}\tiny{\textsc{CVPR'20}} & ReseNet101 & 66.1 & 65.9\\
             SEAM\cite{wang2020self}\tiny{\textsc{CVPR'20} } &  ResNet38 & 64.5 & 65.7  \\
             CIAN\cite{Fan2020CIANCA}\tiny{\textsc{AAAI'20}} & ResNet101 &  64.3 & 65.3 \\ 
             ECSNet\cite{sun2021ecs}\tiny{\textsc{ICCV'20}} & ResNet38 &  66.6 & 67.6 \\
             CONTA\cite{dong_2020_conta}\tiny{\textsc{Nuerips'20}} &ResNet101 &66.1 & 66.7\\
             BES\cite{chen2020weakly}\tiny{\textsc{ECCV'20}} &ResNet101 &65.7 & 66.6\\
             AdvCAM\cite{lee2021anti}\tiny{\textsc{CVPR'21} } & ResNet50 & 68.1 &  68.0 \\ 
             CPN\cite{zhang2021complementary}\tiny{\textsc{ICCV'21} } & ResNet38 &  67.8 & 68.5 \\
             EDAM\cite{wu2021embedded}\tiny{\textsc{CVPR'21} } &  ResNet101 &  52.83 &  58.18 \\
             CSE\cite{kweon2021unlocking}\tiny{\textsc{ICCV'21} } & ResNet38 &68.4 & 68.2\\
             MCTformer\cite{xu2022multi}\tiny{\textsc{CVPR'22} } &  Resnet38 &  71.9 & 71.6 \\
             CLIMS\cite{xie2022clims}\tiny{\textsc{CVPR'22} } & Resnet50 &  70.4 &  70.0\\
             SIPE\cite{chen2022self}\tiny{\textsc{CVPR'22} } & Resnet101 &  68.8 & 69.7\\
             AdvCAM+W-OoD\cite{lee2022weakly}\tiny{\textsc{CVPR'22} } & Resnet38 &  70.7 & 70.1\\
            \hline
               \rowcolor{gray!10}
            PAMR\cite{araslanov2020single}\tiny{\textsc{CVPR'20} } &ResNet38 & 62.7  & 64.3\\
              \rowcolor{gray!18}
             MCIS\cite{sun2020mining}\tiny{\textsc{ECCV'20} }& ResNet101 &66.2 & 66.9\\
               \rowcolor{gray!10}
            ICD \cite{Fan_2020_CVPR}\tiny{\textsc{CVPR'20} } &  Resnet101 &  64.1 &   64.3 \\
              \rowcolor{gray!18}
            AFA\cite{ru2022learning}\tiny{\textsc{CVPR'22} } & MiT-B1 &  66.0 & 66.3\\
              \rowcolor{gray!10}
             MCTformer$^{\star}$\cite{xu2022multi}\tiny{\textsc{CVPR'22} } &  Resnet38 &  68.2 & 68.4 \\  
               \rowcolor{gray!18}
             {\bf ViT-PCM Ours} & ResNet 101 &  {\bf 70.3} &  {\bf 70.9} \\
         \hline
         \multicolumn{4}{c}{{\bf Table D}: mIoU(\%) on PascalVOC2012 val set.}\\
         \hline
       \rowcolor[gray]{.85}
         Method &  ViT-S/8 & ViT-S/16 & ViT-B/16\\ 
         \hline\hline
         DINO\cite{caron2021emerging}\tiny{\textsc{ICCV'21} } & 44.7 & 45.9 & -\\
         {\bf ViT-PCM Ours} & - & {\bf 74.55} &  {\bf 77.25}\\
         \hline
         

        \end{tabular}
        }%
  \end{minipage}
\end{minipage}\\
 \noindent
{\bf Semantic Segmentation Verification Tasks} The verification task of the WSSS methods on PascalVOC 2012 tests the final pseudo-mask (FPM), and the results are reported in Table {\color{red}B}. We divide the methods into two:   those which are boosted (or, according to the definition in PAMR \cite{araslanov2020single} are multi-stage) and those which are end-to-end, highlighted in grey. For the methods considered, the boosted ones improve the mIoU\% w.r.t. the BPM  on average of 9.8\%, while the end-to-end methods improve on average 4.2\%. Our ViT-PCM not being boosted improves by 2.39\% on the val set and decreases on the test set. Our ViT-PCM has the best accuracy among the end-to-end methods, with 70.3\% and 70.9\% on \textit{val} and \textit{test} sets.  Our method   is second to MCT-Former\cite{xu2022multi} on the test set w.r.t. all methods (boosted and end-to-end). However, MCT-Former end-to-end version is second to ViT-PCM, on both the val and test sets.

\noindent 
In Table {\color{red}C} we also evaluate our method on MS-COCO 2014 dataset \cite{lin2014microsoft}. Our ViT-PCM achieves 45.03 mIoU\% on \textit{val} set. We reported only the last methods (2022) with the highest performance.
\noindent
Table {\color{red}D} compares our foreground maps with DINO \cite{caron2021emerging} maps on the PascalVOC 2012 val set.
\noindent
\begin{figure}
\begin{minipage}[b]{0.49\textwidth}
\centering
 \includegraphics[width=0.9\linewidth]{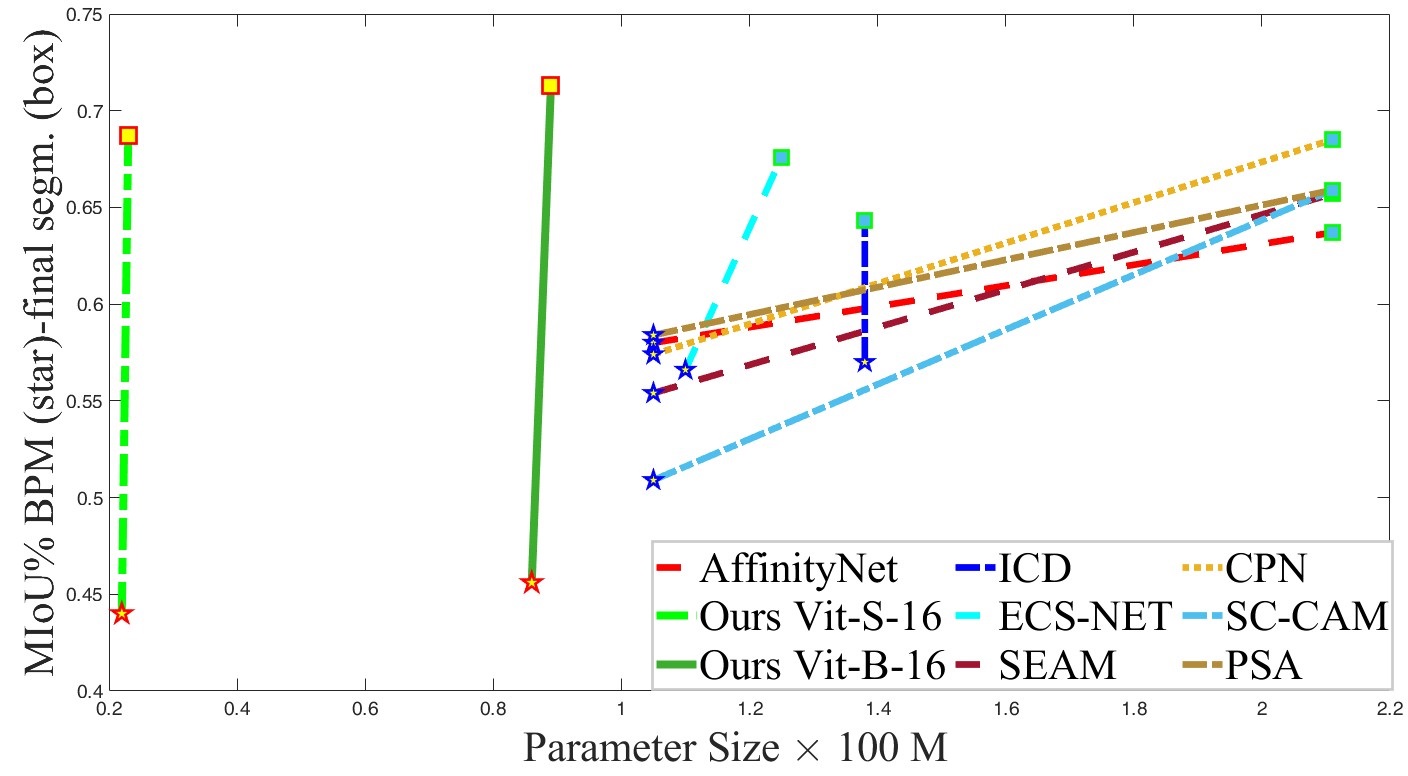}
\caption{Networks parameters consumed from the BPM to the final-segmentation in ours and other methods, against mIoU\% on PascalVoc2012 val. set. }
\label{fig:cost}
\end{minipage}%
\hfill%
\begin{minipage}[b]{0.49\textwidth}
\centering
\resizebox{0.99\textwidth}{!}{%
   \begin{tabular}{c|c|c|c|c}
{\bf Backbone} & Params (M) & Localization &  mIoU (\%)   & pixAcc (\%)\\
   \hline
   \hline
     Resnet50v2    &   25 &   CAM        & \textbf{27.8}  & 72.7\\
          &   &   PCM     & 25.2  & \textbf{76.0}\\
     \cline{1-5}
     Xception    &   23 &   CAM        & \textbf{37.8}  & 76.5\\
         &   & PCM    & 36.5  & \textbf{79.5}\\
     \cline{1-5}
     ViT-S/16  &   \textbf{22} &   CAM        & 29.3  & 55.0\\
         &   & PCM        & \textbf{43.3}  & \textbf{80.1}\\
    \hline
    \end{tabular}%
    }
\tabcaption{Comparison between CAM \cite{zhou2016learning} and PCM (our Patch Class Mapping) on PascalVOC2012 \textit{val} set. The Table reports the best results obtained with Multi-Label BCE loss and L2 regularization loss in all experiments, for both CAM and PCM.}
\label{table:CAM-PCM}
\end{minipage}%
\end{figure}
\noindent
Figure \ref{fig:cost} shows the ratio between the parameters consumed to obtain the BPM and the final segmentation mask, against the mIoU\% on the val set of PascalVOC2012. 
A $\star$ marker specifies the BPM, and a $\Box$ marker specifies the final segmentation mask, ours in red and the others in blue. Our ViT-PCM, with backbone ViT-S/16,  is green-dashed, and ViT-B/16 is green-continuous. We can observe that most of the shown methods are multi-stage (see also \cite{araslanov2020single,ru2022learning}), and boosting the BPM asks for a significant increase of parameters. 
\noindent
Table \ref{table:CAM-PCM} shows the accuracy between CAM and PCM on different backbones and the amount of parameters required. We made this table to understand whether it would be profitable to use CAM with ViT. As shown in the table, we can see that the combination ViT and PCM is the best solution.

\begin{figure}[t]
\centering
\begin{subfigure}[b]{.11\linewidth}
\includegraphics[width=\linewidth]{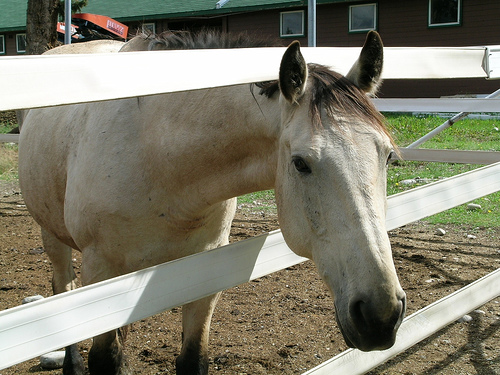}
\end{subfigure}
\begin{subfigure}[b]{.11\linewidth}
\includegraphics[width=\linewidth]{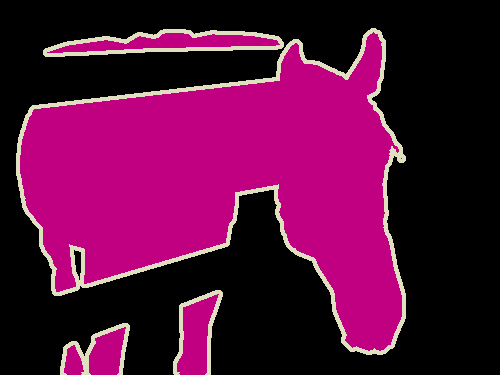}
\end{subfigure}
\begin{subfigure}[b]{.11\linewidth}
\includegraphics[width=\linewidth]{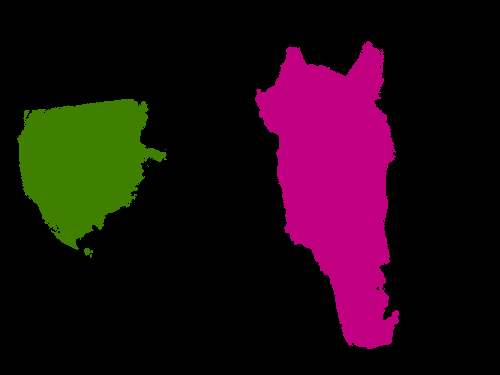}
\end{subfigure}
\begin{subfigure}[b]{.11\linewidth}
\includegraphics[width=\linewidth]{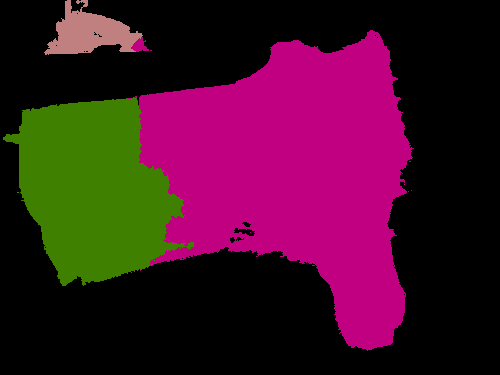}
\end{subfigure}
\begin{subfigure}[b]{.11\linewidth}
\includegraphics[width=\linewidth]{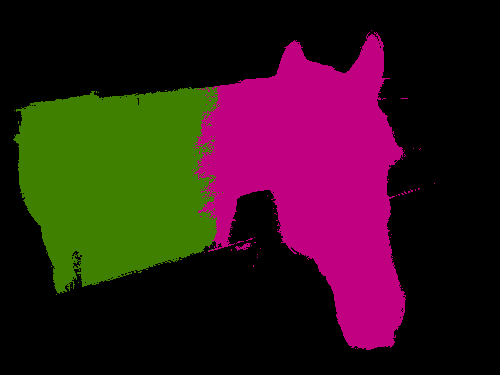}
\end{subfigure}
\begin{subfigure}[b]{.11\linewidth}
\includegraphics[width=\linewidth]{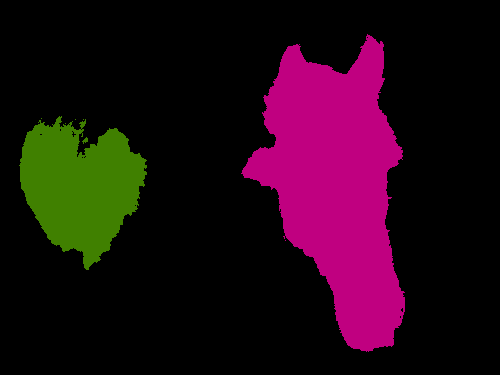}
\end{subfigure}
\begin{subfigure}[b]{.11\linewidth}
\includegraphics[width=\linewidth]{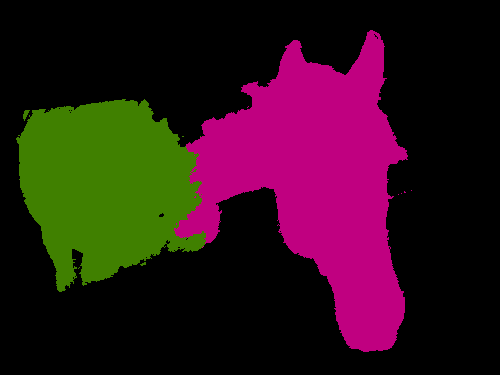}
\end{subfigure}
\begin{subfigure}[b]{.11\linewidth}
\includegraphics[width=\linewidth]{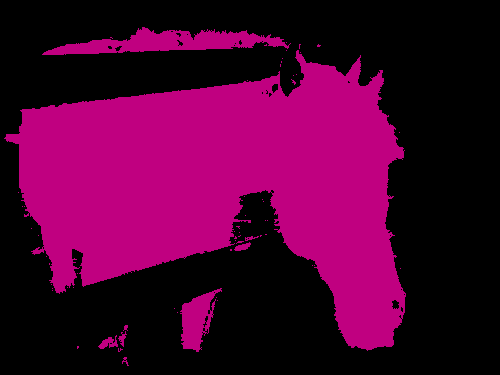}
\end{subfigure}
\begin{subfigure}[b]{.11\linewidth}
\includegraphics[width=\linewidth]{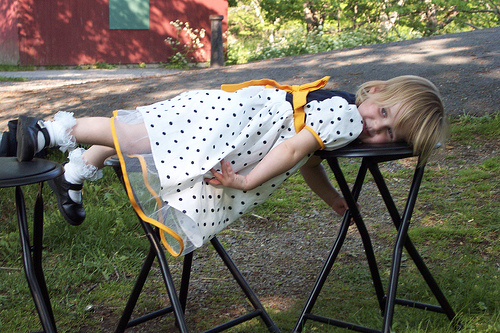}
\end{subfigure}
\begin{subfigure}[b]{.11\linewidth}
\includegraphics[width=\linewidth]{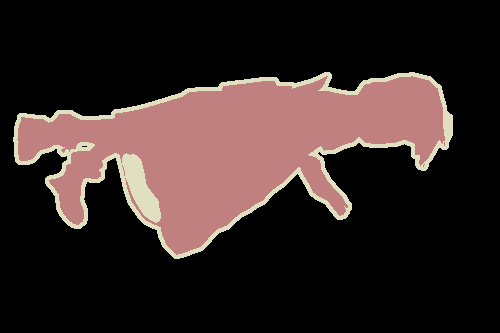}
\end{subfigure}
\begin{subfigure}[b]{.11\linewidth}
\includegraphics[width=\linewidth]{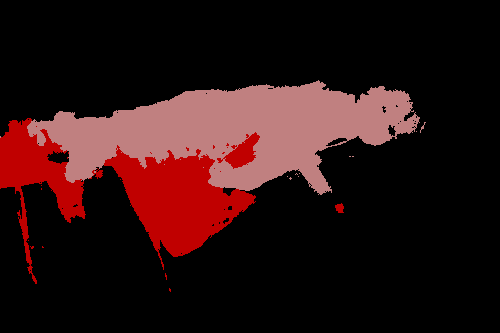}
\end{subfigure}
\begin{subfigure}[b]{.11\linewidth}
\includegraphics[width=\linewidth]{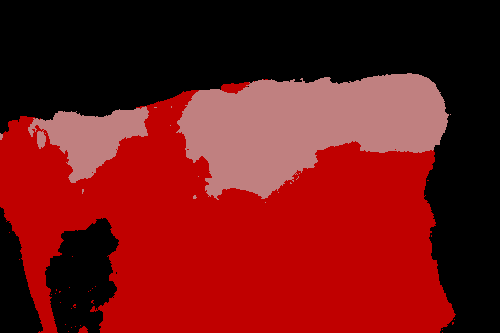}
\end{subfigure}
\begin{subfigure}[b]{.11\linewidth}
\includegraphics[width=\linewidth]{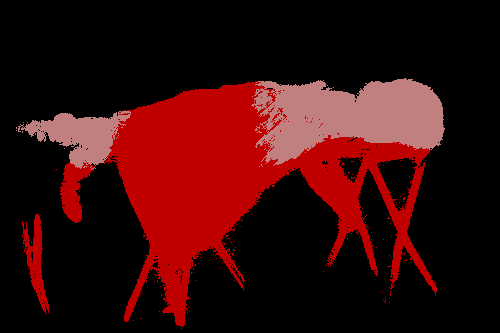}
\end{subfigure}
\begin{subfigure}[b]{.11\linewidth}
\includegraphics[width=\linewidth]{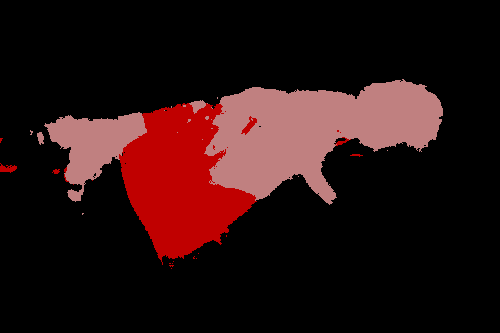}
\end{subfigure}
\begin{subfigure}[b]{.11\linewidth}
\includegraphics[width=\linewidth]{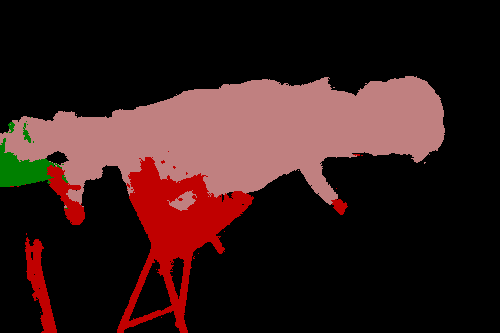}
\end{subfigure}
\begin{subfigure}[b]{.11\linewidth}
\includegraphics[width=\linewidth]{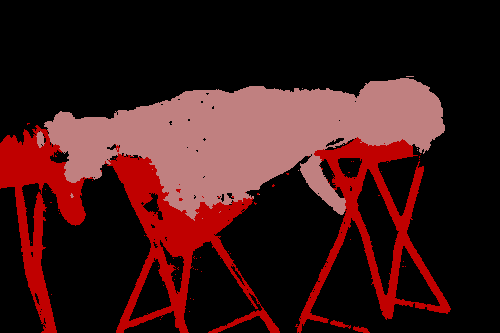}
\end{subfigure}
\begin{subfigure}[b]{.11\linewidth}
\includegraphics[width=\linewidth]{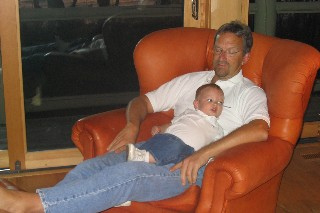}
\end{subfigure}
\begin{subfigure}[b]{.11\linewidth}
\includegraphics[width=\linewidth]{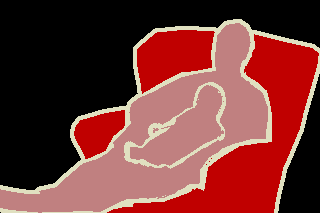}
\end{subfigure}
\begin{subfigure}[b]{.11\linewidth}
\includegraphics[width=\linewidth]{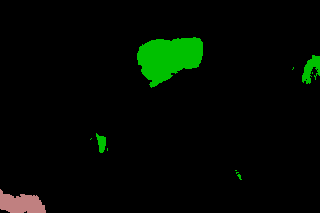}
\end{subfigure}
\begin{subfigure}[b]{.11\linewidth}
\includegraphics[width=\linewidth]{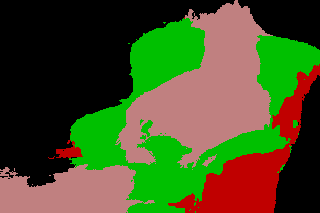}
\end{subfigure}
\begin{subfigure}[b]{.11\linewidth}
\includegraphics[width=\linewidth]{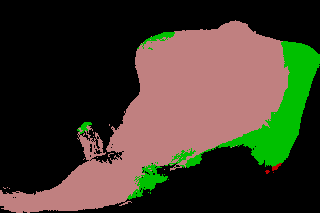}
\end{subfigure}
\begin{subfigure}[b]{.11\linewidth}
\includegraphics[width=\linewidth]{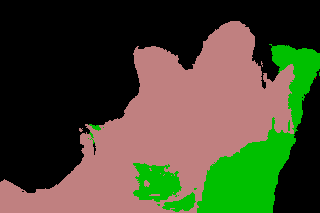}
\end{subfigure}
\begin{subfigure}[b]{.11\linewidth}
\includegraphics[width=\linewidth]{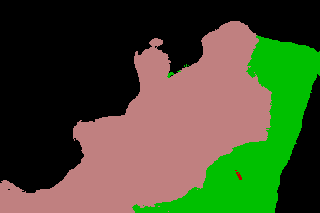}
\end{subfigure}
\begin{subfigure}[b]{.11\linewidth}
\includegraphics[width=\linewidth]{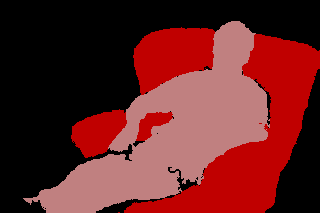}
\end{subfigure}
\begin{subfigure}[b]{.11\linewidth}
\includegraphics[width=\linewidth]{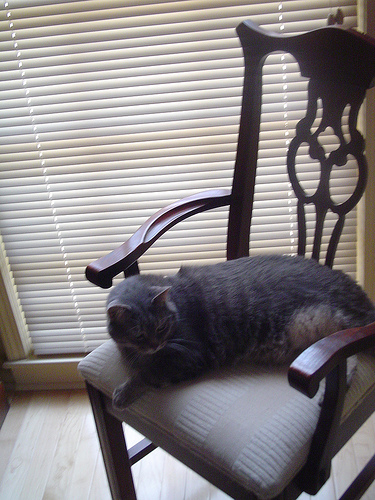}
\end{subfigure}
\begin{subfigure}[b]{.11\linewidth}
\includegraphics[width=\linewidth]{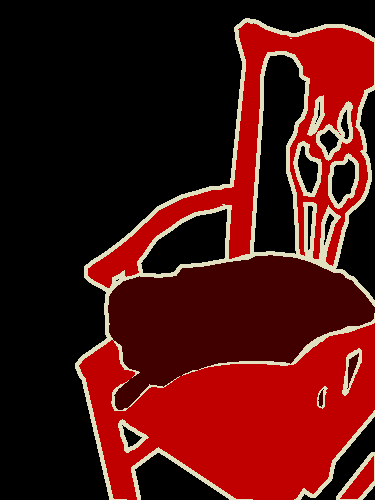}
\end{subfigure}
\begin{subfigure}[b]{.11\linewidth}
\includegraphics[width=\linewidth]{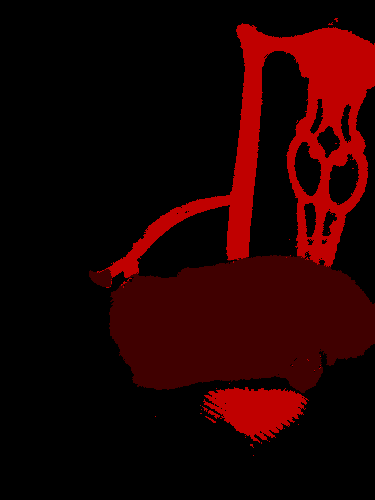}
\end{subfigure}
\begin{subfigure}[b]{.11\linewidth}
\includegraphics[width=\linewidth]{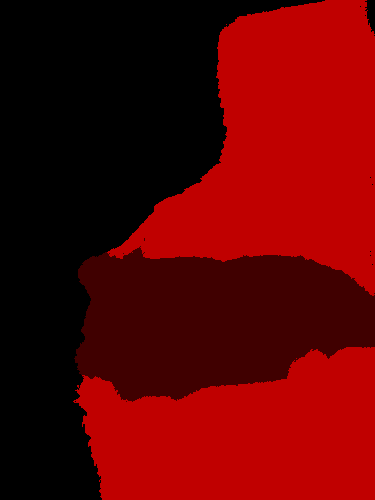}
\end{subfigure}
\begin{subfigure}[b]{.11\linewidth}
\includegraphics[width=\linewidth]{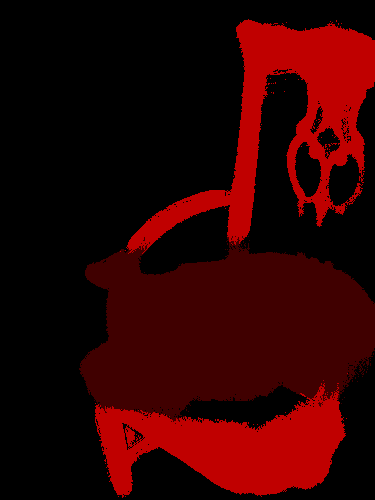}
\end{subfigure}
\begin{subfigure}[b]{.11\linewidth}
\includegraphics[width=\linewidth]{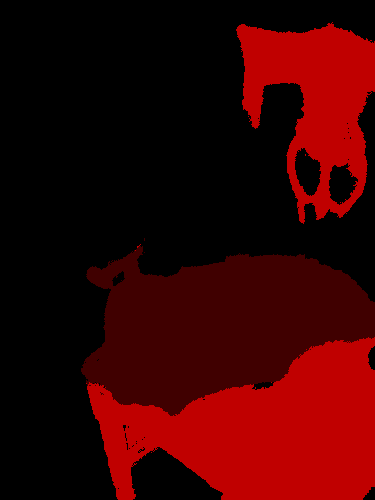}
\end{subfigure}
\begin{subfigure}[b]{.11\linewidth}
\includegraphics[width=\linewidth]{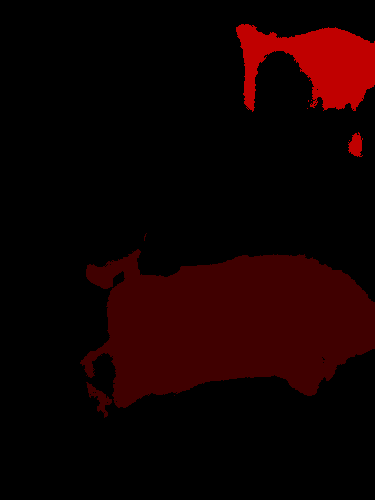}
\end{subfigure}
\begin{subfigure}[b]{.11\linewidth}
\includegraphics[width=\linewidth]{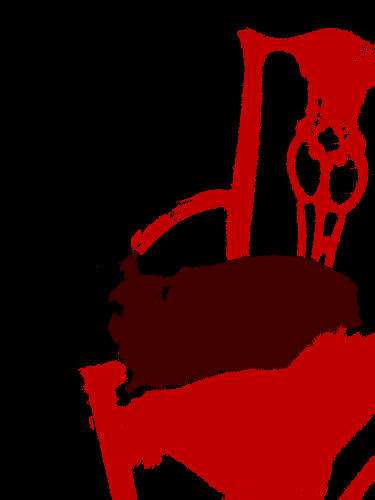}
\end{subfigure}
\begin{subfigure}[b]{.11\linewidth}
\includegraphics[width=\linewidth]{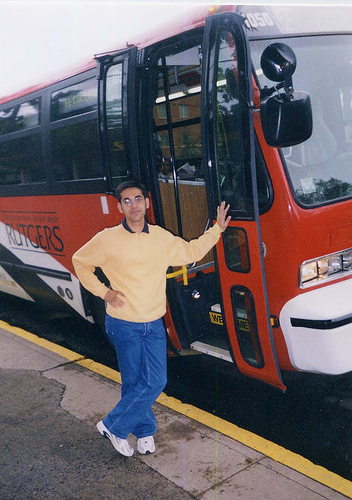}
\end{subfigure}
\begin{subfigure}[b]{.11\linewidth}
\includegraphics[width=\linewidth]{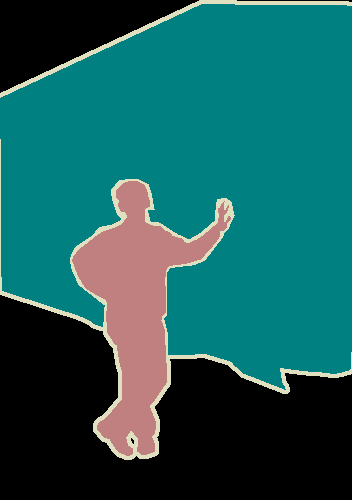}
\end{subfigure}
\begin{subfigure}[b]{.11\linewidth}
\includegraphics[width=\linewidth]{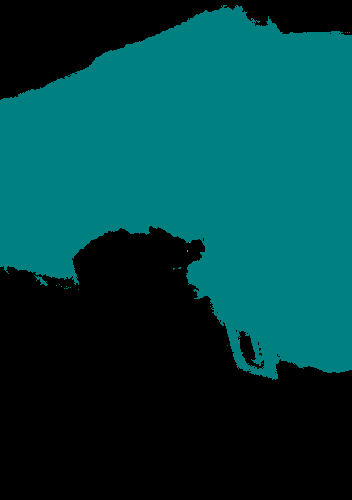}
\end{subfigure}
\begin{subfigure}[b]{.11\linewidth}
\includegraphics[width=\linewidth]{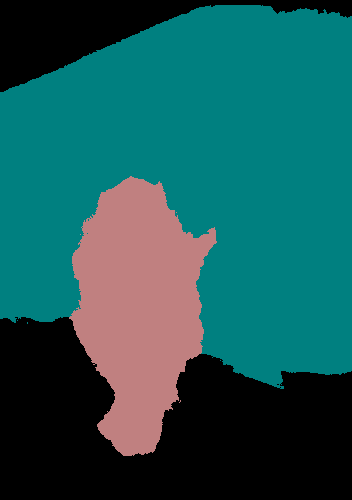}
\end{subfigure}
\begin{subfigure}[b]{.11\linewidth}
\includegraphics[width=\linewidth]{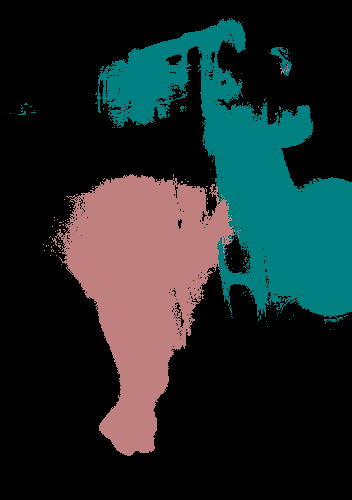}
\end{subfigure}
\begin{subfigure}[b]{.11\linewidth}
\includegraphics[width=\linewidth]{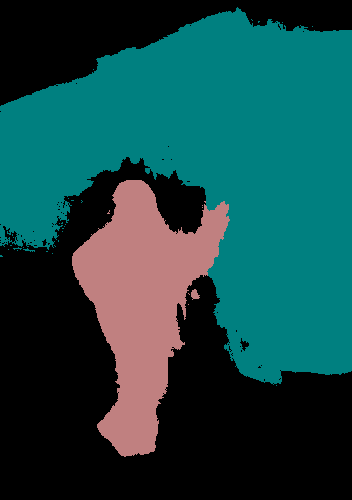}
\end{subfigure}
\begin{subfigure}[b]{.11\linewidth}
\includegraphics[width=\linewidth]{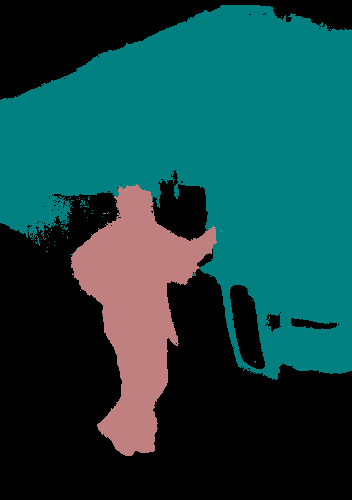}
\end{subfigure}
\begin{subfigure}[b]{.11\linewidth}
\includegraphics[width=\linewidth]{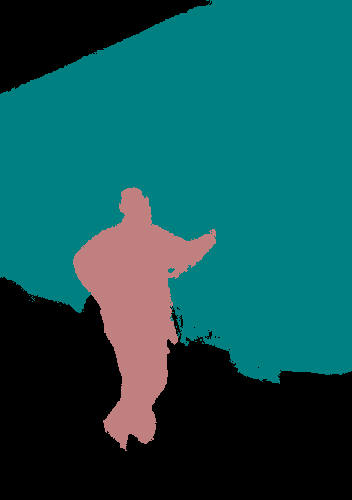}
\end{subfigure}
\begin{subfigure}[b]{.11\linewidth}
\includegraphics[width=\linewidth]{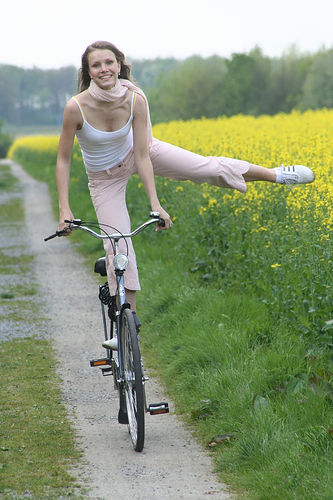}
\caption{rgb}\label{fig:dola}
\end{subfigure}
\begin{subfigure}[b]{.11\linewidth}
\includegraphics[width=\linewidth]{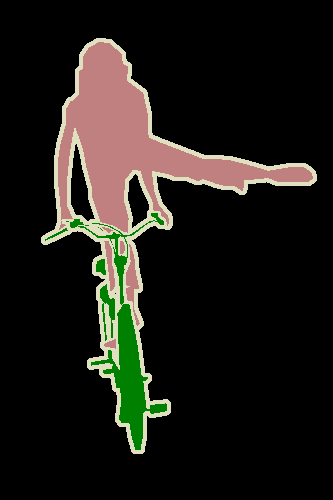}
\caption{GT}\label{fig:dolb}
\end{subfigure}
\begin{subfigure}[b]{.11\linewidth}
\includegraphics[width=\linewidth]{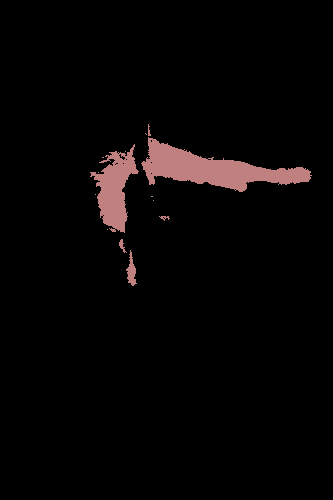}
\caption{ICD}\label{fig:dolc}
\end{subfigure}
\begin{subfigure}[b]{.11\linewidth}
\includegraphics[width=\linewidth]{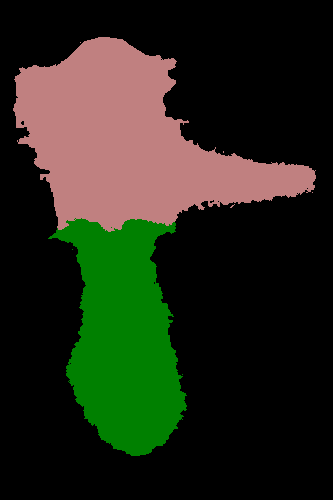}
\caption{CIAN}\label{fig:dold}
\end{subfigure}
\begin{subfigure}[b]{.11\linewidth}
\includegraphics[width=\linewidth]{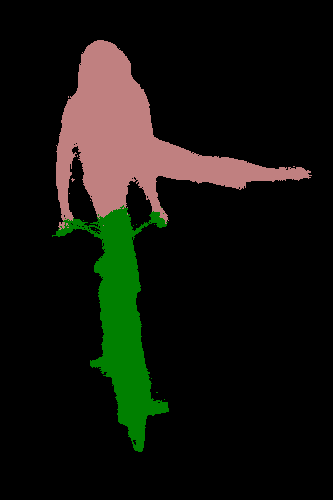}
\caption{MCIS}\label{fig:dole}
\end{subfigure}
\begin{subfigure}[b]{.11\linewidth}
\includegraphics[width=\linewidth]{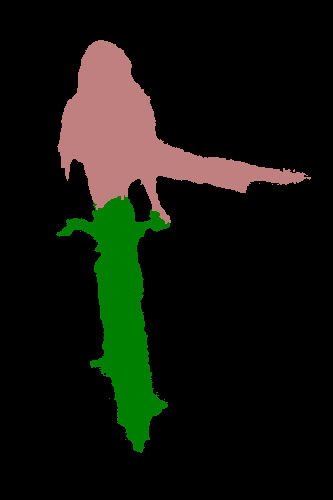}
\caption{EPS}\label{fig:dolf}
\end{subfigure}
\begin{subfigure}[b]{.11\linewidth}
\includegraphics[width=\linewidth]{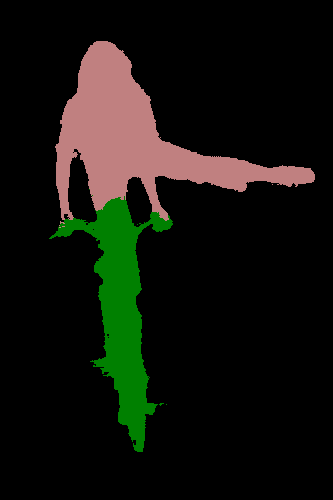}
\caption{EDAM}\label{fig:dolg}
\end{subfigure}
\begin{subfigure}[b]{.11\linewidth}
\includegraphics[width=\linewidth]{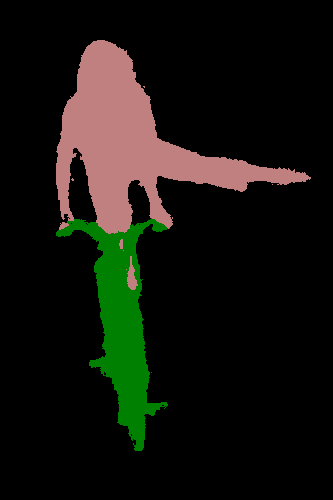}
\caption{Ours}\label{fig:dolh}
\end{subfigure}

\caption{Qualitative comparison on Pascal VOC 2012 validation set.}\label{fig:qual6}
\end{figure}

\noindent
Figure \ref{fig:qual6} compares our qualitative results on Pascal VOC 2012 val set with other approaches whose implementation we have used to generate the images; therefore, they might be biased.

\subsection{Limitations}
We observed that ViT-PCM is biased on the most discriminative features.
Many approaches to WSSS highlight the improvements due to processing pixel relations, boundaries, and neighbourhoods. We have used only the conditioning from HV-BiLSTM, which might not be the best solution. On the other hand, some recent approaches have explored contrastive loss for foreground-background learning with no image-level supervision. Since the background is our Achille's heel, we could have explored this idea. Another bottleneck of our approach is the final scaling to map patches to pixels, where we perform a rough scaling to keep the resources limited.

\typeout{---------------------Conclusions -----------------------}
\section{Conclusions}

We presented an innovative, simple and end-to-end method, ViT-PCM, based on ViT for generating baseline pseudo-masks (BPM) with precise localization and higher quality than those obtained from the more involved CAM CNN-based architectures. We obtained new state-of-the-art in BPM generation with 67.7 \% mIoU on PascalVOC 2012 {\em train set}  and 71.4\% mIoU using  CRF only in post-processing. These results demonstrate this work's high contribution to the field of WSSS. Therefore, we hope that others will continue in this direction. In the supplementary files, we report more analysis and results. The code is available at \url{https://github.com/deepplants/ViT-PCM}.
\bibliographystyle{styles/splncs04}
\bibliography{egbib.bib}

\end{document}


\pagestyle{headings}
\mainmatter
\def\ECCVSubNumber{2601}  

\title{Max Pooling with Vision Transformers reconciles class and shape in weakly supervised semantic segmentation\\ \normalfont{Supplementary Material}  } 


\titlerunning{ViT-PC for WSSS by Image-Class Labels}
%
\titlerunning{ViT-PC for WSSS by Image-Class Labels}

\author{Simone Rossetti\inst{1,2}\orcid{0000-0002-5344-7872},
Damiano Zappia\inst{1}\orcid{0000-0002-5726-2488},
Marta Sanzari\inst{2}\orcid{0000-0002-4640-9122},\\
Marco Schaerf\inst{1,2}\orcid{0000-0002-2016-1966},
Fiora Pirri\inst{1,2}\orcid{0000-0001-8665-9807}}
\authorrunning{Rossetti et al.}
%
\institute{DeepPlants,
\email{@deepplants.com}\\
\and
DIAG, Sapienza
\email{@diag.uniroma1.it}}

\maketitle
\section{Derivation of the ${\mathcal L_{MCE}}$ optmization}
\setcounter{equation}{0}
Here to keep this supplementary section self contained we reintroduce the whole paragraph comprehensive of the parts we omitted in the  main paper.

Let us indicate by $f$ the network taking inputs from a dataset $\mathcal{D}{=}\{\langle X_{in},t\rangle\}$.  Here $X_{in} {\in} {\mathbb R}^{h{\times} w{\times}3}$ indicates an input images, possibly obtained from an augmented and transformed set,  $t{\in} \{0,1\}^{K}$ are the ground truth binary labels, and $K$ is the number of classes defined by the category set ${\mathcal C}{=}\{0,1,{\ldots}, K\}$. The output of $f$ is a tensor $\hat{Y}\in {\mathcal C}^{h{\times} w}$ which is a {\em baseline pseudo-mask}.

ViT is part of $f$. We recall that ViT partitions the image $X$, resized image of the original  $X_{in}$,  into $s$ patches of size $(d{\times} d {\times} 3)$. In particular, we are interested in the feature maps $F{\in} {\mathbb R}^{s\times e}$, with $s{=}(n/d)^2$, with $n{=}w{=}h$.  The feature maps $F$  are the encoded representations of the patches, obtained by ViT.  $F$ represent the basis functions specifying the patches internal structure.  

\subsubsection{Explicit search by Global Max-Pooling}
Given $F{\in} {\mathbb R}^{s\times e}$, we consider also a weight matrix $W{\in}{\mathbb R}^{e{\times}K}$  whose weights are taken into account in the  optimization method described below. More precisely, we estimate the baseline pseudo-mask $\hat{Y}$, training the weights $W$  with only image-level class labels as supervision,  minimizing the multilabel classification error. 

The first objective  is to minimize the multilabel classification prediction error (MCE). Thus, given the ground truth binary labels $t$ defined above, and recalling that $K$ are the number of classes,  we model the multi-label classification using $K$ independent Bernoulli distributions and $K$ binary cross-entropy losses (BCE):
\begin{equation}\label{eq:bce}
    \mathcal{L}_{MCE}=\frac{1}{K}\sum_{k\in{\mathcal C}} BCE(t_k,y_k)=-\frac{1}{K}\sum_{k\in {\mathcal C}} t_k\log(y_k)+(1-t_k)\log(1-y_k).
\end{equation}
Here, $y{\in} {\mathbb R}^{K}$  is obtained as described in the following.  
Let $Z {=} softmax(A)$ with $A {=} F W$, hence both $A$ and $Z$ are in  ${\mathbb R}^{s{\times}K}$, since $W{\in}{\mathbb R}^{e{\times}K}$. $Z$ are the semantic segmentation predictions, needing to be projected into class predictions. We do so using global max pooling (GMP):
\begin{equation}\label{eq:gmp}
y_k = GMP(Z^k) = \max (Z^k) = Z^k_i, \mbox{ for some  } i{\in}\{0,1,{\ldots},s\}
\end{equation}
where,
\begin{equation}\label{eq:output}
Z^k {=} softmax(A^k) {=} \begin{pmatrix}
\frac{\exp(A_{1}^k)}{\sum_{c\in {\mathcal C}}\exp(A_{1}^c)} \\
\vdots \\
\frac{\exp(A_{s}^k)}{\sum_{c\in {\mathcal C}} \exp(A_{s}^c)} 
\end{pmatrix} \text{ and }  A_{j}^k{=} F_jW^k   
\end{equation}
As defined above,  $F_j$ is the feature map of patch $U_j$, while  $A_j^k$ is the logit of patch $U_j$, $j{=}0,{\ldots},s$ with respect to class $k\in\{0,1,{\ldots},K\}$, where the $k{=}0$ class is predicted within the optimization. 


The relative error back propagation of ${\mathcal L}_{MCE}$ w.r.t weights $W$ is given by:
\begin{equation}
    \frac{\partial \mathcal{L}_{MCE}}{\partial W}{=}\sum_{k=0}^K\frac{\partial BCE(t_k,y_k)}{\partial W}
\end{equation}
To simplify we analyze the gradient with respect to each column $q$ of the weights $W$, with $q{=}0,1,{\ldots},K$. Applying the chain rule:

\begin{equation}\label{eq:chain}
    \frac{\partial BCE(t_k,y_k)}{\partial W^q}=\frac{\partial BCE(t_k,y_k)}{\partial y_k}\frac{\partial y_k}{\partial \max(Z^k)}\frac{\partial \max(Z^k)}{\partial A^q}\frac{\partial A^q}{\partial W^q}
\end{equation}
 Each term of Eq. (\ref{eq:chain}) is derived in the following:
\begin{equation}
    \frac{\partial BCE(t_k,y_k)}{\partial y_k}=-\frac{t_k}{y_k}+\frac{1-t_k}{1-y_k}=\frac{y_k-t_k}{y_k(1-y_k)}
\end{equation}
\begin{equation}\label{eq:dydz}
    \frac{\partial y_k}{\partial \max (Z^k)}=
    1 \mbox{, since Eq. (\ref{eq:gmp})}
\end{equation} 
By Eq. (\ref{eq:dydz}) we instantiate $\max(Z^k)$ as  $Z_i^k$. Rewriting $\frac{\partial \max(Z^k)}{\partial A^q}$ as $\frac{\partial Z_i^k}{\partial A^q}$  we get the Jacobian of 
$\frac{\partial Z_i^k}{\partial A^q}$ as:
\begin{equation}
\begin{array}{lc}
\frac{\partial Z_i^k}{\partial A^q} = \left(
\begin{array}{ccc}
\frac{\partial Z_i^k}{\partial A_0^0} &{\ldots} &\frac{\partial Z_i^k}{\partial A_0^K}\\
\vdots& \ddots &\vdots\\
\frac{\partial Z_i^k}{\partial A_s^0}&{\ldots} &\frac{\partial Z_i^k}{\partial A_s^K}
\end{array}\right)
\end{array}
\end{equation}
hence for $j{\in} \{0,1,{\ldots},s\}$:
\begin{equation}\label{eq:dzda}
\begin{split}
    \frac{\partial Z_{i}^k}{\partial A_j^q} &=
    \begin{cases}
    0 & \mbox{ if }  i\neq j\\
    \frac{\exp(A_{j}^k)}{\sum_{c\in {\mathcal C}} \exp(A_{j}^c)}-\Big(\frac{\exp(A_{j}^k)}{\sum_{c\in {\mathcal C}} \exp(A_{j}^c)}\Big)^2 & \mbox{ if } i=j \mbox{ and }  q=k\\
    -\frac{\exp(A_{j}^k)\exp(A_{j}^q)}{\big(\sum_{c\in {\mathcal C}} \exp(A_{j}^c)\big)^2} &
    \mbox{ if } i=j \mbox{ and } q\neq k 
    \end{cases}\\
\end{split}
\end{equation}
From Eq. (\ref{eq:dydz}) $y_k{=} Z_i^k$, and from the above Eq. (\ref{eq:dzda}) we instantiate  $A^q$ with  $A_i^q$,  then from  Eq. (\ref{eq:dzda}) it follows that:
\begin{equation}\label{eq:dzdaR}
\begin{split}
\frac{\partial Z_{i}^k}{\partial A_i^q} &=\begin{cases}
    y_k(1-y_k) & q=k\\
    -y_k Z_{i}^q & q\neq k \\
    \end{cases} 
    \end{split}
\end{equation}
Consider the following matrix:
\begin{equation}
\begin{array}{l}
Z =
   \begin{array}{lr}
     \left.\underbrace{ \left(
    \begin{array}{ccccc}
     Z_0^0 &{\ldots} & Z_0^q &{\ldots} & Z_0^K\\
       \vdots& \ddots &\vdots& \ddots &\vdots\\
      Z_s^0 &{\ldots} & Z_s^q &{\ldots} &Z_s^K
     \end{array}\right) 
     } \ \ \ \ \right\}
   \sum_{k=0}^K Z_i^{k}=1, \ \ i {=} 0,\ldots, s
   \end{array}\\
   \hspace*{1.1cm}  \max(Z^q) = Z^q_j\\  
    \hspace*{1.1cm} j{=}0,{\ldots}, s,\\
     \hspace*{1.1cm} q{\in}\{0,{\ldots},K\}
  \end{array}
\end{equation}

From the above setting, which is like the vector of Eq. (\ref{eq:output}) repeated up to $K$, we can see that for each row we have a categorical distribution, and it sums to one.
More precisely, consider  the above  matrix $Z\in {\mathbb R}^{s{\times}K}$, we have that each $Z_i^q$ is the softmax of $A_i^q$, and indicates the probability that the corresponding patch $U_i$ is of class $q$. Namely:
\begin{equation}
p(U_i^0,{\ldots},U_i^K)  = \prod_{k=0}^K Z_i^{[q=k]}
\end{equation}
Which is, indeed, the joint mass function for patch $U_i$ of a categorical distribution with probabilities $Z_i^0,{\ldots}, Z_i^K$.

On the other hand, along each column, taking the maximum for each of them, we obtain
\begin{equation}
GMP(Z) = (\max(Z^0), \max(Z^1),{\ldots}, \max(Z^K)). 
\end{equation} 
$GMP(Z)$ then gives the probability for the $q$-th class to appear in the image $X_{in}$, it specifies, indeed, a multi-label classification. Note that, since $\max(Z^q) = Z_i^q$, for some $i{=}0,{\ldots},s$, we also know the location of the category, with respect to the patch $U_i^q$.

From the last two terms of the r.h.s. of eq.(\ref{eq:chain}) and the definition of $A$ we obtain that:
\begin{equation}
    \frac{\partial A^q_{i}}{\partial W^q} = F_{i}.
\end{equation}
We can see that $i = 0,\ldots, s$  depends on the choice of $A$, which in turns depends on the index of $max(Z^k)$. 

 Finally, we get the error backpropagation with respect to the network weights:
\begin{equation}\label{eq:backprop}
    \frac{\partial BCE(t_k,y_k)}{\partial W^q}= F_{i} \cdot
    \begin{cases}
    y_k -t_k & q=k\\
    Z_{i}^q \frac{t_k-y_k}{1-y_k} & q\neq k
    \end{cases}
\end{equation}
The gradients have size $\frac{\partial y_k}{\partial Z^k}{\in} \mathbb{R}^{s}$, $\frac{\partial Z^k}{\partial A^q}{\in}\mathbb{R}^{s\times{s}}$,  $\frac{\partial A^q}{\partial W^q}{\in} \mathbb{R}^{s\times{e}}$, and 
$\frac{\partial BCE(t_k,y_k)}{\partial W^q}{\in} \mathbb{R}^{e}$. Note that in equation (\ref{eq:backprop}), according to equation (\ref{eq:gmp}),
the subscript $i$, varying in $0,\ldots,s$  concerns the GMP computed with respect to a specific class $k$, and either the choice of the column $q$ for $W$ is equal to such a $k$ or it is not. We consider both cases, relative to the index where $Z^k$ is maximum.  

To keep track of the index $i$ w.r.t. the specific class, for notational purpose we indicate by $i_a$  the location at which the value $Z^a$ is maximum,  which we use improperly as a subscript also for the feature vectors $F$.  Let us consider again   the column $q$ of the weights $W$, this column will be updated by the quantity:
\begin{equation}\label{eq:backproploss_SUPP}
\begin{split}
   \frac{\partial \mathcal{L}_{MCE}}{\partial W^q}&=
   \frac{\partial BCE(t_q,y_q)}{\partial W^q}
   +
   \displaystyle{\sum_{\substack{k\in C\\k\neq q}}
   \frac{\partial BCE(t_k,y_k)}{\partial W^q}}\\
   &= -F_{i_q}(t_q - y_q) + \displaystyle{\sum_{\substack{k\in C\\k\neq q}} F_{i_k}Z_{i_k}^q \frac{t_k -y_k}{1-y_k}  } 
\end{split}
\end{equation}
Eq. \ref{eq:backproploss_SUPP} specifies the linear-search mechanism of the proposed optimization, iteratively selecting the most representative features $F_{i_q}$ of each category $q$. At each step, the optimization  updates the full column rank matrix $W{\in} {\mathbb R}^{s{\times}e}$ and returns the minimum error norm solution, which  separates the feature vector space $\mathbb{R}^e$ into $K$ linear sub-spaces. 
Considering the optimization manifold,   the vector  $W^q$ moves  in the direction of the best representative feature vector $F_{i_u}$, with either $u$ being of the same category of the chosen column $q$, or not. More precisely, at each iteration, $W^q$ moves in the direction of $F_{i_q}$  according to the error value $(t_q - y_q)$, and in the direction $F_{i_k}$ according to the term $Z_{i_k}^q \frac{t_k -y_k}{1-y_k}$, for any category $k$, with $k\neq q$.

More specifically, when the term $\frac{(t_k -y_k)}{1-y_k}{=}1$,  and the category $k{\neq} q$ is considered,   $W^q$ moves in the direction opposite to the best representative feature vector $F_{i_k}$. On the other hand, when $t_k=0$  the term considered is $ -(Z_{i_k}^k \frac{y_k}{1-y_k})$ which is added to $W^q$, for its  updating. Note that, in this case,  the update term is increasingly small, since $y_k{\ll}1- y_k$ as $y_k{\to}0$.  
This optimization method, based on iterative learning and stochastic gradient descent, induces a separation in the space of patch features, according to the multilabel classification.
\section{Further Experiments and Results}

\subsection{Qualitative Results on Pascal VOC 2012}
\begin{figure}
\centering
\begin{subfigure}{.31\textwidth}
\centering
\includegraphics[width=1.015\linewidth]{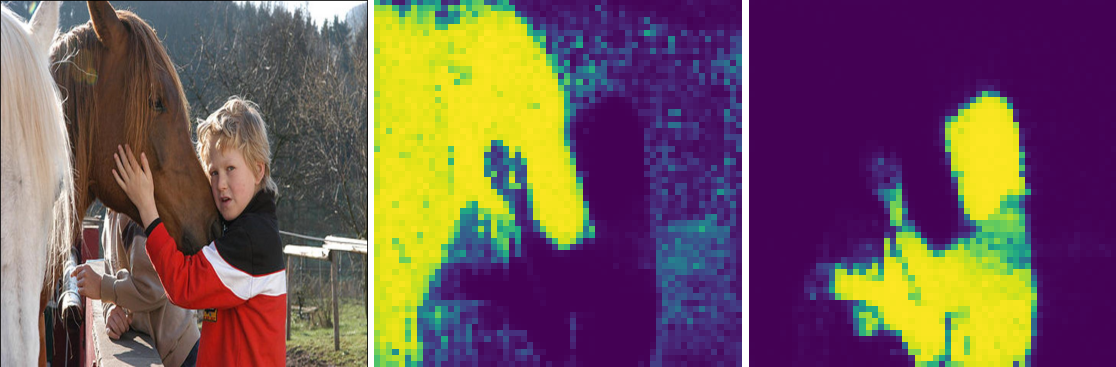}
\end{subfigure}
\begin{subfigure}{.31\textwidth}
\centering
\includegraphics[width=1.015\linewidth]{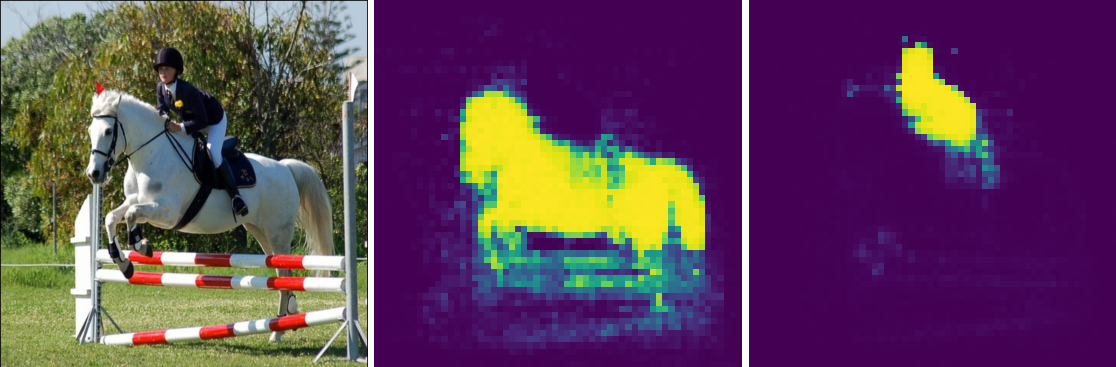}
\end{subfigure}
\begin{subfigure}{.31\textwidth}
\centering
\includegraphics[width=1.015\linewidth]{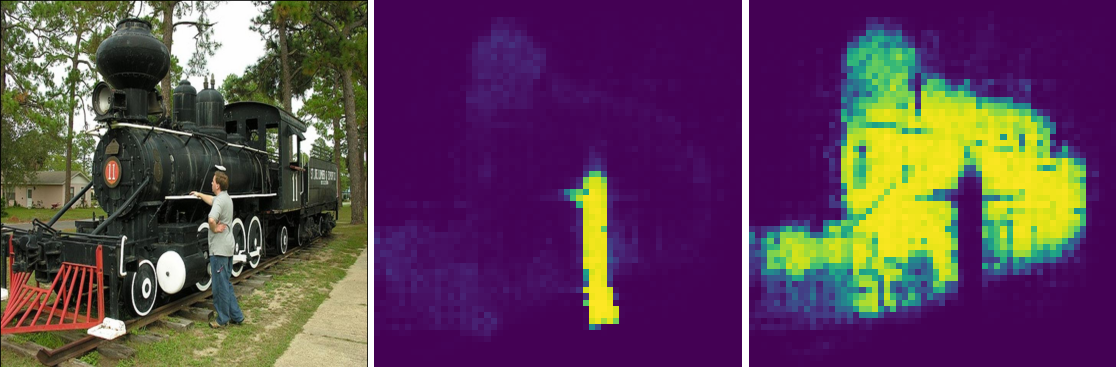}
\end{subfigure}\\
\begin{subfigure}{.31\textwidth}
\centering
\includegraphics[width=1.015\linewidth]{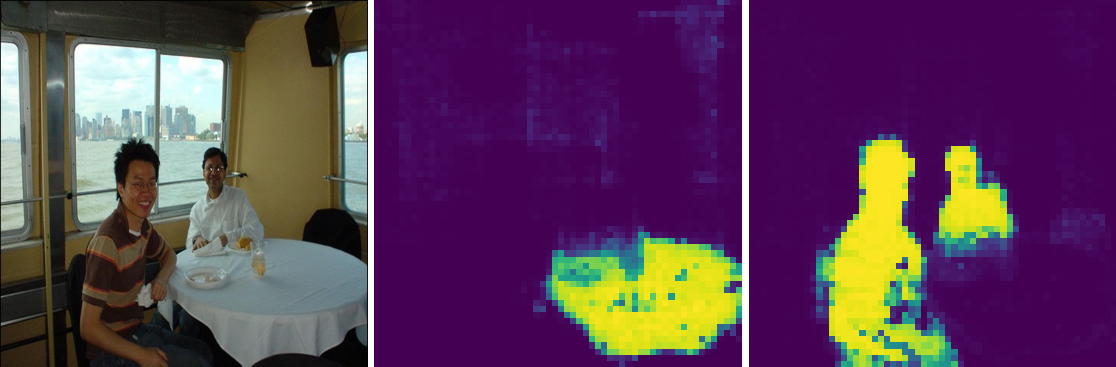}
\end{subfigure}
\begin{subfigure}{.31\textwidth}
\centering
\includegraphics[width=1.015\linewidth]{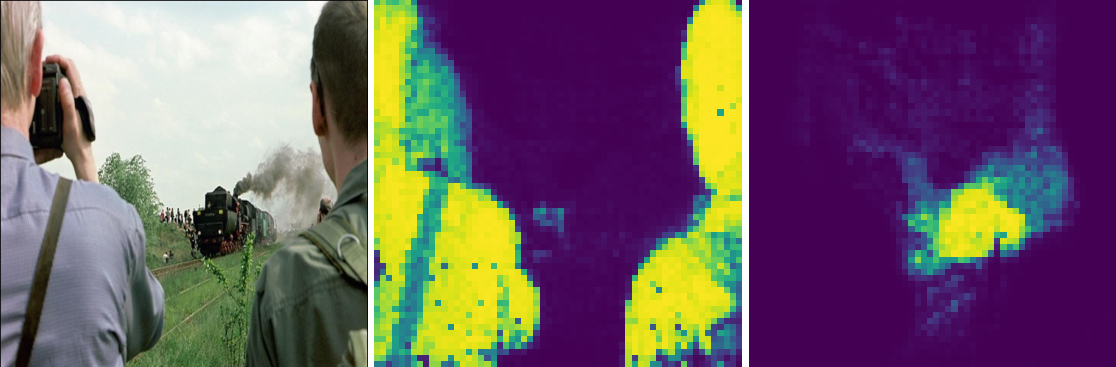}
\end{subfigure}
\begin{subfigure}{.31\textwidth}
\centering
\includegraphics[width=1.015\linewidth]{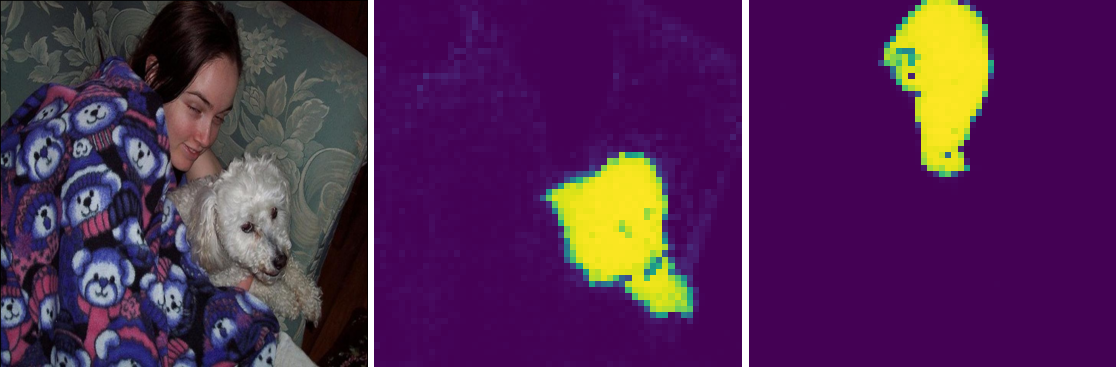}
\end{subfigure}\\
\begin{subfigure}{.31\textwidth}
\centering
\includegraphics[width=1.015\linewidth]{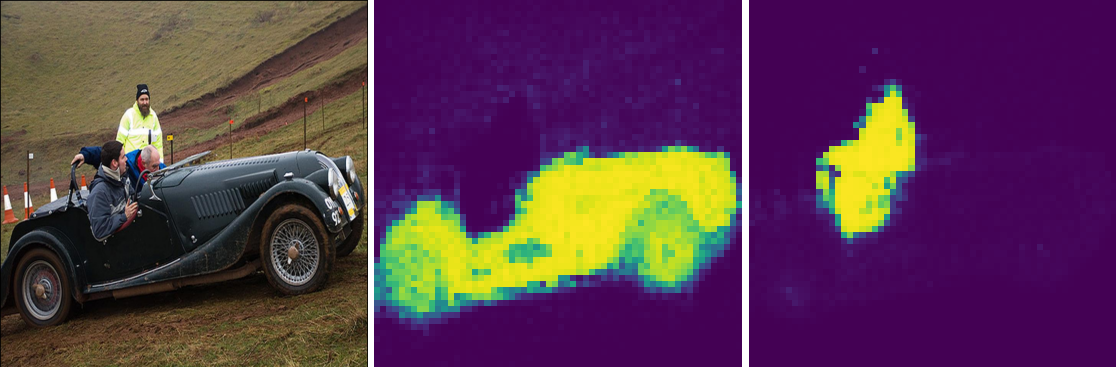}
\end{subfigure}
\begin{subfigure}{.31\textwidth}
\centering
\includegraphics[width=1.015\linewidth]{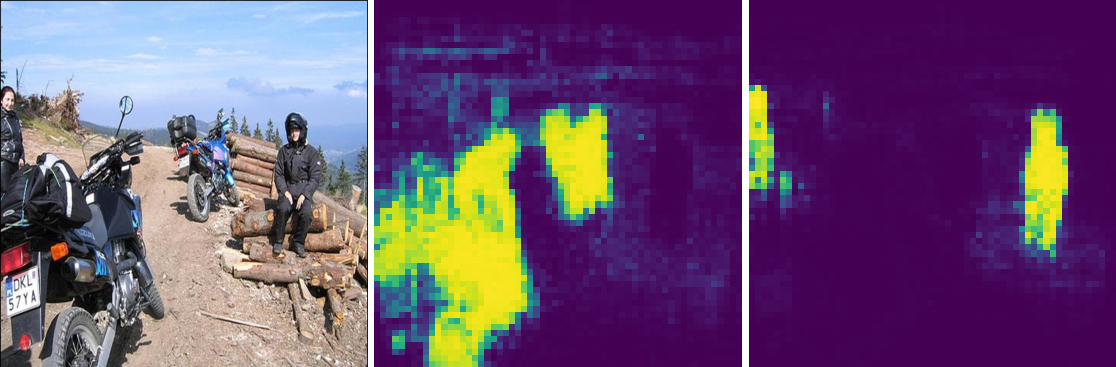}
\end{subfigure}
\begin{subfigure}{.31\textwidth}
\centering
\includegraphics[width=1.015\linewidth]{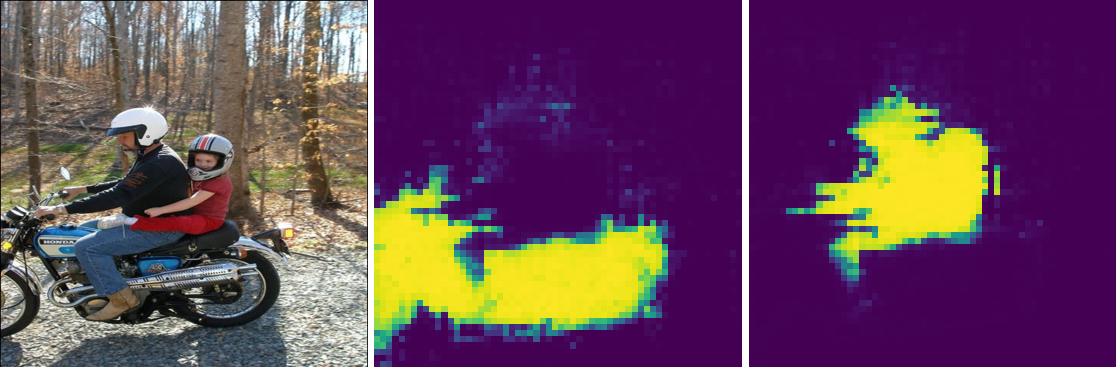}
\end{subfigure}\\
\begin{subfigure}{.31\textwidth}
\centering
\includegraphics[width=1.015\linewidth]{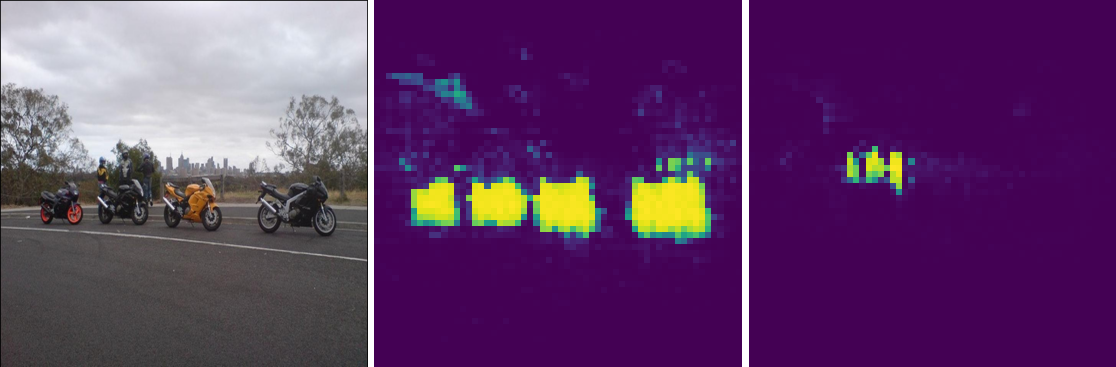}
\end{subfigure}
\begin{subfigure}{.31\textwidth}
\centering
\includegraphics[width=1.015\linewidth]{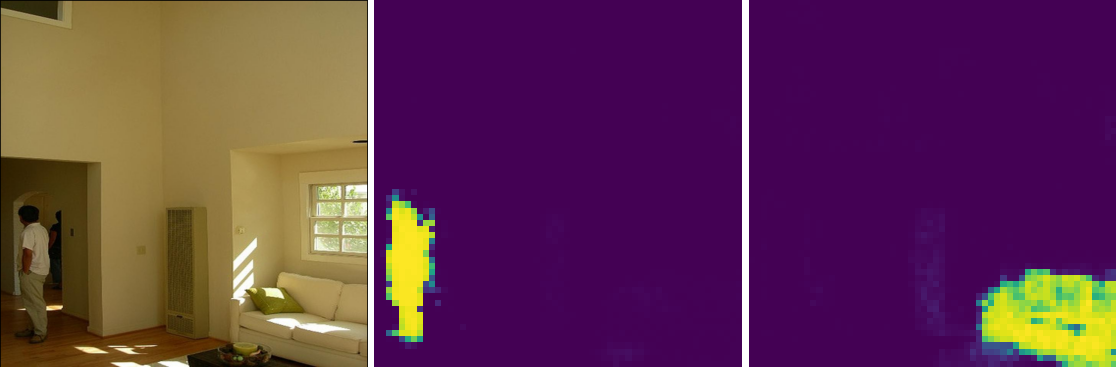}
\end{subfigure}
\begin{subfigure}{.31\textwidth}
\centering
\includegraphics[width=1.015\linewidth]{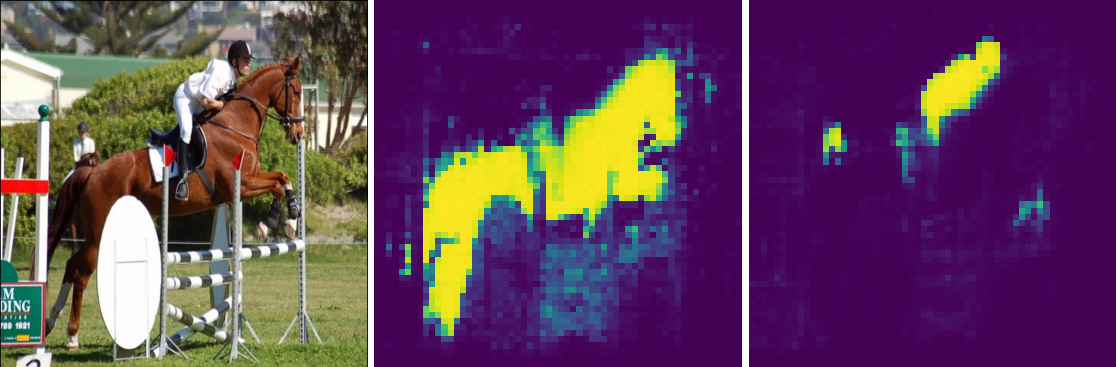}
\end{subfigure}\\
\begin{subfigure}{.31\textwidth}
\centering
\includegraphics[width=1.015\linewidth]{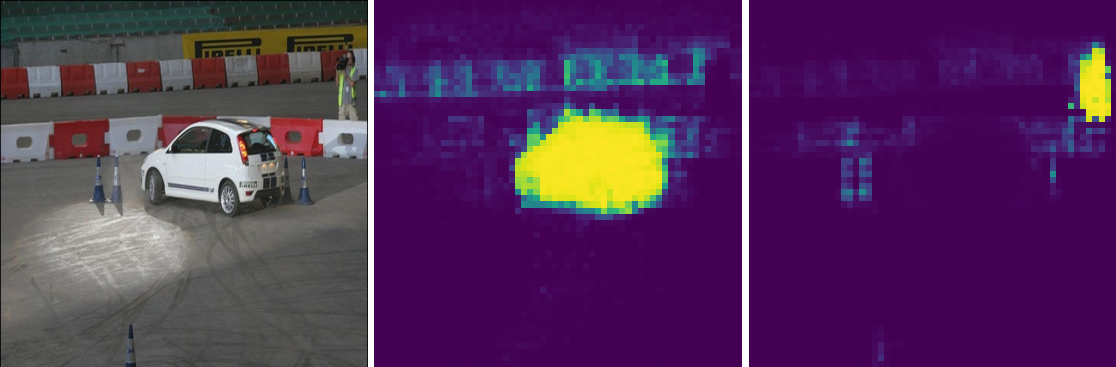}
\end{subfigure}
\begin{subfigure}{.31\textwidth}
\centering
\includegraphics[width=1.015\linewidth]{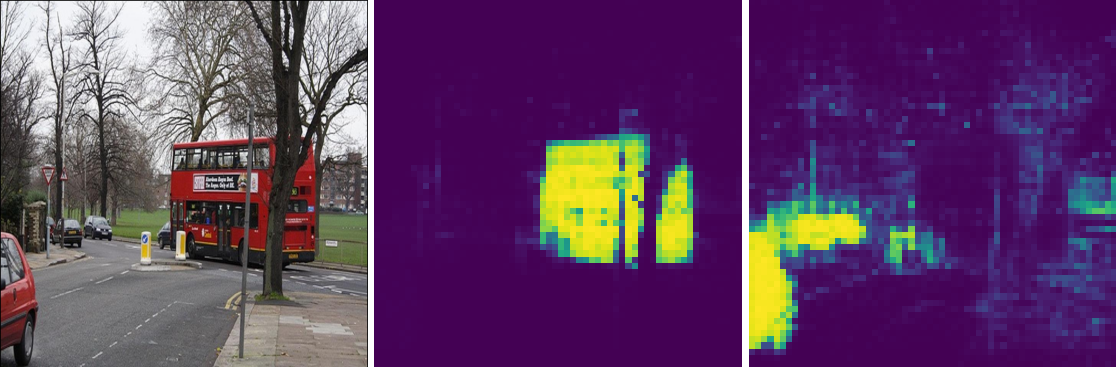}
\end{subfigure}
\begin{subfigure}{.31\textwidth}
\centering
\includegraphics[width=1.015\linewidth]{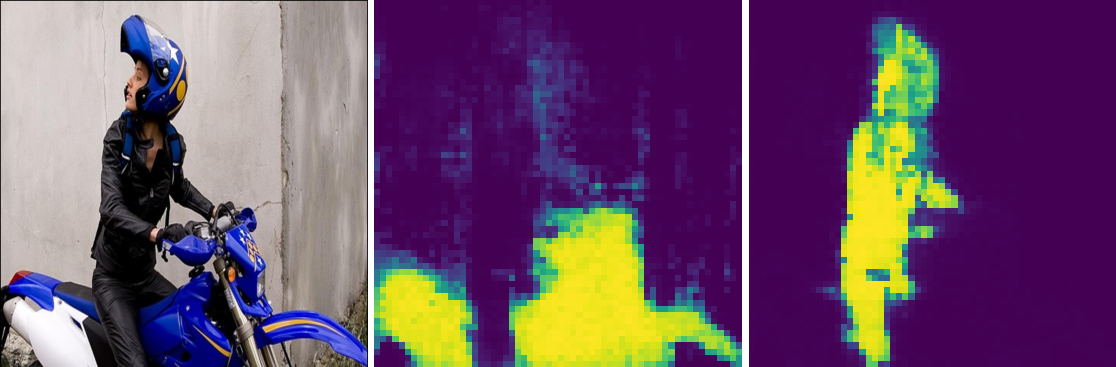}
\end{subfigure}\\
\begin{subfigure}{.31\textwidth}
\centering
\includegraphics[width=1.015\linewidth]{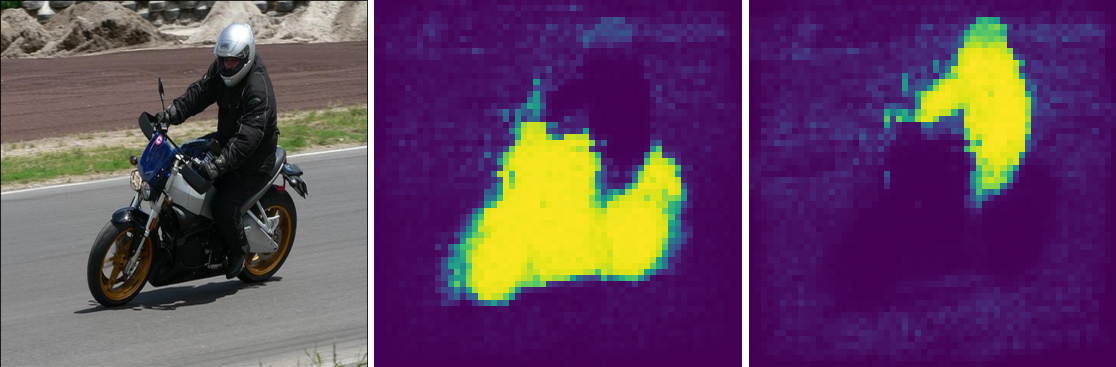}
\end{subfigure}
\begin{subfigure}{.31\textwidth}
\centering
\includegraphics[width=1.015\linewidth]{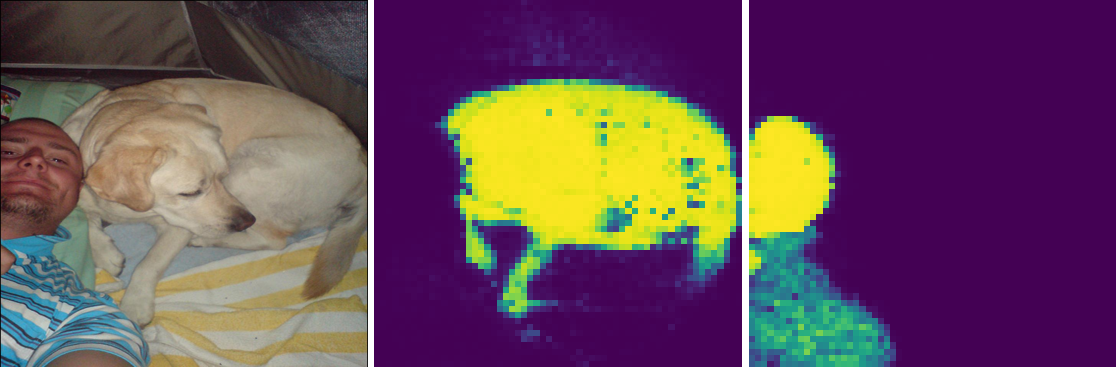}
\end{subfigure}
\begin{subfigure}{.31\textwidth}
\centering
\includegraphics[width=1.015\linewidth]{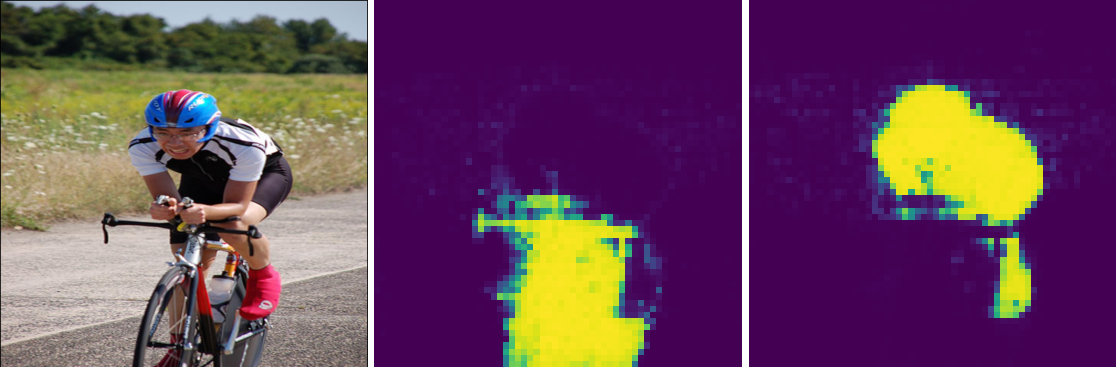}
\end{subfigure}\\
\begin{subfigure}{.31\textwidth}
\centering
\includegraphics[width=1.015\linewidth]{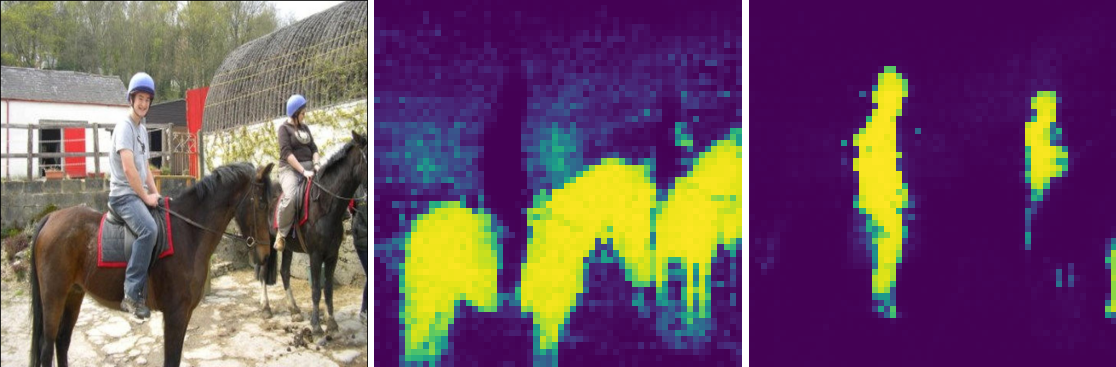}
\end{subfigure}
\begin{subfigure}{.31\textwidth}
\centering
\includegraphics[width=1.015\linewidth]{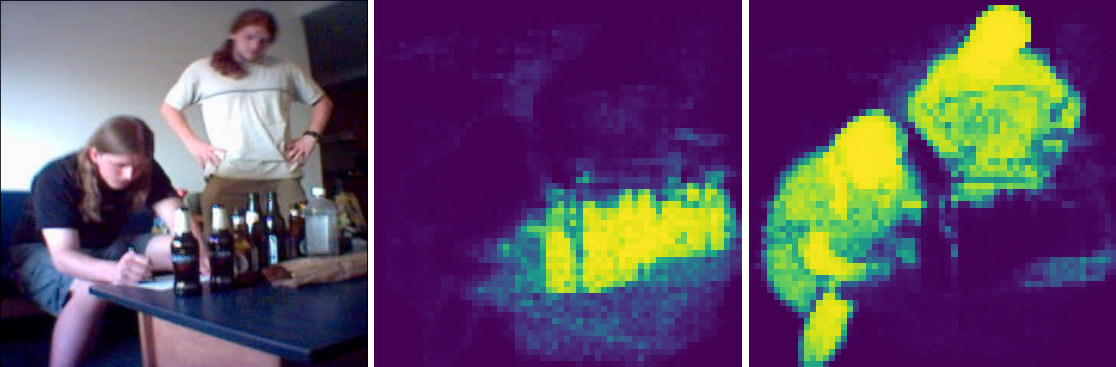}
\end{subfigure}
\begin{subfigure}{.31\textwidth}
\centering
\includegraphics[width=1.015\linewidth]{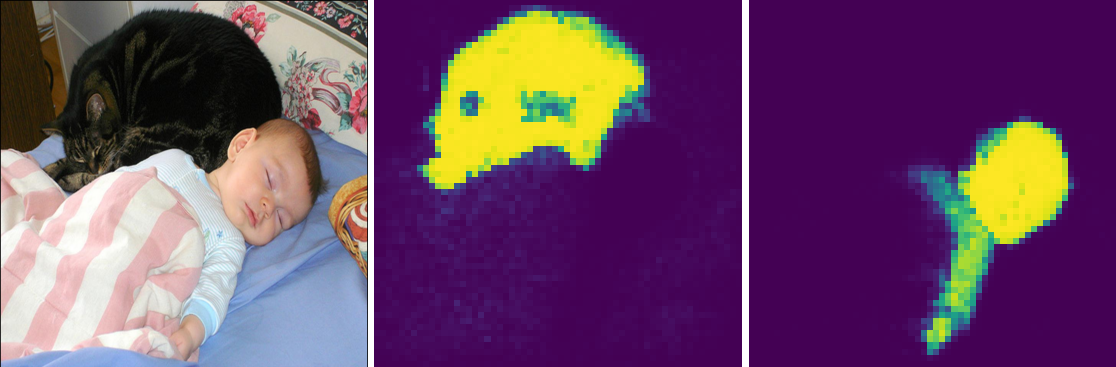}
\end{subfigure}\\
\begin{subfigure}{.31\textwidth}
\centering
\includegraphics[width=1.015\linewidth]{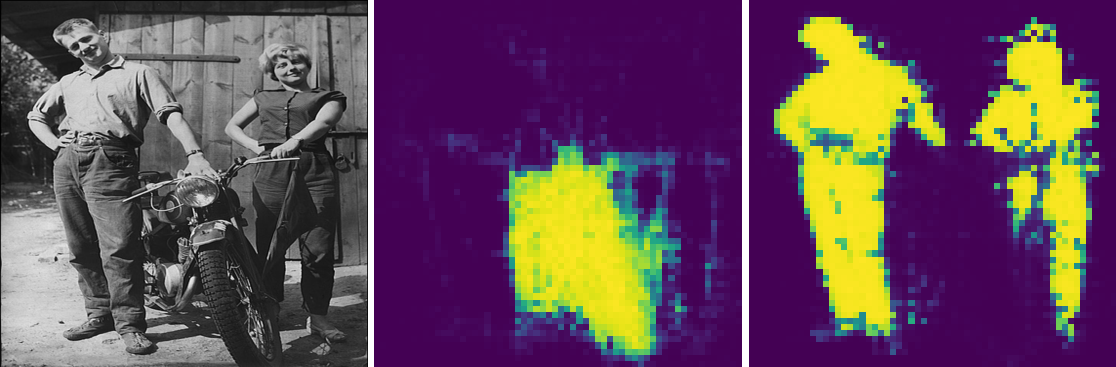}
\end{subfigure}
\begin{subfigure}{.31\textwidth}
\centering
\includegraphics[width=1.015\linewidth]{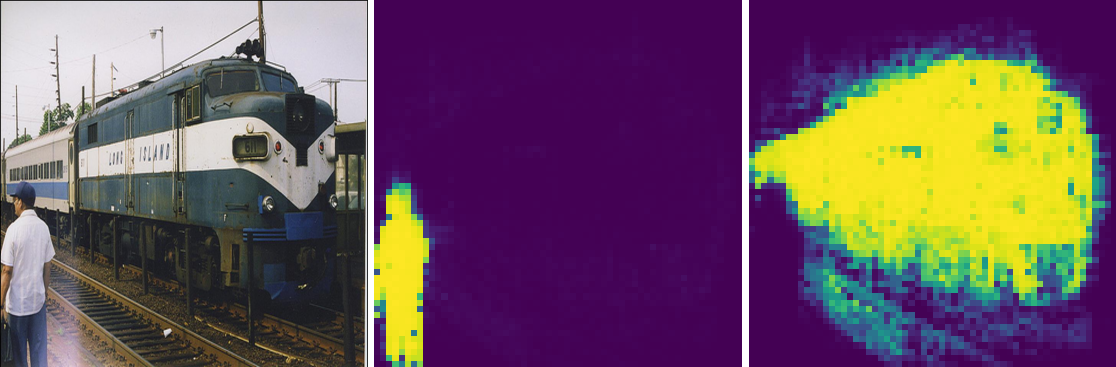}
\end{subfigure}
\begin{subfigure}{.31\textwidth}
\centering
\includegraphics[width=1.015\linewidth]{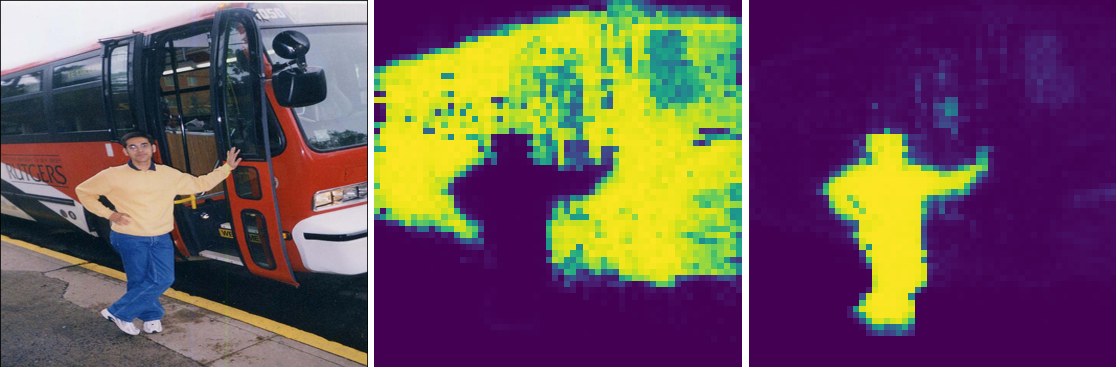}
\end{subfigure}\\
\begin{subfigure}{.31\textwidth}
\centering
\includegraphics[width=1.015\linewidth]{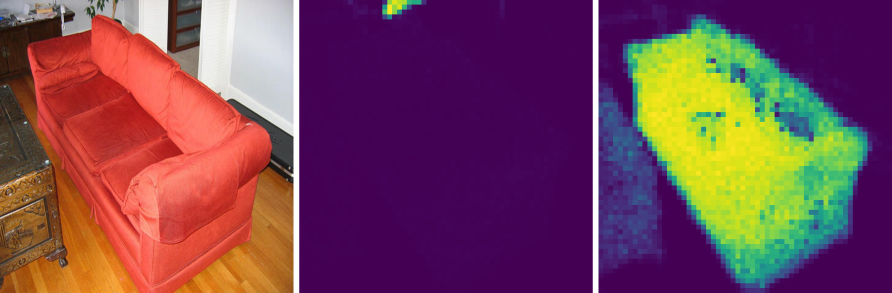}
\end{subfigure}
\begin{subfigure}{.31\textwidth}
\centering
\includegraphics[width=1.015\linewidth]{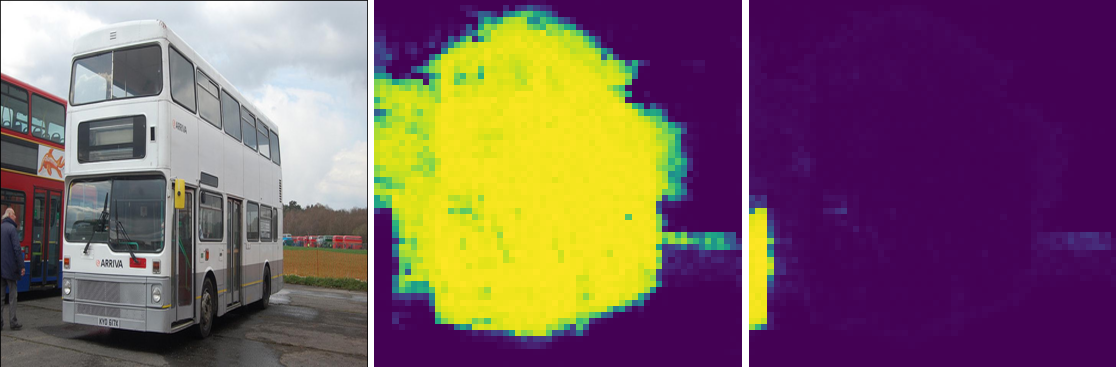}
\end{subfigure}
\begin{subfigure}{.31\textwidth}
\centering
\includegraphics[width=1.015\linewidth]{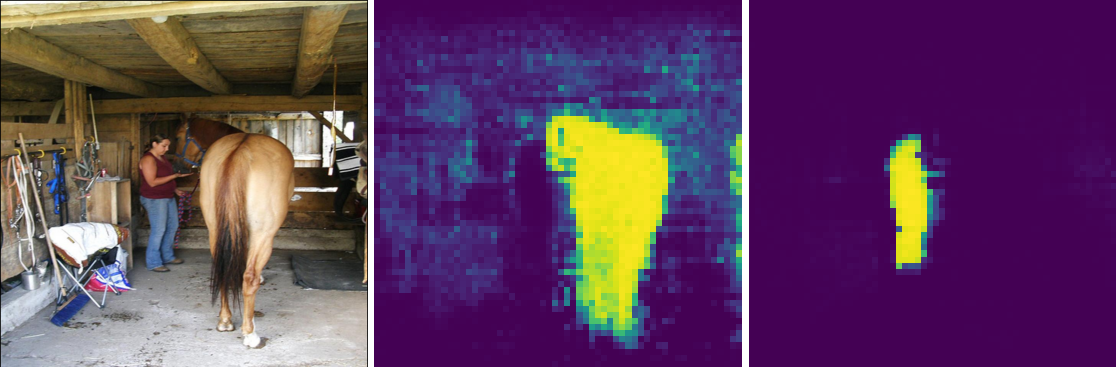}
\end{subfigure}\\
\begin{subfigure}{.31\textwidth}
\centering
\includegraphics[width=1.015\linewidth]{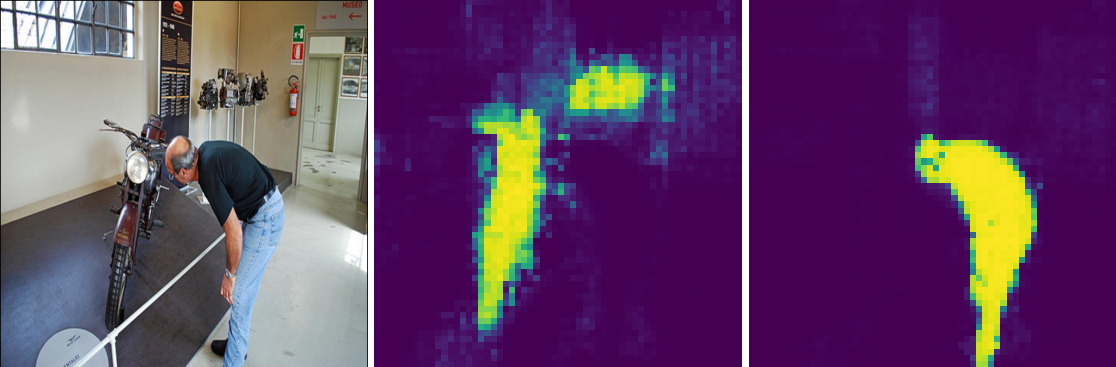}
\end{subfigure}
\begin{subfigure}{.31\textwidth}
\centering
\includegraphics[width=1.015\linewidth]{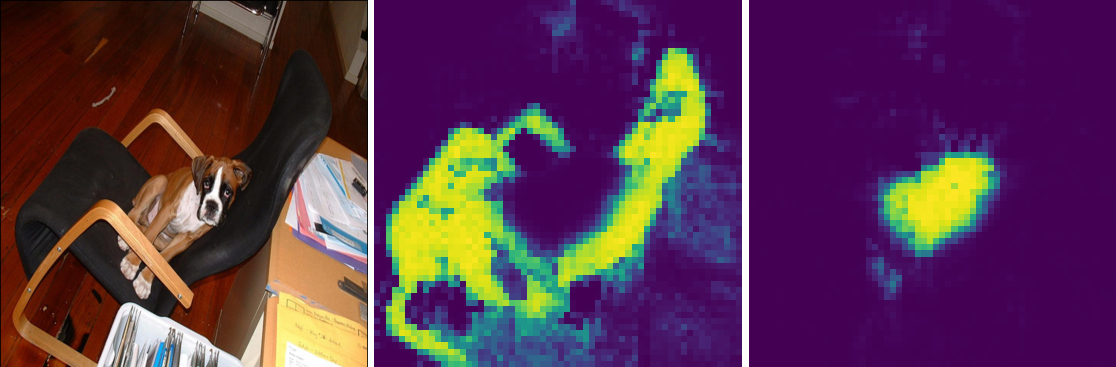}
\end{subfigure}
\begin{subfigure}{.31\textwidth}
\centering
\includegraphics[width=1.015\linewidth]{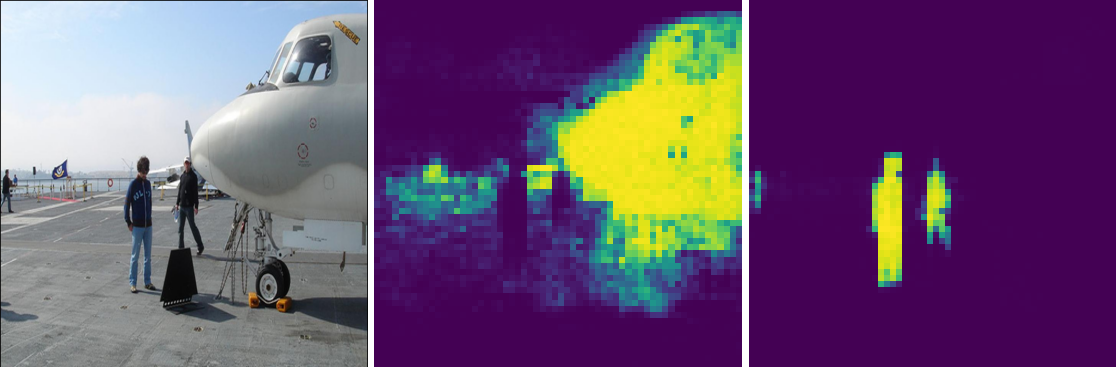}
\end{subfigure}\\
\begin{subfigure}{.31\textwidth}
\centering
\includegraphics[width=1.015\linewidth]{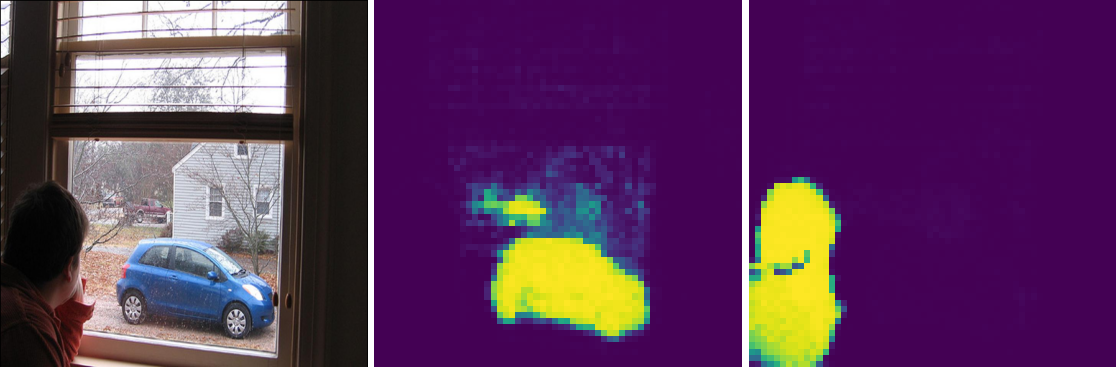}
\end{subfigure}
\begin{subfigure}{.31\textwidth}
\centering
\includegraphics[width=1.015\linewidth]{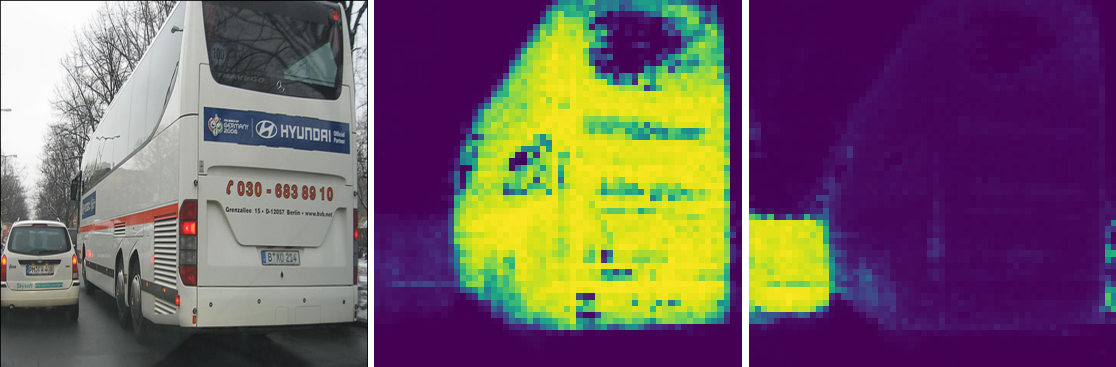}
\end{subfigure}
\begin{subfigure}{.31\textwidth}
\centering
\includegraphics[width=1.015\linewidth]{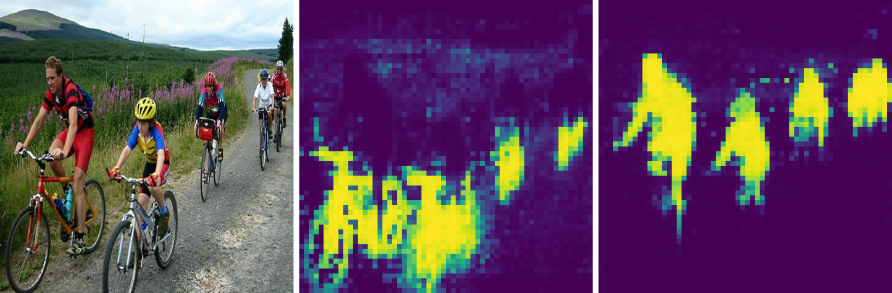}
\end{subfigure}\\
\begin{subfigure}{.31\textwidth}
\centering
\includegraphics[width=1.015\linewidth]{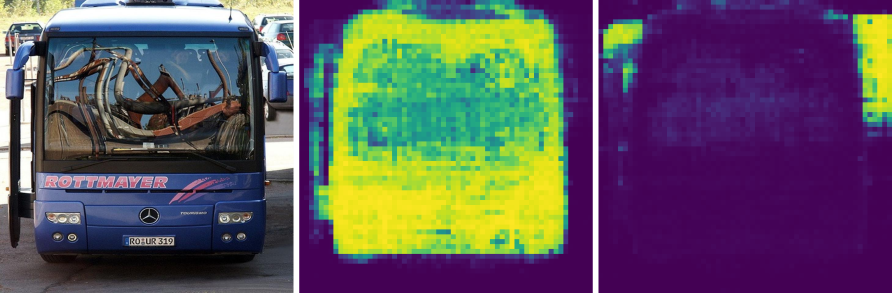}
\end{subfigure}
\begin{subfigure}{.31\textwidth}
\centering
\includegraphics[width=1.015\linewidth]{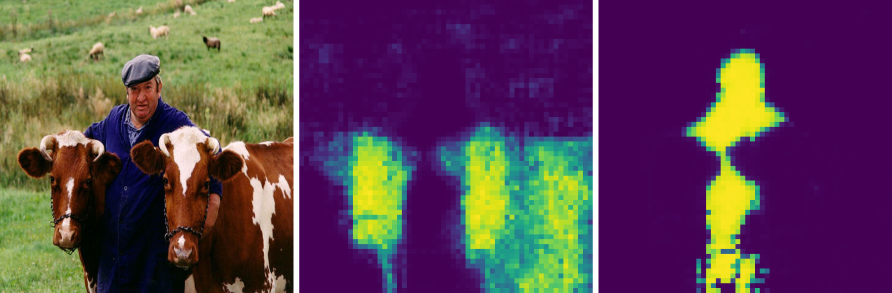}
\end{subfigure}
\begin{subfigure}{.31\textwidth}
\centering
\includegraphics[width=1.015\linewidth]{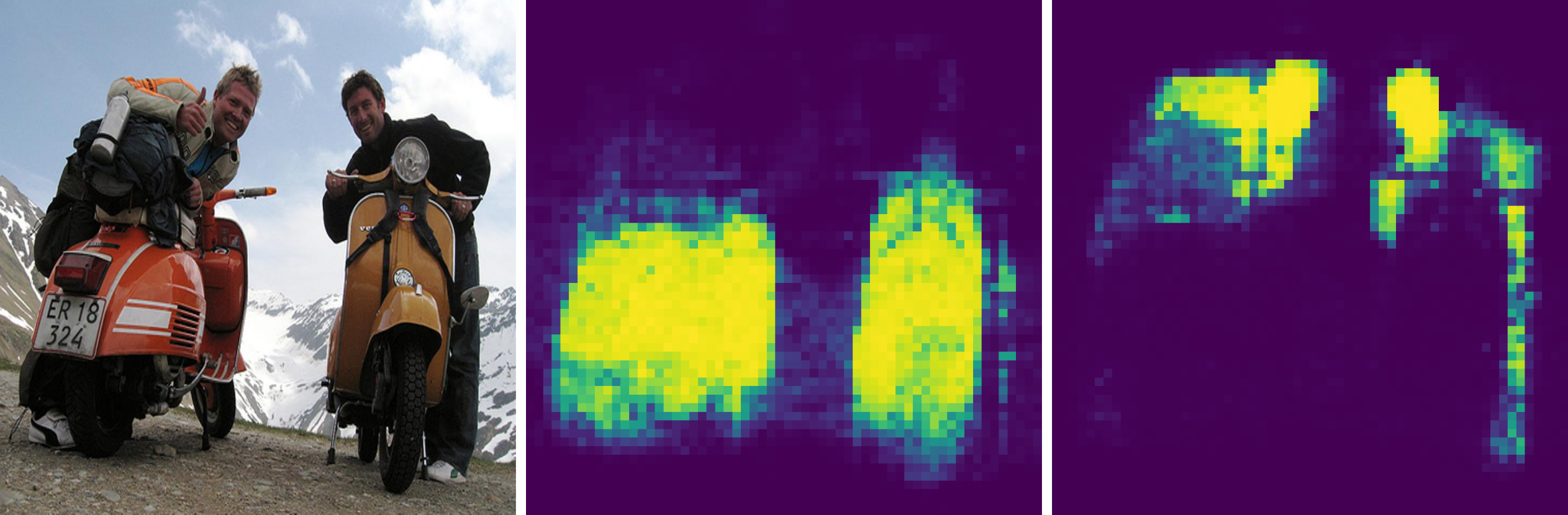}
\end{subfigure}\\
\begin{subfigure}{.31\textwidth}
\centering
\includegraphics[width=1.015\linewidth]{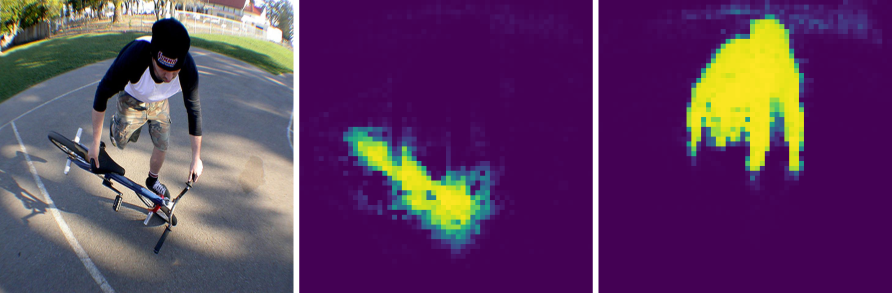}
\end{subfigure}
\begin{subfigure}{.31\textwidth}
\centering
\includegraphics[width=1.015\linewidth]{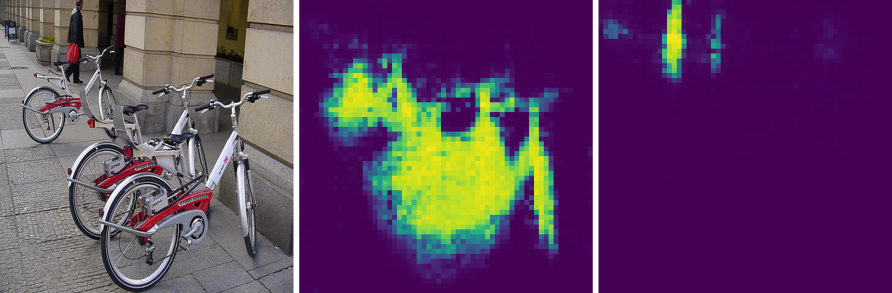}
\end{subfigure}
\begin{subfigure}{.31\textwidth}
\centering
\includegraphics[width=1.015\linewidth]{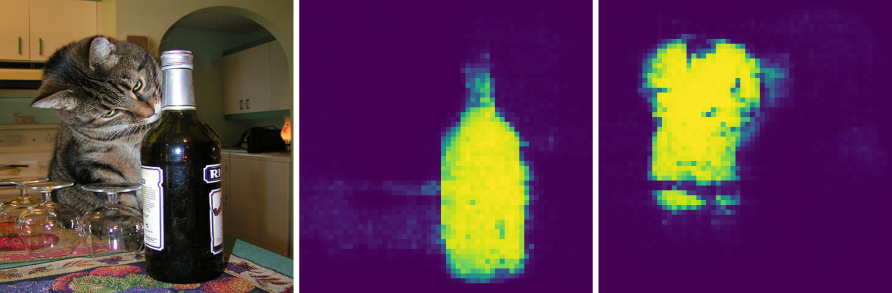}
\end{subfigure}\\
\hspace{-0.45em}
\begin{subfigure}{.413333\textwidth}
\centering
\includegraphics[width=1.015\linewidth]{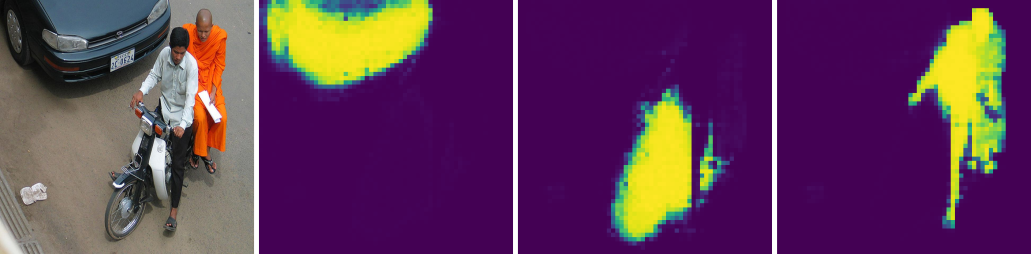}
\end{subfigure}\hspace{0.22em}
\begin{subfigure}{.516666\textwidth}
\centering
\includegraphics[width=1.015\linewidth]{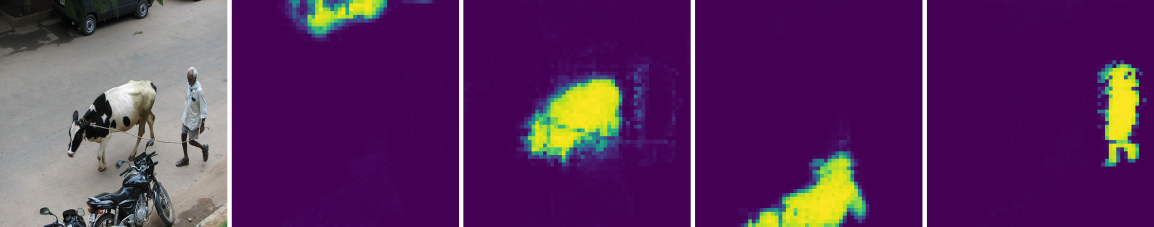}
\end{subfigure}\\
\caption{Each row shows the original images (two or three per row) and the BPM for each class appearing in the image on its right. The BPM  of size $60{\times}60$ are  resized to the original image size.  }\label{fig:BPM}
\end{figure}
\begin{figure}
\centering
\includegraphics[width=0.95\linewidth]{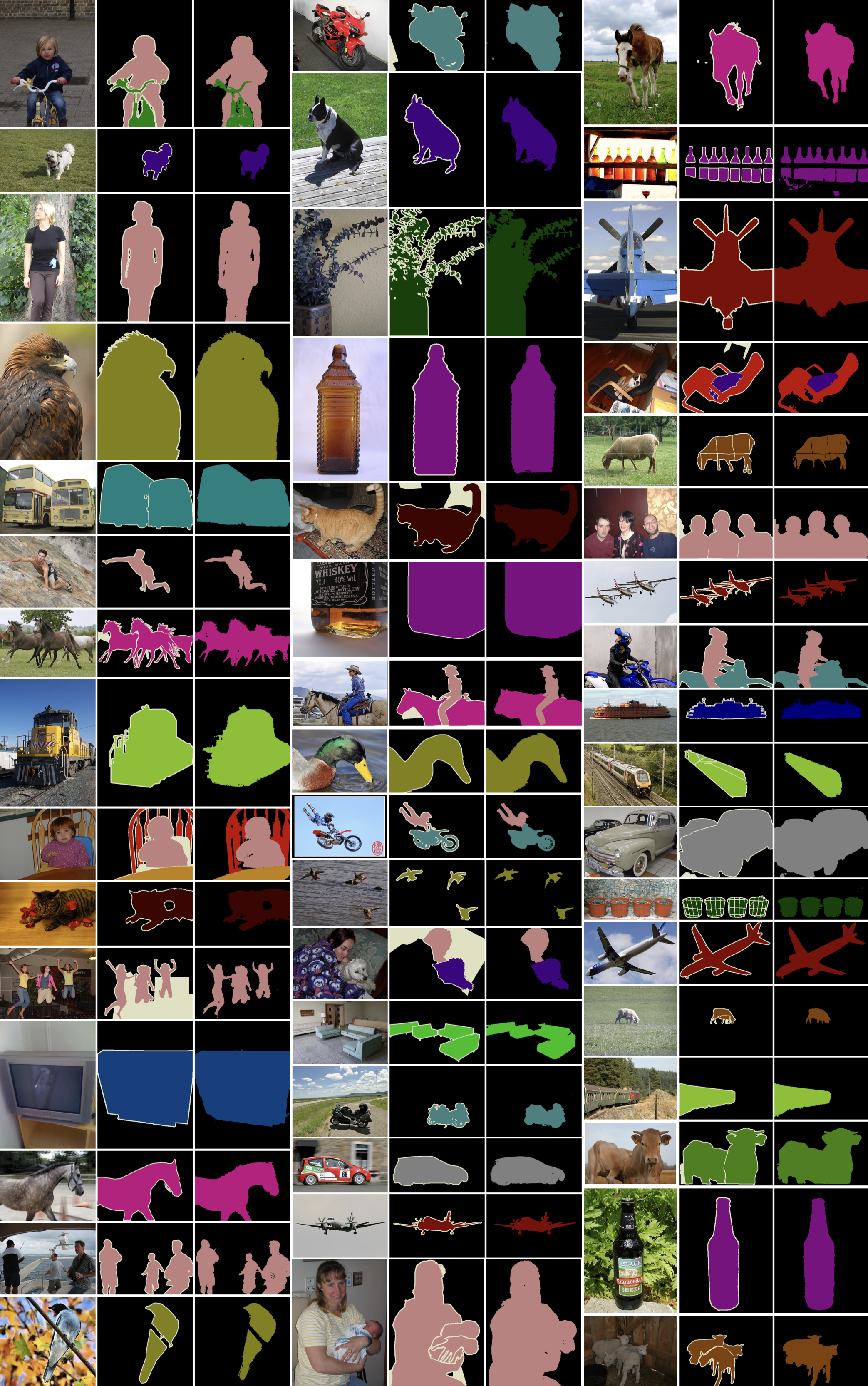}
\caption{Original image, ground-truth and final segmentation mask verified  with DeepLab V2 \cite{Chen2018DeepLabSI} trained on our BPM.}\label{fig:good-preds}
\end{figure}

In Figure \ref{fig:BPM}, we show more examples of the BMP inferred by ViT-PCM supervised by image-class labels. There are two or three original images per row, and the BPM for each class appearing in the image is on its right. We can note that the shapes are pretty accurate, but for the noise on the shape contour. 
In Figure \ref{fig:good-preds}, we show the best quality final pseudo-masks obtained by BPM plus CRF post-processing, verified by  DeepLabV2 trained on our BPM. The CRF of \cite{krahenbuhl2011efficient}, used as post-processing of the BPM, removes much of the noise, though it cannot improve a non-accurate shape.

\section{Per-Class Comparisons with state-of-the-art on the Verification Task in Pascal VOC 2012}
\vspace*{-0.3cm}
\begin{table}[h!]
\centering
\caption{Per-class performance comparison with the state-of-the-art WSSS methods  on the verification task with final-segmentation masks, in terms of mIoU\% on PASCAL VOC val set and test set. }
\label{tab:Table2}
\resizebox{\textwidth}{!}{%
\begin{tabular}{c | c  c  c  c  c  c  c  c  c  c  c  c  c  c  c  c  c  c  c  c c| c} 
 \hline
 \multicolumn{1}{c}{   Pascal VOC2012 val set}\\
\hline
 Method & bkg & aero & bike & bird & boat & btl & bus & car & cat & chair & cow & table & dog & horse & mbk & person & plant & sheep & sofa & train & tv & mIoU \\ 
 \hline
 CIAN\cite{Fan2020CIANCA}\tiny{\textsc{AAAI'20} } & 83.6 & 59.4 & 35.4 & 53.7 & 39.8 & 56.2 & 79.3 & 73.0 & 79.0 & 28.9 & 67.5 & 54.8 & 74.9 & 68.7 & 74.1 & 58.7 & 49.5 & 72.3 & 33.5 & 57.3 & 50.4 & 59.5 \\
 SEAM \cite{wang2020self}\tiny{\textsc{CVPR'20}} &  88.8 &68.5 &33.3 &85.7 &40.4 &67.3 &78.9 &76.3 &81.9 &29.1 &75.5 &48.1 &79.9 &73.8 &71.4 &75.2 &48.9 &79.8 &40.9 &58.2 &53.0 &64.5  \\
 BES\cite{chen2020weakly}\tiny{\textsc{ECCV'20}}  &  88.9 & 74.1 &29.8 &81.3 &53.3 &69.9 &89.4 &79.8 &84.2 &27.9 &76.9 &46.6 &78.8 &75.9 &72.2 &70.4 &50.8 &79.4 &39.9 &65.3 &44.8 &65.7 \\
 ECSNet\cite{sun2021ecs}\tiny{\textsc{ICCV'20}} &89.8 &68.4 &33.4 &85.6 &48.6 &72.2 &87.4 &78.1 &86.8 &33.0 &77.5 &41.6 &81.7 &76.9 &75.4 &75.6 &46.2 &80.7 &43.9 &59.8 &56.3 &66.6 \\
 AdvCAM\cite{lee2021anti}\tiny{\textsc{CVPR'21} } &  89.5 &76.9 &33.5 &80.3 &63.7 &68.6 &89.7 &77.9 &87.6 &31.6 &77.2 &36.2 &82.6 &78.7 &73.5 &69.8 &51.9 &81.9 &43.8& 70.9 &52.6 &67.5\\
 CPN\cite{zhang2021complementary}\tiny{\textsc{ICCV'21} } & 89.9 &75.1 &32.9 &87.8 &60.9 &69.5 &87.7 &79.5 &89.0 &28.0 &80.9 &34.8 &83.4 &79.7 &74.7 &66.9 &56.5 &82.7 &44.9 &73.1 &45.7 &67.8\\
 CSE\cite{kweon2021unlocking}\tiny{\textsc{ICCV'21} } & 90.2 & 82.9 &35.1 &86.8 &59.4& 70.6 &82.5 &78.1 &87.4 &30.1 &79.4 &45.9 &83.1 &83.4 &75.7 &73.4 &48.1 &89.3 &42.7 &60.4 & 52.3 &68.4\\
 W-OoD\cite{lee2022weakly}\tiny{\textsc{CVPR'22} } &  91.2 & 80.1 & 34.0& 82.5 &68.5 &72.9 &90.3 &80.8 &89.3 & 32.3 &78.9 &31.1 &83.6 &79.2 &75.4 &74.4 &58.0 &81.9 &45.2 &81.3 &54.8 &69.8  \\
 MCT-Former\cite{xu2022multi}\tiny{\textsc{CVPR'22} } &  91.9 & 78.3 &39.5 &89.9 &55.9 &76.7 &81.8 &79.0 &90.7 &32.6 &87.1 &57.2 &87.0 &84.6 &77.4 &79.2 &55.1 &89.2 &47.2 &70.4 &58.8 &71.9 \\
 \hline
 \rowcolor{gray!13}
  \multicolumn{1}{c}{  End-to-end methods}\\
  \hline
\rowcolor{gray!21}
            PAMR\cite{araslanov2020single}\tiny{\textsc{CVPR'20} } & 88.7 & 70.4  &35.1  &75.7  &  51.9 & 65.8 & 71.9 & 64.2 & 81.1 & 30.8 & 73.3 & 28.1 & 81.6 & 69.1 &  62.6&  74.8 &48.6& 71.0 & 40.1 & 68.5 & 64.3 &62.7  \\
\rowcolor{gray!13}
            ICD\cite{Fan_2020_CVPR}\tiny{\textsc{CVPR'20} } & 82.4 & 67.6 & 46.1 & 63.5 & 51.9 & 53.2 & 76.1 & 68.6 & 74.6 & 24.4 & 71.2 & 31.4 & 62.1 & 70.6 & 73.0 & 10.5 & 49.1 & 74.6 & 31.6 & 69.0 & 33.4 & 56.4 \\ 
\rowcolor{gray!21}
            AFA\cite{ru2022learning}\tiny{\textsc{CVPR'22} } & 89.9 &79.5   &31.2  &80.7  &67.2  &61.9  &81.4  &65.4  & 82.3 &28.7  &83.4  &41.6 & 82.2 &75.9 &70.2  &69.4  & 53.0  &85.9  &44.1  &64.2  &50.9  & 66.0 \\
\rowcolor{gray!13}
MCT-Former$^{\star}$\cite{xu2022multi}\tiny{\textsc{CVPR'22} } & 90.6 & 71.8 &37.5& 85.1 &52.9 &68.8 &78.8 &78.7 &87.1 &28.4 &78.9 &53.0 &83.9 &78.2 &76.8 &76.4 &54.1& 80.1& 46.0 &71.6 &54.3 &68.2 \\
\rowcolor{gray!21}
           {\bf ViT-PCM Ours} & 91.2& {\bf 86.0}&	37.8&	83.7&	{\bf 67.1}&	70.2&	{\bf 90.4}&	{\bf 85.0} &	90.2&	29.5&	82.1&	{\bf 57.3}&	84.1&	78.3&	{\bf 77.7} &	{\bf 83.5}&	53.0&	78.7&	22.7&	{\bf 82.6}&	44.8& 70.3 \\ 
         \hline

 \hline
\multicolumn{1}{c}{   Pascal VOC2012 test set}\\
 \hline

 CIAN\cite{Fan2020CIANCA} & 82.1 & 57.6 & 28.5 & 49.2 & 36.5 & 58.9 & 84.6 & 72.4 & 76.6 & 23.3 & 68.4 & 47.0 & 72.1 & 66.8 & 70.6 & 61.2 & 39.4 & 64.1 & 34.6 & 55.8 & 47.4 & 57.0 \\
 AdvCam\cite{wang2020self}\tiny{\textsc{CVPR'20}} &  90.1 &81.2 &33.6 &80.4 &52.4 &66.6 &87.1 &80.5 &87.2 &28.9 &80.1 &38.5 &84.0 &83.0 &79.5 &71.9 &47.5 &80.8 &59.1& 65.4& 49.7& 68.0  \\
 CPN\cite{zhang2021complementary}\tiny{\textsc{ICCV'21} } & 90.4 &79.8 &32.9 &85.8 &52.9 &66.4 &87.2 &81.4 &87.6 &28.2 &79.7 &50.2 &82.9 &80.4 &78.9& 70.6 &51.2 &83.4& 55.4& 68.5 &44.6 &68.5\\
 W-OoD\cite{lee2022weakly}\tiny{\textsc{CVPR'22} } &91.4 & 85.3 & 32.8 & 79.8 & 59.0 & 68.4 & 88.1 &82.2 &88.3 &27.4 &76.7 &38.7 &84.3 &81.1 &80.3 &72.8 &57.8 &82.4 &59.5& 79.5 &52.6 &69.9  \\
 MCT-Former\cite{xu2022multi}\tiny{\textsc{CVPR'22} } &  92.3 & 84.4 &37.2 &82.8 &60.0 &72.8 &78.0 &79.0 &89.4 &31.7 &84.5 &59.1 &85.3 &83.8 &79.2 &81.0 &53.9 &85.3 &60.5 &65.7 &57.7 &71.6 \\
 \hline
 \rowcolor{gray!13}
 \multicolumn{1}{c}{  End-to-end methods}\\
  \hline
 \rowcolor{gray!13}
            PAMR\cite{araslanov2020single}\tiny{\textsc{CVPR'20} } & 89.2 & 73.4 & 37.3 &68.3 &45.8 &68.0& 72.7 &64.1 &74.1 &32.9 &74.9 &39.2 &81.3 &74.6 &72.6 &75.4 &58.1 &71.0 &48.7 &67.7 &60.1& 64.3 \\
\rowcolor{gray!21}
           ICD\cite{Fan_2020_CVPR} & 83.7 & 75.3 & 31.4 & 68.8 & 56.1 & 63.4 & 87.6 & 77.2 & 76.6 & 25.0 & 72.4 & 37.2 & 67.4 & 73.0 & 70.1 & 7.6 & 46.0 & 79.8 & 31.2 & 75.0 & 33.3 & 59.0\\ 
\rowcolor{gray!13}
MCT-Former$^{\star}$\cite{xu2022multi}\tiny{\textsc{CVPR'22} } & 90.9 & 76.0 & 37.2 & 79.1 & 54.1 & 69.0 & 78.1& 78.0 &86.1 & 30.3 &79.5 & 58.3 & 81.7 & 81.1 &77.0 & 76.4 & 49.2 &80.0 &55.1 &65.4 &54.5 &68.4 \\
\rowcolor{gray!21}
{\bf ViT-PCM Ours} &  91.1&	{\bf 88.9}&	39.0&	{\bf 87.0}&	58.8&	69.4&	{\bf 89.4}&	{\bf 85.4}&	{\bf 89.9}&	30.7&	82.6&	{\bf 62.2}&	{\bf 85.7}&	83.6&	{\bf 79.7}&	{\bf 81.6}&	52.1&	82.0&	26.5&	{\bf 80.3}&	42.4 &70.9  \\ 
         \hline
\end{tabular}
}
\end{table}

In Table \ref{tab:Table1} we report the per-class comparison with other WSSS methods on val and test set of Pascal VOC 2012. Not all methods report both the val and the test set. We divide the methods into two sets: those using boosting and those end-to-end. An explanation of the computational effort of boosting w.r.t. the end-to-end networks is also given in their paper supplements in PAMR \cite{araslanov2020single}. It is interesting to note that, according to the PAMR's authors,  methods such as PSA\cite{ahn2018learning}, and IRNet\cite{ahn2019weakly}, have three stages and additionally train a standalone segmentation network. The end-to-end methods are highlighted in grey. Thanks to MCT-Former, operated in the two versions (boosted and end-to-end, this last indicated by a ${\star}$), we can appreciate the difference between the two approaches. Among all methods, our ViT-PCM is second to MCT-Former, boosted with PSA\cite{ahn2018learning}. W.r.t the end-to-end methods ViT-PCM advances the state-of-the-art on all categories and improves the results of 2.1\% on the val set and 2.5\% on the test set.

\vspace*{-0.5cm}

\section{Qualitative results on MS-COCO val set}
\begin{figure}
\centering
\includegraphics[width=0.95\linewidth]{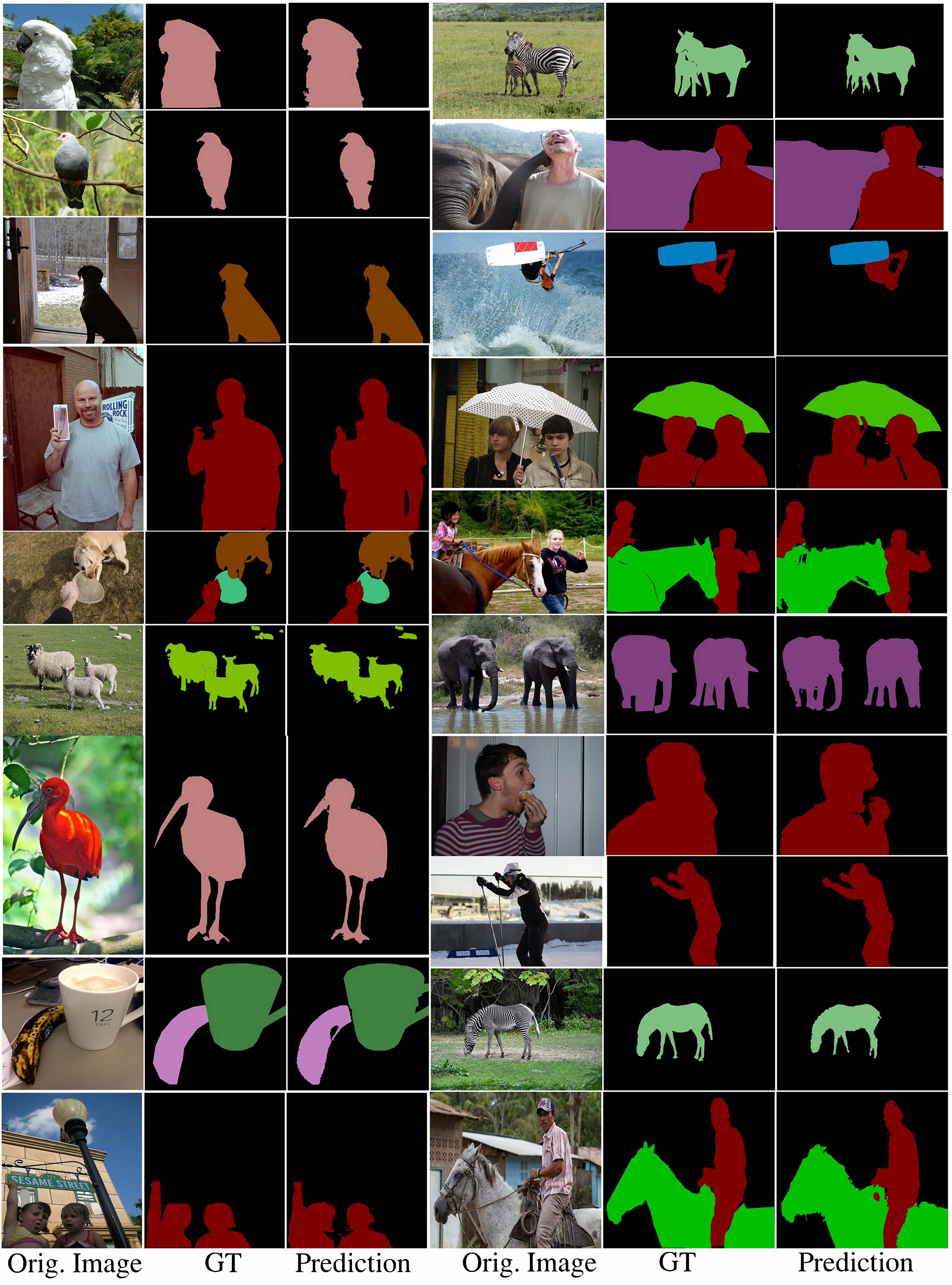}
\caption{Qualitative results on MS-COCO 2014 val set.}\label{fig:good-preds}
\end{figure}
In Figure \ref{fig:good-preds}, we show some results on the final-segmentation masks for MS-COCO 2014 val set. All the shown images are chosen among those with mIoU\% greater than 95\%. We can note that some results are even better than the ground-truth annotations, e.g. the man-eating or the two elephants, where the annotation does not consider the space inside the shapes, which instead ViT-PCM recognizes. 

\section{Per-Class Comparisons with state-of-the-art on Verification Task in MS-COCO 2014}
\vspace*{-0.8cm}
\begin{table}[h!]
\centering
\caption{Per-class performance comparison with the state-of-the-art WSSS methods  on the verification task with final-segmentation masks, in terms of mIoU\% on MS-COCO 2014 \textit{val} set. }\label{tab:Table1}
\resizebox{\textwidth}{!}{%
\begin{tabular}{c  c  c  c |  c  c  c  c  } 
 \hline
 Class & AuxSegNet\tiny{\textsc{ICCV'21} } & MCT-Former\tiny{\textsc{CVPR'22} } & ViT-PCM & 
 Class & AuxSegNet\tiny{\textsc{ICCV'21} } & MCT-Former\tiny{\textsc{CVPR'22} } & ViT-PCM \\
 &  \cite{xu2021leveraging}  & \cite{xu2022multi} & Ours   & 
 & \cite{xu2021leveraging}  & \cite{xu2022multi} & Ours \\
         \hline
background &82.0&82.4 & 81.9
  &wine-glass&32.1&27.0& 38.2\\
person &65.4&62.6& 62.4
  &cup&29.3&29.& 40.9\\
bicycle&43.0&47.4& 54.3
   &fork&5.4&13.9& 33.3\\
car&34.5&47.2& 49.2
   &knife&1.4&12.0& 31.0\\
motorcycle&66.2&63.7& 70.3 
   &spoon&1.4&6.6& 21.4\\
airplane&60.3&64.7& 74.5
   &bowl&19.5&22.4& 36.2\\
bus&63.1&64.5& 76.0
   &banana&46.9&63.2& 58.6\\
train&57.3&64.5& 61.2
   &apple&40.4&44.4& 52.1\\
truck&38.9&44.8& 45.3
    &sandwich&39.4&39.7& 57.1\\
boat&30.1&42.3& 47.8
   &orange&52.9&63.0& 55.8\\
traffic-light&40.4&49.9& 22.2
   &broccoli&36.0&51.2& 53.5\\
fire-hydrant&72.7&73.2& 78.8
   &carrot&13.9&40.0& 45.0\\
stop-sign&40.3&76.6& 11.0
   & hot-dog&46.1&53.0& 41.4\\
parking-meter&59.8&64.4& 65.5
   &pizza&62.0&62.2& 77.6\\
bench&16.0&32.8& 42.6
   &donut&43.9&55.7& 39.4\\
bird&61.0&62.6& 67.0
    &cake&30.6&47.9& 63.0\\
cat&68.6&78.2& 20.4
    &chair&11.4&22.8& 35.6\\
dog &66.9&68.2& 71.7
    &couch&14.5&35.0& 41.7\\
horse&55.6&65.8& 68.6
    &potted-plant&2.1&13.5& 37.9\\
sheep&61.4&70.1& 67.2
    &bed&20.5&48.6& 53.2 \\
cow&60.7&68.3& 70.4
    &dining-table&9.5&12.9& 29.4\\
elephant&76.1&81.6&83.3 
    &toilet&57.8&63.1& 67.3\\
bear&73.0&80.1& 74.2
    &tv &36.0&47.9& 38.7\\
zebra&80.8&83.0 & 72.6
    &laptop&35.2&49.5& 51.7\\
giraffe&71.6&76.9& 67.3
    &mouse&13.4&13.4& 13.9\\
backpack&11.3&14.6& 24.3
    &remote&23.6&41.9& 34.2\\
umbrella&35.0&61.7& 67.7
     &keyboard&17.9&49.8& 65.0\\
handbag&2.2&4.5& 19.4
     &cellphone&49.9&54.1& 56.8\\
tie&14.7&25.2& 19.0
     &microwave&28.7&38.0& 50.2\\
suitcase&31.7&46.8 &47.6 
     &oven&13.3&29.9& 35.8\\
frisbee&1.0&43.8& 38.1
     & toaster&0.0&0.0& 13.8\\
skis&8.1&12.8& 20.3
     &sink&21.0&28.0& 14.3\\
snowboard&7.6&31.4&41.6 
     &refrigerator&16.6&40.1& 44.9 \\
sports-ball&28.8&9.2& 7.1
     &book&8.7&32.2& 40.6\\
kite&27.3&26.3& 41.5
     & clock&34.4& 43.2& 51.3\\
baseball-bat&2.2&0.9 &2.3 
    &vase&25.9&22.6& 25.0\\
baseball-glove&1.3&0.7& 5.0
    & scissors&16.6&32.9& 48.1\\
skateboard&15.2&7.8& 10.3
    &teddy-bear&47.3&61.9& 53.9\\
surfboard&17.8&46.5& 45.9
    &hair-drier&0.0&0.0& 13.4\\
tennis-racket&47.1&1.4& 16.1
    &toothbrush&1.4&12.2& 33.1\\
\cline{5-8}
bottle&33.2&31.1& 41.5
    & mIoU\% & 33.9 & 42.0  & 45.0\\
    \hline
\end{tabular}
}
\end{table}
 In Table \ref{tab:Table2} we expose per-class comparison on MS-COCO 2014 val set. We compare our results with AugSegNet \cite{xu2021leveraging} and with MCT-Former \cite{xu2022multi}. ViT-PCM outperforms the other two, though the results per class are highly variable. For example, on the class {\em stop-sign} we have an accuracy of 11.0\% while MCT-Former obtains 76.6\% and AugSegNet 40.3\%. On the other hand, for tiny objects such as {\em fork}, {\em knife} and {\em spoon}, we obtain resp. 33.3\%, 31.0\% and 21.4\% against a much lower accuracy obtained by the two competitors.

\clearpage

\bibliographystyle{styles/splncs04}
\bibliography{egbib.bib}